\documentclass{article}
\usepackage{arxiv}

\usepackage[utf8]{inputenc}

\usepackage{xcolor}
\usepackage{rotating}
\usepackage{xstring} %
\usepackage[nointegrals]{wasysym} %
\usepackage{float}
\usepackage{amssymb}

\usepackage{hyperref}       %
\usepackage{url}            %
\usepackage{booktabs}       %
\usepackage{tabularx}

\usepackage{nicefrac}       %
\usepackage{microtype}      %
\usepackage{optidef}

\usepackage[mode=buildnew]{standalone}%
\usepackage{pgfplots}
\usepgfplotslibrary{groupplots}
\usepackage{pgfplotstable}

\pgfplotsset{compat=newest}
\DeclareUnicodeCharacter{2212}{−}
\usepackage{tikz}
\usetikzlibrary{arrows.meta}
\usepackage{siunitx}
\usepackage{bm}
\usepackage{placeins}
\usepackage{graphicx}
\usepackage{natbib}

\usepackage{algpseudocode}
\algrenewcommand{\algorithmicrequire}{\textbf{Input:}}
\algrenewcommand{\algorithmicensure}{\textbf{Output:}}
\algnewcommand\algorithmicand{\textbf{and}}

\usepackage{caption}
\usepackage{subcaption}

\DeclareUnicodeCharacter{2217}{$\ast$}

\newcommand{\dimF}{n}

\newcommand{\queueSize}{Q}
\newcommand{\popSize}{K}

\DeclarePairedDelimiterX{\infdivx}[2]{(}{)}{%
	#1\;\delimsize\|\;#2%
}
\newcommand{\KLdiv}{\mathrm{KL}\infdivx}
\DeclarePairedDelimiter{\norm}{\lVert}{\rVert}

\newcommand{\dx}{\mathrm{d\bm{x}}}
\newcommand{\vecT}[1]{#1^\mathrm{T}}
\newcommand{\tr}[1]{\mathrm{tr}(#1)}
\DeclareMathOperator{\vect}{vec}

\DeclareMathOperator*{\argmin}{arg\,min}

\DeclarePairedDelimiter\ceil{\lceil}{\rceil}
\DeclarePairedDelimiter\floor{\lfloor}{\rfloor}

\newcommand{\bbobdatapath}{figures/ppdata/} %
\providecommand{\bbobecdfcaptionsinglefunctionssingledim}[1]{
Empirical cumulative distribution of simulated (bootstrapped)
             runtimes, measured in number of objective function evaluations,
             divided by dimension (FEvals/DIM) for the $51$ 
             targets $10^{[-8..2]}$ in dimension #1.
}

\providecommand{\bbobECDFslegend}[1]{
Bootstrapped empirical cumulative distribution of the number of objective function evaluations divided by dimension (FEvals/DIM) for $51$ targets with target precision in $10^{[-8..2]}$ for all functions and subgroups in #1-D. As reference algorithm, the best algorithm from BBOB 2009 is shown as light thick line with diamond markers.
}
\providecommand{\bbobppfigslegend}[1]{
Expected running time (\ERT\ in number of $f$-evaluations
                    as $\log_{10}$ value), divided by dimension for target function value $10^{-8}$
                    versus dimension. Slanted grid lines indicate quadratic scaling with the dimension. Different symbols correspond to different algorithms given in the legend of #1. Light symbols give the maximum number of function evaluations from the longest trial divided by dimension. Black stars indicate a statistically better result compared to all other algorithms with $p<0.01$ and Bonferroni correction number of dimensions (six).  
Legend: 
{\color{NavyBlue}$\circ$}: \algorithmA
, {\color{Magenta}$\diamondsuit$}: \algorithmB
, {\color{Orange}$\star$}: \algorithmC
, {\color{CornflowerBlue}$\triangledown$}: \algorithmD
}
\definecolor{NavyBlue}{HTML}{000080}
\definecolor{Magenta}{HTML}{FF00FF}
\definecolor{Orange}{HTML}{FFA500}
\definecolor{CornflowerBlue}{HTML}{6495ED}
\definecolor{YellowGreen}{HTML}{9ACD32}
\definecolor{Gray}{HTML}{BEBEBE}
\definecolor{Yellow}{HTML}{FFFF00}
\definecolor{GreenYellow}{HTML}{ADFF2F}
\definecolor{ForestGreen}{HTML}{228B22}
\definecolor{Lavender}{HTML}{FFC0CB}
\definecolor{SkyBlue}{HTML}{87CEEB}
\definecolor{NavyBlue}{HTML}{000080}
\definecolor{Goldenrod}{HTML}{DDF700}
\definecolor{VioletRed}{HTML}{D02090}
\definecolor{CornflowerBlue}{HTML}{6495ED}
\definecolor{LimeGreen}{HTML}{32CD32}

\providecommand{\pptablesfooter}{
\end{tabularx}
}
\providecommand{\pptablesheader}{
\begin{tabularx}{1.0\textwidth}{@{}c@{}|*{7}{@{}r@{}X@{}}|@{}r@{}@{}l@{}}
$\Delta f_\mathrm{opt}$ & \multicolumn{2}{@{\,}l@{\,}}{1e1} & \multicolumn{2}{@{\,}l@{\,}}{1e0} & \multicolumn{2}{@{\,}l@{\,}}{1e-1} & \multicolumn{2}{@{\,}l@{\,}}{1e-2} & \multicolumn{2}{@{\,}l@{\,}}{1e-3} & \multicolumn{2}{@{\,}l@{\,}}{1e-5} & \multicolumn{2}{@{\,}l@{\,}}{1e-7} & \multicolumn{2}{|@{}l@{}}{\#succ}\\\hline
}

\providecommand{\pptablesfooter}{
\end{tabularx}
}
\providecommand{\pptablesheader}{
\begin{tabularx}{1.0\textwidth}{@{}c@{}|*{7}{@{}r@{}X@{}}|@{}r@{}@{}l@{}}
$\Delta f_\mathrm{opt}$ & \multicolumn{2}{@{\,}l@{\,}}{1e1} & \multicolumn{2}{@{\,}l@{\,}}{1e0} & \multicolumn{2}{@{\,}l@{\,}}{1e-1} & \multicolumn{2}{@{\,}l@{\,}}{1e-2} & \multicolumn{2}{@{\,}l@{\,}}{1e-3} & \multicolumn{2}{@{\,}l@{\,}}{1e-5} & \multicolumn{2}{@{\,}l@{\,}}{1e-7} & \multicolumn{2}{|@{}l@{}}{\#succ}\\\hline
}

\providecommand{\pptablesfooter}{
\end{tabularx}
}
\providecommand{\pptablesheader}{
\begin{tabularx}{1.0\textwidth}{@{}c@{}|*{7}{@{}r@{}X@{}}|@{}r@{}@{}l@{}}
$\Delta f_\mathrm{opt}$ & \multicolumn{2}{@{\,}l@{\,}}{1e1} & \multicolumn{2}{@{\,}l@{\,}}{1e0} & \multicolumn{2}{@{\,}l@{\,}}{1e-1} & \multicolumn{2}{@{\,}l@{\,}}{1e-2} & \multicolumn{2}{@{\,}l@{\,}}{1e-3} & \multicolumn{2}{@{\,}l@{\,}}{1e-5} & \multicolumn{2}{@{\,}l@{\,}}{1e-7} & \multicolumn{2}{|@{}l@{}}{\#succ}\\\hline
}

\providecommand{\pptablesfooter}{
\end{tabularx}
}
\providecommand{\pptablesheader}{
\begin{tabularx}{1.0\textwidth}{@{}c@{}|*{7}{@{}r@{}X@{}}|@{}r@{}@{}l@{}}
$\Delta f_\mathrm{opt}$ & \multicolumn{2}{@{\,}l@{\,}}{1e1} & \multicolumn{2}{@{\,}l@{\,}}{1e0} & \multicolumn{2}{@{\,}l@{\,}}{1e-1} & \multicolumn{2}{@{\,}l@{\,}}{1e-2} & \multicolumn{2}{@{\,}l@{\,}}{1e-3} & \multicolumn{2}{@{\,}l@{\,}}{1e-5} & \multicolumn{2}{@{\,}l@{\,}}{1e-7} & \multicolumn{2}{|@{}l@{}}{\#succ}\\\hline
}

\providecommand{\pptablesfooter}{
\end{tabularx}
}
\providecommand{\pptablesheader}{
\begin{tabularx}{1.0\textwidth}{@{}c@{}|*{7}{@{}r@{}X@{}}|@{}r@{}@{}l@{}}
$\Delta f_\mathrm{opt}$ & \multicolumn{2}{@{\,}l@{\,}}{1e1} & \multicolumn{2}{@{\,}l@{\,}}{1e0} & \multicolumn{2}{@{\,}l@{\,}}{1e-1} & \multicolumn{2}{@{\,}l@{\,}}{1e-2} & \multicolumn{2}{@{\,}l@{\,}}{1e-3} & \multicolumn{2}{@{\,}l@{\,}}{1e-5} & \multicolumn{2}{@{\,}l@{\,}}{1e-7} & \multicolumn{2}{|@{}l@{}}{\#succ}\\\hline
}

\providecommand{\algsfolder}{CAS_M_MORE_CMA-E_xNESa/}
\providecommand{\algorithmA}{CAS-MORE}
\providecommand{\algorithmB}{MORE}
\providecommand{\algorithmC}{CMA-ES}
\providecommand{\algorithmD}{xNESas}
\providecommand{\bbobecdfcaptionsinglefunctionssingledim}[1]{
Empirical cumulative distribution of simulated (bootstrapped)
             runtimes, measured in number of objective function evaluations,
             divided by dimension (FEvals/DIM) for the $51$ 
             targets $10^{[-8..2]}$ in dimension #1.
}

\providecommand{\bbobECDFslegend}[1]{
Bootstrapped empirical cumulative distribution of the number of objective function evaluations divided by dimension (FEvals/DIM) for $51$ targets with target precision in $10^{[-8..2]}$ for all functions and subgroups in #1-D. As reference algorithm, the best algorithm from BBOB 2009 is shown as light thick line with diamond markers.
}
\providecommand{\bbobppfigslegend}[1]{
Expected running time (\ERT\ in number of $f$-evaluations
                    as $\log_{10}$ value), divided by dimension for target function value $10^{-8}$
                    versus dimension. Slanted grid lines indicate quadratic scaling with the dimension. Different symbols correspond to different algorithms given in the legend of #1. Light symbols give the maximum number of function evaluations from the longest trial divided by dimension. Black stars indicate a statistically better result compared to all other algorithms with $p<0.01$ and Bonferroni correction number of dimensions (six).  
Legend: 
{\color{NavyBlue}$\circ$}: \algorithmA
, {\color{Magenta}$\diamondsuit$}: \algorithmB
, {\color{Orange}$\star$}: \algorithmC
}
\definecolor{NavyBlue}{HTML}{000080}
\definecolor{Magenta}{HTML}{FF00FF}
\definecolor{Orange}{HTML}{FFA500}
\definecolor{CornflowerBlue}{HTML}{6495ED}
\definecolor{YellowGreen}{HTML}{9ACD32}
\definecolor{Gray}{HTML}{BEBEBE}
\definecolor{Yellow}{HTML}{FFFF00}
\definecolor{GreenYellow}{HTML}{ADFF2F}
\definecolor{ForestGreen}{HTML}{228B22}
\definecolor{Lavender}{HTML}{FFC0CB}
\definecolor{SkyBlue}{HTML}{87CEEB}
\definecolor{NavyBlue}{HTML}{000080}
\definecolor{Goldenrod}{HTML}{DDF700}
\definecolor{VioletRed}{HTML}{D02090}
\definecolor{CornflowerBlue}{HTML}{6495ED}
\definecolor{LimeGreen}{HTML}{32CD32}

\providecommand{\pptablesfooter}{
\end{tabularx}
}
\providecommand{\pptablesheader}{
\begin{tabularx}{1.0\textwidth}{@{}c@{}|*{7}{@{}r@{}X@{}}|@{}r@{}@{}l@{}}
$\Delta f_\mathrm{opt}$ & \multicolumn{2}{@{\,}l@{\,}}{1e1} & \multicolumn{2}{@{\,}l@{\,}}{1e0} & \multicolumn{2}{@{\,}l@{\,}}{1e-1} & \multicolumn{2}{@{\,}l@{\,}}{1e-2} & \multicolumn{2}{@{\,}l@{\,}}{1e-3} & \multicolumn{2}{@{\,}l@{\,}}{1e-5} & \multicolumn{2}{@{\,}l@{\,}}{1e-7} & \multicolumn{2}{|@{}l@{}}{\#succ}\\\hline
}

\providecommand{\pptablesfooter}{
\end{tabularx}
}
\providecommand{\pptablesheader}{
\begin{tabularx}{1.0\textwidth}{@{}c@{}|*{7}{@{}r@{}X@{}}|@{}r@{}@{}l@{}}
$\Delta f_\mathrm{opt}$ & \multicolumn{2}{@{\,}l@{\,}}{1e1} & \multicolumn{2}{@{\,}l@{\,}}{1e0} & \multicolumn{2}{@{\,}l@{\,}}{1e-1} & \multicolumn{2}{@{\,}l@{\,}}{1e-2} & \multicolumn{2}{@{\,}l@{\,}}{1e-3} & \multicolumn{2}{@{\,}l@{\,}}{1e-5} & \multicolumn{2}{@{\,}l@{\,}}{1e-7} & \multicolumn{2}{|@{}l@{}}{\#succ}\\\hline
}

\providecommand{\pptablesfooter}{
\end{tabularx}
}
\providecommand{\pptablesheader}{
\begin{tabularx}{1.0\textwidth}{@{}c@{}|*{7}{@{}r@{}X@{}}|@{}r@{}@{}l@{}}
$\Delta f_\mathrm{opt}$ & \multicolumn{2}{@{\,}l@{\,}}{1e1} & \multicolumn{2}{@{\,}l@{\,}}{1e0} & \multicolumn{2}{@{\,}l@{\,}}{1e-1} & \multicolumn{2}{@{\,}l@{\,}}{1e-2} & \multicolumn{2}{@{\,}l@{\,}}{1e-3} & \multicolumn{2}{@{\,}l@{\,}}{1e-5} & \multicolumn{2}{@{\,}l@{\,}}{1e-7} & \multicolumn{2}{|@{}l@{}}{\#succ}\\\hline
}

\providecommand{\pptablesfooter}{
\end{tabularx}
}
\providecommand{\pptablesheader}{
\begin{tabularx}{1.0\textwidth}{@{}c@{}|*{7}{@{}r@{}X@{}}|@{}r@{}@{}l@{}}
$\Delta f_\mathrm{opt}$ & \multicolumn{2}{@{\,}l@{\,}}{1e1} & \multicolumn{2}{@{\,}l@{\,}}{1e0} & \multicolumn{2}{@{\,}l@{\,}}{1e-1} & \multicolumn{2}{@{\,}l@{\,}}{1e-2} & \multicolumn{2}{@{\,}l@{\,}}{1e-3} & \multicolumn{2}{@{\,}l@{\,}}{1e-5} & \multicolumn{2}{@{\,}l@{\,}}{1e-7} & \multicolumn{2}{|@{}l@{}}{\#succ}\\\hline
}

\providecommand{\pptablesfooter}{
\end{tabularx}
}
\providecommand{\pptablesheader}{
\begin{tabularx}{1.0\textwidth}{@{}c@{}|*{7}{@{}r@{}X@{}}|@{}r@{}@{}l@{}}
$\Delta f_\mathrm{opt}$ & \multicolumn{2}{@{\,}l@{\,}}{1e1} & \multicolumn{2}{@{\,}l@{\,}}{1e0} & \multicolumn{2}{@{\,}l@{\,}}{1e-1} & \multicolumn{2}{@{\,}l@{\,}}{1e-2} & \multicolumn{2}{@{\,}l@{\,}}{1e-3} & \multicolumn{2}{@{\,}l@{\,}}{1e-5} & \multicolumn{2}{@{\,}l@{\,}}{1e-7} & \multicolumn{2}{|@{}l@{}}{\#succ}\\\hline
}

\providecommand{\algsfolder}{CAS_M_CMA-E_xNESa/}
\providecommand{\algorithmA}{CAS MORE}
\providecommand{\algorithmB}{CMA-ES}
\providecommand{\algorithmC}{xNESas}
\providecommand{\bbobecdfcaptionsinglefunctionssingledim}[1]{
Empirical cumulative distribution of simulated (bootstrapped)
             runtimes, measured in number of objective function evaluations,
             divided by dimension (FEvals/DIM) for the $51$ 
             targets $10^{[-8..2]}$ in dimension #1.
}

\providecommand{\bbobECDFslegend}[1]{
Bootstrapped empirical cumulative distribution of the number of objective function evaluations divided by dimension (FEvals/DIM) for $51$ targets with target precision in $10^{[-8..2]}$ for all functions and subgroups in #1-D. As reference algorithm, the best algorithm from BBOB 2009 is shown as light thick line with diamond markers.
}
\providecommand{\bbobppfigslegend}[1]{
Expected running time (\ERT\ in number of $f$-evaluations
                    as $\log_{10}$ value), divided by dimension for target function value $10^{-8}$
                    versus dimension. Slanted grid lines indicate quadratic scaling with the dimension. Different symbols correspond to different algorithms given in the legend of #1. Light symbols give the maximum number of function evaluations from the longest trial divided by dimension. Black stars indicate a statistically better result compared to all other algorithms with $p<0.01$ and Bonferroni correction number of dimensions (six).  
Legend: 
{\color{NavyBlue}$\circ$}: \algorithmA
, {\color{Magenta}$\diamondsuit$}: \algorithmB
, {\color{Orange}$\star$}: \algorithmC
, {\color{CornflowerBlue}$\triangledown$}: \algorithmD
, {\color{red}$\varhexagon$}: \algorithmE
}
\definecolor{NavyBlue}{HTML}{000080}
\definecolor{Magenta}{HTML}{FF00FF}
\definecolor{Orange}{HTML}{FFA500}
\definecolor{CornflowerBlue}{HTML}{6495ED}
\definecolor{YellowGreen}{HTML}{9ACD32}
\definecolor{Gray}{HTML}{BEBEBE}
\definecolor{Yellow}{HTML}{FFFF00}
\definecolor{GreenYellow}{HTML}{ADFF2F}
\definecolor{ForestGreen}{HTML}{228B22}
\definecolor{Lavender}{HTML}{FFC0CB}
\definecolor{SkyBlue}{HTML}{87CEEB}
\definecolor{NavyBlue}{HTML}{000080}
\definecolor{Goldenrod}{HTML}{DDF700}
\definecolor{VioletRed}{HTML}{D02090}
\definecolor{CornflowerBlue}{HTML}{6495ED}
\definecolor{LimeGreen}{HTML}{32CD32}

\providecommand{\pptablesfooter}{
\end{tabularx}
}
\providecommand{\pptablesheader}{
\begin{tabularx}{1.0\textwidth}{@{}c@{}|*{7}{@{}r@{}X@{}}|@{}r@{}@{}l@{}}
$\Delta f_\mathrm{opt}$ & \multicolumn{2}{@{\,}l@{\,}}{1e1} & \multicolumn{2}{@{\,}l@{\,}}{1e0} & \multicolumn{2}{@{\,}l@{\,}}{1e-1} & \multicolumn{2}{@{\,}l@{\,}}{1e-2} & \multicolumn{2}{@{\,}l@{\,}}{1e-3} & \multicolumn{2}{@{\,}l@{\,}}{1e-5} & \multicolumn{2}{@{\,}l@{\,}}{1e-7} & \multicolumn{2}{|@{}l@{}}{\#succ}\\\hline
}

\providecommand{\pptablesfooter}{
\end{tabularx}
}
\providecommand{\pptablesheader}{
\begin{tabularx}{1.0\textwidth}{@{}c@{}|*{7}{@{}r@{}X@{}}|@{}r@{}@{}l@{}}
$\Delta f_\mathrm{opt}$ & \multicolumn{2}{@{\,}l@{\,}}{1e1} & \multicolumn{2}{@{\,}l@{\,}}{1e0} & \multicolumn{2}{@{\,}l@{\,}}{1e-1} & \multicolumn{2}{@{\,}l@{\,}}{1e-2} & \multicolumn{2}{@{\,}l@{\,}}{1e-3} & \multicolumn{2}{@{\,}l@{\,}}{1e-5} & \multicolumn{2}{@{\,}l@{\,}}{1e-7} & \multicolumn{2}{|@{}l@{}}{\#succ}\\\hline
}

\providecommand{\pptablesfooter}{
\end{tabularx}
}
\providecommand{\pptablesheader}{
\begin{tabularx}{1.0\textwidth}{@{}c@{}|*{7}{@{}r@{}X@{}}|@{}r@{}@{}l@{}}
$\Delta f_\mathrm{opt}$ & \multicolumn{2}{@{\,}l@{\,}}{1e1} & \multicolumn{2}{@{\,}l@{\,}}{1e0} & \multicolumn{2}{@{\,}l@{\,}}{1e-1} & \multicolumn{2}{@{\,}l@{\,}}{1e-2} & \multicolumn{2}{@{\,}l@{\,}}{1e-3} & \multicolumn{2}{@{\,}l@{\,}}{1e-5} & \multicolumn{2}{@{\,}l@{\,}}{1e-7} & \multicolumn{2}{|@{}l@{}}{\#succ}\\\hline
}

\providecommand{\pptablesfooter}{
\end{tabularx}
}
\providecommand{\pptablesheader}{
\begin{tabularx}{1.0\textwidth}{@{}c@{}|*{7}{@{}r@{}X@{}}|@{}r@{}@{}l@{}}
$\Delta f_\mathrm{opt}$ & \multicolumn{2}{@{\,}l@{\,}}{1e1} & \multicolumn{2}{@{\,}l@{\,}}{1e0} & \multicolumn{2}{@{\,}l@{\,}}{1e-1} & \multicolumn{2}{@{\,}l@{\,}}{1e-2} & \multicolumn{2}{@{\,}l@{\,}}{1e-3} & \multicolumn{2}{@{\,}l@{\,}}{1e-5} & \multicolumn{2}{@{\,}l@{\,}}{1e-7} & \multicolumn{2}{|@{}l@{}}{\#succ}\\\hline
}

\providecommand{\pptablesfooter}{
\end{tabularx}
}
\providecommand{\pptablesheader}{
\begin{tabularx}{1.0\textwidth}{@{}c@{}|*{7}{@{}r@{}X@{}}|@{}r@{}@{}l@{}}
$\Delta f_\mathrm{opt}$ & \multicolumn{2}{@{\,}l@{\,}}{1e1} & \multicolumn{2}{@{\,}l@{\,}}{1e0} & \multicolumn{2}{@{\,}l@{\,}}{1e-1} & \multicolumn{2}{@{\,}l@{\,}}{1e-2} & \multicolumn{2}{@{\,}l@{\,}}{1e-3} & \multicolumn{2}{@{\,}l@{\,}}{1e-5} & \multicolumn{2}{@{\,}l@{\,}}{1e-7} & \multicolumn{2}{|@{}l@{}}{\#succ}\\\hline
}

\renewcommand{\algsfolder}{CAS_M_MORE_CMA-E_xNESa/} %
\graphicspath{{\bbobdatapath\algsfolder}}

\newcommand{\ERT}{\ensuremath{\mathrm{ERT}}}

\newcommand{\Df}{\ensuremath{\Delta f}}

\newcommand{\fopt}{\ensuremath{f_\mathrm{opt}}}
\newcommand{\ftarget}{\ensuremath{f_\mathrm{t}}}

\title{Regret-Aware Black-Box Optimization with Natural Gradients, Trust-Regions and Entropy Control}

\author{Maximilian Hüttenrauch \texttt{maximilian.huettenrauch@kit.edu} \\
       Department of Computer Science \\
       Karlsruhe Institute of Technology \\
       Karlsruhe
        \AND
       Gerhard Neumann \texttt{gerhard.neumann@kit.edu} \\
       Department of Computer Science \\
       Karlsruhe Institute of Technology \\
       Karlsruhe}
       
\renewcommand{\shorttitle}{Regret-Aware Black-Box Optimization}

\begin{document}

\maketitle

\begin{abstract}%
Most successful stochastic black-box optimizers, such as CMA-ES, use rankings of the individual samples to obtain a new search distribution. Yet, the use of rankings also introduces several issues such as (i) the underlying optimization objective is often unclear, i.e., we do not optimize the expected fitness. Further, (ii) while these algorithms typically produce a high-quality mean estimate of the search distribution, the produced samples can have poor quality as these algorithms are ignorant of the regret. Lastly, (iii) noisy fitness function evaluations may result in solutions that are highly sub-optimal on expectation. 
In contrast, stochastic optimizers that are motivated by policy gradients, such as the Model-based Relative Entropy Stochastic Search (MORE) algorithm, directly optimize the expected fitness function without the use of rankings. MORE can be derived by applying natural policy gradients and compatible function approximation, and is using information theoretic constraints to ensure the stability of the policy update. While MORE does not suffer from the listed limitations (i-iii), it often can not achieve state of the art performance in comparison to ranking based methods. Hence, 
we improve MORE by (a) decoupling the update of the mean and covariance of the search distribution allowing for more aggressive updates on the mean while keeping the update on the covariance conservative, (b) an improved entropy scheduling technique based on an evolution path which results in faster convergence and (c) a simplified and more effective model learning approach in comparison to the original paper.
We compare our algorithm to state of the art black-box optimization algorithms on standard optimization tasks as well as on episodic RL tasks in robotics where it is also crucial to have small regret. Our algorithm obtains a comparable performance in terms of the best found solution on the optimization benchmarks. Yet, it clearly outperforms ranking-based methods in terms of regret for the RL tasks and can also deal with noisy fitness evaluations. 
\end{abstract}

\section{Introduction}
\label{sec:intro}
Stochastic-Search algorithms \citep{spall2005introduction} are problem independent algorithms well-suited for black-box optimization (BBO) \citep{larson2019derivative} of an objective function.
They only require function evaluations and are used when the objective function cannot be modeled analytically and no gradient information is available.
This is often the case for real world problems such as robotics \citep{chatzilygeroudis2017black}, medical applications \citep{winter2008registration}, or forensic identification \citep{ibanez2009experimental}.

Typically, these algorithms maintain a search distribution over the optimization variables of the objective function.
Solution candidates are sampled, evaluated, and the parameters of the search distribution (e.g. mean and covariance for Gaussian search distributions) are then updated towards a more promising direction. 
This process is repeated until a satisfactory solution quality is found or a pre-defined budget of objective function evaluations is reached.

In this paper, we re-introduce Model-based Relative Entropy Stochastic Search (MORE) \citep{NIPS2015_36ac8e55}, a versatile, general purpose stochastic-search optimization algorithm.
Using insights from reinforcement learning (RL) and information-theoretic trust-regions, it aims to update the parameters of a Gaussian search distribution in the direction of the natural gradient \citep{kakade2001natural}.
To this end, MORE uses compatible function approximation \citep{sutton1999policy, pajarinen2019compatible} and learn a quadratic surrogate model of the objective function
and bound the Kullback Leibler (KL) divergence between subsequent search distributions.
In its original formulation, a bound on the Kullback-Leibler divergence and a bound on the loss of entropy between the old and new search distribution additionally acts as an exploration-exploitation trade-off to prevent pre-mature convergence of the algorithm.

In contrast to most other successful stochastic search algorithms, that utilize a ranking of objective function values for updating the search distribution's parameters \citep{hansen2016cma, wierstra2014natural, rubinstein2004cross}, MORE directly incorporates objective function values into the model learning process.
This is especially important for reinforcement learning tasks, where we also want to minimize the regret, i.e., also the quality of the generated samples during exploration matters and not just the quality of the final mean estimate of the search distribution.  
The regret is defined as expected difference in costs between the current solution and an optimal solution. As an RL-based algorithm, MORE is regret aware and, hence, the search distribution avoids regions with high costs, while ranking-based approaches are ignorant to these regions and can easily produce samples in these regions.

However, the original MORE approach suffers from (a) the need for conservative KL bounds to keep the covariance updates stable, (b), a fixed entropy decreasing schedule resulting in suboptimal exploration and slow convergence, and (c), it employs an unnecessary complicated and inaccurate model learning method.
We aim to fix these issues by (a) splitting the KL divergence into separate updates for the mean and covariance with separate trust region bounds, (b) introducing an adaptive entropy schedule based on an evolution path, and (c) using ordinary least squares with appropriate data pre-processing techniques to simplify the model learning process.

We empirically evaluate our algorithm on simulated robotics tasks, as well as a set of benchmark optimization functions.
While we are competitive with state-of-the-art algorithms such as CMA-ES on the benchmark optimization functions,  the RL experiments clearly show the strength of MORE in finding well performing solutions in noisy conditions while simultaneously minimizing regret.

\section{Related Work}
The MORE algorithm can be seen as an instance of Evolution Strategies \citep{beyer2002evolution} from the broader class of Evolutionary Algorithms.
The usual procedure involves sampling from the search distribution, evaluating the candidates on the objective function, and improve the search distribution based on the candidates with the highest fitness while discarding poorly performing candidates.
In the following sections, we describe in more depth current state of the art algorithms from this category and review common design choices and algorithmic procedures.

\subsection{Ranking-based Algorithms}
Instead of directly incorporating function values into the optimization process, many algorithms apply a ranking based transformation before updating parameters.
While this makes the optimization process invariant to certain transformations and adds to the robustness of algorithms, it can have severe downsides in reinforcement learning problems, where regret is an additional notion of performance.
In noisy problems, it leads to an over- or underestimation of a samples performance and requires averaging over several sample evaluations in order to obtain a reliable estimate for the ranking \citep{hansen2008method, heidrich2009hoeffding}.

The cross-entropy method \citep{rubinstein2004cross, botev2013cross, amos2020differentiable} is one of the simplest representatives of ranking based algorithms. It evolves the search distribution by only incorporating an elite set of samples into the next generation which can be seen as a rank-based update.

The Covariance Matrix Adaptation - Evolution Strategy (CMA-ES) \citep{hansen2016cma} performs well established heuristics to update the mean and covariance matrix of a Gaussian search distribution as an interpolation of the weighted sample mean and covariance and the old mean and covariance. The weights of the samples are chosen according to the ranking of the samples.
Building on the basic CMA-ES approach, many derivative algorithms exist, mainly aiming at improving sample efficiency.
Notable extensions are those which include restarts with increasing population sizes \citep{auger2005restart}, restarts that alternate between large and small population sizes \citep{hansen2009benchmarking}, a version with decreasing step-sizes \citep{loshchilov2012alternative}, or incorporating second order information into the update of the covariance matrix \citep{auger2004ls}.

Another class of algorithms using rank-based fitness shaping are those from the family of Natural Evolution Strategies (NES) \citep{wierstra2014natural}.
Instead of updating the search distribution parameters with heuristics, NES follow a sample-based search gradient based on the same objective as MORE. Yet, due to the use of rankings, the connection to the natural gradient of the original expected reward objective is lost.
We describe the relation to MORE in more detail in the next section.

Contrary to these algorithms, MORE does not discard poorly performing candidates in its optimization process and only applies common data pre-processing techniques such as standardization before feeding them to the learning algorithm.
There also exist variations of the CMA-ES using surrogate models such as \citep{loshchilov2012self} or \citep{hansen2019global}.
Whereas these algorithms use the surrogate to generate approximate function values, which are then again used to generate a ranking, MORE directly uses the model parameters to update the search distribution parameters.

\subsection{Natural Gradients for Black-box Optimization}
The natural gradient is often used in optimization as it has shown to be more effective when a parameter space has a certain underlying structure \citep{amari1998natural}.
Important algorithms belong to the family of Natural Evolution Strategies (NES) \citep{wierstra2014natural}.
They use a Taylor approximation of the KL-divergence between subsequent updates to estimate a search gradient in the direction of the natural gradient.
Instead of information theoretic trust-regions on the update as used in MORE, they use either fixed \citep{sun2009efficient, glasmachers2010exponential} or heuristically updated learning rates \citep{wierstra2014natural}.
NES also uses ranking-based fitness shaping. The ranking is required to improve the robustness of the algorithm, yet, it also changes the objective and the search direction does not correspond to the natural gradient anymore. In contrast, MORE is inherently robust due to the used trust regions and therefore, does not require a ranking-based transformation of the rewards. 
The ROCK$\ast$ algorithm presented in \citep{hwangbo2014rock} uses the natural gradient on a global approximation of the objective generated with kernel regression.

\subsection{Trust-region Algorithms}
Updates bounded by a trust-region are a common approach in the reinforcement learning literature.
As MORE is based on the policy search objective, we also review some of the recent work in this domain.
A closely related method to MORE is the Relative Entropy Policy Search (REPS) for step-based reinforcement learning \citep{peters2010relative} and its episodic formulation (albeit derived in a contextual setting) in \citep{kupcsik2013data}.
Both use samples to approximate the objective and the trust regions which does not guarantee that the KL-trust region is enforced for the new parametric policy in practice. 
Layered direct policy search (LaDIPS) \citep{end2017layered} extends MORE to the contextual setting.

The ideas have then been taken towards reinforcement learning with neural network function approximation where they spawned state of the art algorithms such as TRPO \citep{schulman2015trust}.
The usage of natural gradients can be derived by using a second order approximation of the KL-divergence, resulting in the Fisher information matrix used in NES.
Similar sample based approximations of the trust region are also employed by recent deep reinforcement learning techniques such as Maximum Posteriori Policy Optimization (MPO) \citep{abdolmaleki2018maximum}.
Compatible function approximation for deep reinforcement learning has been explored in \citep{pajarinen2019compatible}.
Trust-regions can also be enforced in the policy function directly by using differentiable trust-region layers as introduced in \citep{otto2021differentiable}.

\subsection{Other Black-Box Optimization Approaches}
There exists a wide variety of stochastic search algorithm classes for black-box optimization, each with their own benefits and drawbacks.
Classic algorithms such as Nelder-Mead \citep{nelder1965simplex} use a simplex to find the minimum of a function.
Bayesian optimization techniques such as Gaussian processes, for example, aim at finding global optima in low dimensional problem domains \citep{osborne2009gaussian}. However, they suffer from high computation time and scale poorly with problem dimensionality and data points.
Other directions are genetic algorithms \citep{holland1992genetic} which are easy to implement but suffer from the need for good heuristics and random search \citep{zabinsky2010random, price1983global}. While both approaches can yield good results, they are not computationally efficient.

\subsection{Broader Scope}
The MORE algorithm finds application in a variety of fields, such as variational inference \citep{arenz2018efficient} or density estimation \citep{becker2019expected}. 
In the context of trajectory optimization, ideas from the MORE algorithm are explored in the MOTO algorithm \citep{akrour2018model}.

\section{Model-Based Relative Entropy Stochastic Search}
The goal of the Model-Based Relative Entropy Stochastic Search (MORE) algorithm \citep{NIPS2015_36ac8e55} is to find a parameter $\bm x \in \mathbb{R}^\dimF$ that maximizes \footnote{Equivalently, a cost can be minimized by maximizing the negative objective function.} a possibly noisy objective function $f(\bm x)$.
We consider the black-box scenario where we only access to potentially noisy sample evaluations, i.e.,  no expected function values and no gradients of the objective function are available.
Stochastic search can be seen as a special form of policy search \cite{deisenroth2013survey}, where the search distribution $\pi(\bm x)$ to be learned is a distribution of the problem's variables $\bm x$.
In robotics, for example, the objective function describes the outcome of a task and we are often interested in a distribution over controller parameters that performs well in expectation.
Thus, we frame stochastic search as maximizing the expected fitness function, i.e., 
\begin{equation}
    \max_{\pi} \mathbb{E}_{\bm x \sim \pi} [f(\bm{x})],
    \label{eq:ps_obj}
\end{equation}
which is a well known objective in policy search \cite{deisenroth2013survey}.

\subsection{Trust-Region Objective}
\label{sec:more}
Let the current search distribution be $\pi_t(\bm{x}) = \mathcal{N}(\bm{x} \mid \bm{\mu}_{\pi_t}, \bm{\Sigma}_{\pi_t})$ and the new search distribution be $\pi(\bm{x}) = \mathcal{N}(\bm{x} \mid \bm{\mu}_\pi, \bm{\Sigma}_{\pi})$ \footnote{Where unambiguous, we leave out the subscript $\pi$ for easier notation.}.
The optimization problem is formally given by
\begin{maxi}|l|
	{\pi}{\int_{\bm{x}} \pi(\bm{x}) f(\bm{x}) \dx}
	{}{}
	\addConstraint{\KLdiv{\pi(\bm{x})}{\pi_t(\bm{x})}}{\leq \epsilon}
	\addConstraint{H\left(\pi(\bm{x})\right)}{\geq \beta}
	\addConstraint{\int_{\bm{x}} \pi(\bm{x}) \dx}{= 1}
	\label{eq:more_const_opt}
\end{maxi}
where $\epsilon$ and $\beta$ are hyper-parameters controlling the exploration-exploitation trade-off and $H(\pi(\bm{x}))$ denotes the Shannon entropy of the search distribution.

MORE optimizes the objective from Equation \ref{eq:more_const_opt} by substituting the expected objective value with a learned quadratic approximation
\begin{equation}
    f(\bm{x}) \approx \hat{f}(\bm{x}) = - 1 / 2 \, \bm{x}^\mathrm{T} \bm{A} \bm{x} + \bm{x}^\mathrm{T} \bm{a} + a_0 \label{eq:quad_model}.
\end{equation}
Using a quadratic surrogate and a Gaussian search distribution, the optimal solution in each iteration would be to set the mean of the search distribution to the optimum of the surrogate and collapse the search distribution to a point estimate which would prevent it from further exploration. 
To control the exploration-exploitation trade-off, additional constraints on the updated search distribution need to be introduced.
First, the KL-divergence between the current search distribution and the solution of the optimization problem is upper bounded which ensures that the mean and covariance only slowly change and the approximate model is not over-exploited.
Furthermore, an additional lower bound on the entropy is introduced to prevent premature convergence.

\subsubsection{Closed Form Updates}
The optimization problem can be solved in closed form using the method of Lagrangian multipliers.
The dual function is given by 
\begin{align*}
	g(\eta, \omega) &= \eta \epsilon - \omega \beta + (\eta + \omega) \log \left( \int \pi_t(\bm{x})^{\frac{\eta}{\eta + \omega}} \exp \left(\frac{f(\bm{x})}{\eta + \omega}\right) \dx \right)
\end{align*}
with Lagrangian multipliers $\eta$ and $\omega$.
The new search distribution is then given in terms of the Lagrangian multipliers as
\begin{align*}
	\pi(\bm{x}) \propto \pi_t(\bm{x})^{\frac{\eta}{\eta + \omega}}\exp \left(\frac{f(\bm{x})}{\eta + \omega}\right).
\end{align*}

\subsubsection{Analytic Solution}
The use of a quadratic model is beneficial for two reasons.
First, it is expressive enough to solve a wide range of problems while being computationally easy to obtain.
Second, quadratic features are the compatible features of the Gaussian distribution  which make the integrals tractable.
As shown in Section \ref{sec:nat_grad}, the use of quadratic features also allows us to obtain exact natural gradient updates. 
The optimization problem in terms of the natural parameters $\bm m = \bm \Sigma^{-1} \bm \mu$ and $\bm \Lambda = \bm \Sigma^{-1}$ is solved by minimizing the Lagrangian dual function, which is given as
\begin{align*}
	g(\eta, \omega) =& \eta \epsilon - \omega \beta + \frac{1}{2}\big( \omega k \log(2 \pi) - \eta (\log\vert\bm {\Lambda}^{-1}_{t}\vert + \vecT{\bm{m}_{t}}\bm{\Lambda}_{t}^{-1}\bm{m}_{t}) \\
	&+ (\eta + \omega) (\log\vert\bm{\Lambda}^{-1}\vert + \vecT{\bm{m}}\bm{\Lambda}^{-1}\bm{m})\big),
\end{align*}
and the optimization problem becomes
\begin{mini*}|l|
	{\eta, \omega}{g(\eta, \omega)},
	{}{}
	\addConstraint{\eta}{\ge 0}
	\addConstraint{\omega}{\ge 0}.
\end{mini*}
The update rules for the new search distribution based on the Lagrangian multipliers are then given by
\begin{align*}
	\bm{\Lambda} &= \frac{\eta \bm{\Lambda}_{t} + \bm{A}}{\eta + \omega}, &
	\bm{m} &= \frac{\eta \bm{m}_{t} + \bm{a}}{\eta + \omega}. 
\end{align*}
Mean and covariance can be recovered by the inverse transformations and are given by $\bm \mu = \bm \Lambda^{-1} \bm m$ and $\bm \Sigma = \bm \Lambda^{-1}$.
We can now see that the new search distribution's parameters are an interpolation between the natural parameters of the old distribution and the parameters of the quadratic model from Equation \eqref{eq:quad_model}.

\subsection{Model Learning}

The parameters of the quadratic model can be learned using linear regression with quadratic features.
The original MORE algorithm first uses a probabilistic dimensionality reduction technique to project the problem parameters into a lower dimensional space.
Afterwards, weighted Bayesian linear regression is used to solve for the model parameters in the reduced space.
Lastly, the model parameters are projected back into the original space.
This approach has several drawbacks, as it introduces additional hyper-parameters to the algorithm and involves a sample-based approach to integrate out the projection matrix which is very costly in terms of computation time.
We will later on show that by using appropriate data pre-processing techniques, standard linear least squares is better suited to efficiently fit the quadratic models and also allows a direct connection of MORE to natural gradients.

\subsection{Entropy Control}
The entropy of the search distribution can be controlled with parameter $\beta$.
The bound $\beta$ is chosen such that entropy of the search distribution $H\left(\pi(\bm{x})\right)$ decreases by a certain percentage until a minimum entropy $H^0$ is reached, i.e. $\beta = \gamma (H\left(\pi_t(\bm{x})\right) - H^0) + H^0$.
Alternatively, a simple alternative is to linearly decrease $\beta$ in each iteration until the lower bound $H^0$ is reached, i.e. $\beta = \max(H\left(\pi_t(\bm{x})\right) - \delta, H^0)$.

\subsection{Relation to Natural Gradient}
\label{sec:nat_grad}
MORE can also be derived from natural gradients using compatible function approximation. In difference to other methods such as CMA-ES \cite{akimoto2010bidirectional} or NES, where a relation to natural gradients has also been established, the natural gradient update of MORE is exact since the estimation of the quadratic model is unbiased in expectation and no ranking based reward transformation is used.
On the other hand, other algorithms, involve approximations such as the use of rankings or weighted maximum likelihood estimates.

The natural gradient is given by scaling the "vanilla" gradient with the inverse of the Fisher information matrix (FIM) $\bm F$
\begin{align*}
	\bm{g}_\text{NG} = \alpha \bm F^{-1} \nabla_\theta J
\end{align*}
where $\alpha$ is a learning rate, $J$ is the objective and $\theta$ are the parameters of the search distribution.
In the stochastic search case, the FIM is given by 
\begin{align*}
	\bm F = \mathbb{E}_{\bm{x} \sim \pi}[\nabla_\theta \log \pi(\bm{x})\vecT{\nabla_\theta \log \pi(\bm{x})}]
\end{align*}
which we can approximate with samples. %
With $\nabla_\theta J = \mathbb{E}_{\bm{x} \sim \pi} [\nabla_\theta \log \pi(\bm x) f(\bm x)]$,
the natural gradient is then in approximation
\begin{align}
	\bm{g}_\text{NG} \approx &\alpha \left(\sum_i \nabla_\theta \log \pi(\bm{x}_i)\vecT{\nabla_\theta \log \pi(\bm{x}_i)}\right)^{-1} \sum_i \nabla_\theta \log \pi(\bm x) f(\bm x) \nonumber \\
	&= \left(\vecT{\bm{\Phi}}\bm{\Phi}\right)^{-1}\vecT{\bm{\Phi}}\bm{y} \label{eq:NG}
\end{align}
with
\begin{align*}
	\bm{\Phi} = \begin{bmatrix}
		\nabla_\theta \log \pi(\bm{x}_1) \\
		\vdots \\
		\nabla_\theta \log \pi(\bm{x}_N)
	\end{bmatrix} , \quad \bm{y} = \begin{bmatrix}
	y_1 \\
	\vdots \\
	y_N
\end{bmatrix}.
\end{align*}
Ignoring the learning rate $\alpha$, Equation \ref{eq:NG} resembles the least squares solution to a linear regression problem which is given by
\begin{align*}
	\bm \beta^\ast = \argmin_{\bm \beta} \, \sum_i{(\vecT{\nabla_\theta \log \pi(\bm{x})} \bm \beta - y_i)}^2
\end{align*}
using the function approximator $\vecT{\nabla_\theta \log \pi(\bm{x})} \bm \beta$ where the features are given by the gradient of the log search distribution with respect to the natural parameters of $\pi$. These features are often referred to as compatible features of the policy \citep{sutton1999policy, pajarinen2019compatible}. For a Gaussian distribution with natural parameters $\bm m = \bm \Sigma^{-1} \bm \mu$ and $\bm \Lambda = \bm \Sigma^{-1}$, these features are given by 
\begin{align*}
	\nabla_{\bm m} \log \pi(\bm{x}) &= \vecT{\bm x} + c_1, \\
	\nabla_{\bm \Lambda} \log \pi(\bm{x}) &= - \frac{1}{2} \bm x \vecT{\bm x} + c_2.
\end{align*}
Thus, the compatible features are given by
\begin{align*}
	\phi(\bm x) &= [{\nabla_{\bm m} \log \pi(\bm{x})}, \vect({\nabla_{\bm \Lambda} \log \pi(\bm{x})})] \\
	     &= \begin{bmatrix}
	     	1, \bm x, - \frac{1}{2} \vect( \bm x \vecT{\bm x})
	     \end{bmatrix}.
\end{align*}
Assuming an entropy bound $\beta \rightarrow -\infty$, the update equations in MORE are simply given by
\begin{align*}
	\bm m = \bm m_t + \eta^{-1} \bm a, \quad \bm \Lambda = \bm \Lambda_t + \eta^{-1} \bm A.
\end{align*}
Consequently, if the model is obtained by a standard least squares fit, 
the MORE algorithm is performing an update in the  natural gradient direction where the learning rate is specified indirectly by the trust region $\epsilon$.

\section{Improving the MORE Algorithm}
\label{sec:ca_more}
In its original formulation, MORE has several drawbacks as already pointed out earlier.
In this section, we will introduce a new version of MORE that aims at improving convergence speed and simplifying the model learning approach.
We achieve this by (a) disentangling the trust regions for mean and covariance, (b) an adaptive entropy control mechanism, (c) a simplified but improved model learning approach based on standard linear least squares.
We also propose a robust fitting of the surrogate to stabilize the model fitting process.
We will refer to the new algorithm as \textbf{C}oordinate-\textbf{A}scent MORE with \textbf{S}tep Size Adaptation or CAS-MORE for short.

\subsection{Disentangled Trust Regions}
The trust-region radius $\epsilon$ controls the change of the distributions between subsequent updates and thus has an influence on the speed of convergence. 
While a large trust region should encourage larger steps of the mean, we observed in our experiments  that the result is mainly a large change in the covariance matrix leading to instabilities and divergence of the algorithm, especially on problems where it is difficult to estimate the quadratic model.

Therefore, limiting the KL-divergence between the policies in subsequent iterations is a sub-optimal choice as we have no direct control whether the mean or covariance of the search distribution is updated more aggressively.
To alleviate this problem, we propose to decouple the mean and covariance update by employing a block coordinate ascent strategy for the mean and the covariance matrix which allows for setting different bounds on each of the components.

The optimization problems we will be solving are
\begin{maxi}|l|
	{\bm{\mu}}{\left.\int_{\bm{x}} \pi(\bm{x}) \hat{f}(\bm{x}) \dx\right\rvert_{\bm{\Sigma} = \bm{\Sigma}_t}}
	{}{}
	\addConstraint{\left.\KLdiv{\pi(\bm{x})}{\pi_t(\bm{x})}\right\rvert_{\bm{\Sigma} = \bm{\Sigma}_t}}{\leq \epsilon_\mu}
	 \label{eq:cas_more_mean}
\end{maxi}
for the mean and
\begin{maxi}|l|
	{\bm{\Sigma}}{\left.\int_{\bm{x}} \pi(\bm{x}) \hat{f}(\bm{x}) \dx\right\rvert_{\bm{\mu} = \bm{\mu}_t}}
	{}{}
	\addConstraint{\left.\KLdiv{\pi(\bm{x})}{\pi_t(\bm{x})}\right\rvert_{\bm{\mu} = \bm{\mu}_t}}{\leq \epsilon_\Sigma}
	\label{eq:cas_more_cov}
\end{maxi}

for the covariance in each iteration.
By choosing a small bound $\epsilon_\Sigma$, we can drop the entropy constraint from the optimization as the entropy is not decreasing as quickly.

\subsubsection{Updating the Mean}
We start updating the mean by setting $\bm{\Sigma} = \bm{\Sigma}_t$ and introducing a bound $\epsilon_\mu$ to limit the change of the mean displacement.
The dual function to the problem in Equation \ref{eq:cas_more_mean} is given by
\begin{align*}
    g_\mu(\lambda) &= \lambda \epsilon_\mu + \frac{1}{2} \Big(\bm{m}_\mu(\lambda)^\mathrm{T} \bm{M}_\mu(\lambda)^{-1} \bm{m}_\mu(\lambda) - \lambda \bm{m}_{t}^\mathrm{T} \bm{M}_t^{-1} \bm{m}_t\Big)
\end{align*}
where $\lambda$ is a Lagrangian multiplier and 
\begin{align*}
    \bm{M}_\mu(\lambda) &= \lambda \bm{\Sigma}_{t}^{-1} + \bm{A}, & \quad
    \bm{m}_\mu(\lambda) &= \lambda \bm{\Sigma}_{t}^{-1} \bm{\mu}_{t} + \bm{a}.
\end{align*}
We now solve the dual problem
\begin{mini*}|l|
	{\lambda}{g_\mu(\lambda)}
	{}{}
	\addConstraint{\lambda}{> 0}
\end{mini*}
and obtain the new mean as
\begin{align*}
\bm{\mu}^\ast &= \bm{M}_\mu(\lambda^\ast)^{-1} \bm{m}_\mu(\lambda^\ast) %
\end{align*}
where $\lambda^\ast$ is the solution of the optimization problem which we find using a non-linear optimization algorithm. \footnote{We used the implementation of L-BFGS-B \citep{johnson2014nlopt} provided by the python package NLOpt \citep{johnson2014nlopt}}.

\subsubsection{Updating the Covariance Matrix}
Next, we set $\bm{\mu} = \bm{\mu}_t$ and introduce a bound $\epsilon_\Sigma$ to constrain the change of the covariance.
The dual function for the problem in Equation \ref{eq:cas_more_cov} is given by
\begin{align*}
    g_\Sigma(\nu) &= \nu \epsilon_\Sigma + \frac{1}{2} \nu \Big(\log \vert \bm{\Lambda}(\nu)^{-1}\vert - \log \vert \bm{\Lambda}_{t}^{-1} \vert \Big)
\end{align*}
with
\begin{align*}
    \bm{\Lambda}(\nu) = \frac{\nu \bm{\Sigma}_{t}^{-1} + \bm{A}}{\nu}
\end{align*}
and we need to solve the following optimization problem
\begin{mini*}|l|[0]
	{\nu, \omega}{g_\Sigma(\nu)}
	{}{}
	\addConstraint{\nu}{> 0}
\end{mini*}
where $\nu$ is again a Lagrangian multiplier.
The optimal solution $\bm{\Sigma}^\ast$ in terms of the solution $\nu^\ast$ can be found analogously and is given by
\begin{align*}
\bm{\Sigma}^\ast &= \bm{\Lambda}(\nu^\ast)^{-1}. %
\end{align*}

\subsection{Entropy Control}
\label{sec:csa}

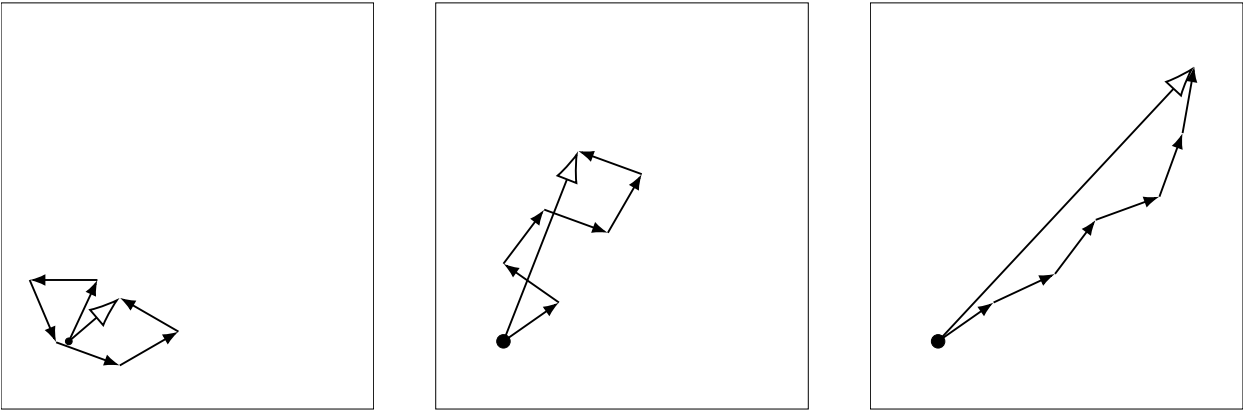
\begin{figure}
    \centering
    \begin{subfigure}[b]{0.3\textwidth}
        \centering
        \begin{tikzpicture}[every node/.style={color=black}]
        \useasboundingbox (-1,-1) rectangle (4.5,5.);
    \draw (-1,-1) rectangle (4.5,5.);
    \path (0,0) coordinate (origin);
    \path (origin) ++(65:1cm) coordinate (P0);
    \path (P0) ++(180:1cm) coordinate (P1);
    \path (P1) ++(293:1cm) coordinate (P2);
    \path (P2) ++(340:1cm) coordinate (P3);
    \path (P3) ++(30:1cm) coordinate (P4);
    \path (P4) ++(150:1cm) coordinate (P5);
    
    \node at (origin) [circle,draw,fill, inner sep=0.1em] {};
    \draw[thick, -Latex] (origin) -- (P0);
    \draw[thick, -Latex] (P0) -- (P1);
    \draw[thick, -Latex] (P1) -- (P2);
    \draw[thick, -Latex] (P2) -- (P3);
    \draw[thick, -Latex] (P3) -- (P4);
    \draw[thick, -Latex] (P4) -- (P5);
    
    \draw[thick, -{Latex[open,scale=2]}] (origin) -- (P5);
\end{tikzpicture} \label{fig:csa_short}
    \end{subfigure}
    \hfill
    \begin{subfigure}[b]{0.3\textwidth}
        \centering
        \begin{tikzpicture}[every node/.style={color=black}]
        \useasboundingbox (-1,-1) rectangle (4.5,5.);
    \draw (-1,-1) rectangle (4.5,5.);
    \path (0,0) coordinate (origin);
    \path (origin) ++(35:1cm) coordinate (P0);
    \path (P0) ++(145:1cm) coordinate (P1);
    \path (P1) ++(53:1cm) coordinate (P2);
    \path (P2) ++(-20:1cm) coordinate (P3);
    \path (P3) ++(60:1cm) coordinate (P4);
    \path (P4) ++(160:1cm) coordinate (P5);
    
    \node at (origin) [circle,draw,fill, inner sep=0.2em] {};
    \draw[thick, -Latex] (origin) -- (P0);
    \draw[thick, -Latex] (P0) -- (P1);
    \draw[thick, -Latex] (P1) -- (P2);
    \draw[thick, -Latex] (P2) -- (P3);
    \draw[thick, -Latex] (P3) -- (P4);
    \draw[thick, -Latex] (P4) -- (P5);
    
    \draw[thick, -{Latex[open,scale=2]}] (origin) -- (P5);
\end{tikzpicture} \label{fig:csa_good}
    \end{subfigure}
    \hfill
    \begin{subfigure}[b]{0.3\textwidth}
        \centering
        \begin{tikzpicture}[every node/.style={color=black}]
    \useasboundingbox (-1,-1) rectangle (4.5,5.);
    \draw (-1,-1) rectangle (4.5,5.);
    \path (0,0) coordinate (origin);
    \path (origin) ++(35:1cm) coordinate (P0);
    \path (P0) ++(25:1cm) coordinate (P1);
    \path (P1) ++(53:1cm) coordinate (P2);
    \path (P2) ++(20:1cm) coordinate (P3);
    \path (P3) ++(70:1cm) coordinate (P4);
    \path (P4) ++(80:1cm) coordinate (P5);
    
    \node at (origin) [circle,draw,fill, inner sep=0.2em] {};
    \draw[thick, -Latex] (origin) -- (P0);
    \draw[thick, -Latex] (P0) -- (P1);
    \draw[thick, -Latex] (P1) -- (P2);
    \draw[thick, -Latex] (P2) -- (P3);
    \draw[thick, -Latex] (P3) -- (P4);
    \draw[thick, -Latex] (P4) -- (P5);
    
    \draw[thick, -{Latex[open,scale=2]}] (origin) -- (P5);
\end{tikzpicture} \label{fig:csa_long}
    \end{subfigure}
	\caption{This figure illustrates the concept of the evolution path in a 2D example and is adapted from \cite{hansen2016cma}. In each figure, arrows with black heads correspond to an update of the mean. The evolution path is a smoothed sum over subsequent mean updates and depicted with an open head. In the left plot, the mean updates show no clear search direction and, as a result, the evolution path is short. The right plot shows the opposite where heavily correlated mean updates lead to a long evolution path. The center plot shows the desired case, where subsequent mean updates are uncorrelated.}\label{fig:csa}
\end{figure}

The standard MORE algorithm is only able to decrease the entropy of the search distribution in each iteration. This might result in slow convergence in particularly if the search distribution is initialized with too small variances\footnote{This is often necessary in case a large initial exploration is generating samples that are very costly or dangerous to evaluate, e.g., in robotics.}. 
On the other hand, a too large covariance can also lead to slow convergence as it takes a long time to settle in on a solution and simply increasing the bound on the covariance can lead to unstable updates.

To solve these issues, we take inspiration from CMA-ES which already makes use of an entropy control mechanism in form of the step-size update \citep{hansen2016cma}.
The crucial property underlying this update is the so-called evolution path which is a smoothed sum over previous mean updates.
The idea of step-size adaptation is the following. Whenever we take steps in the same direction, we could essentially take a single, larger step, while dithering around a constant location indicates no clear search direction. 
In the first case, a larger entropy would allow for bigger steps, in the second case, we need to decrease entropy.
We illustrate the concept in Figure \ref{fig:csa}.
We can achieve this by scaling the covariance after the optimization with a scalar factor 
\begin{align*}
\sigma_{t+1} &= \exp\left(-\frac{\delta_{t+1}}{\dimF}\right)
\end{align*} 
after the optimization steps where $\dimF$ is the problem dimensionality and $\delta_{t+1}$ is the desired change between the entropy in iteration $t$ and the next iteration $t+1$ of the algorithm.
The question is now how to determine a suitable value for $\delta_{t+1}$.

The step size update we propose is inspired by the step size control mechanism used in CMA-ES called cumulative step size adaptation (CSA).
We also use the evolution path $\bm{p}$ which is a vector tracking previous mean updates and compare its length to a hypothetical length of the evolution path resulting from uncorrelated mean updates.
Therefore, the evolution path update needs to be constructed in a way that we can derive a computable desired length from it.
Inspired by CMA-ES, we initialize the evolution path $\bm{p}_0$ to zero and update it as follows
\begin{align*}
		\bm{p}_{t+1} = (1 - c_\sigma) \bm{p}_t + \sqrt{\frac{c_\sigma (2 - c_\sigma)}{2 \epsilon_\mu}} \bm{\Sigma}_t^{-\frac{1}{2}}(\bm{\mu}_{t+1} - \bm{\mu}_t)
	\end{align*}
where $c_\sigma < 1$ is a hyper parameter.
The factors are chosen analogously to CMA-ES such that $(1 - c_\sigma)^2 + \sqrt{c_\sigma (2 - c_\sigma)}^2 = 1$. 
While multiplying the shift in means with the inverse square root of the covariance matrix in CMA-ES makes the evolution path roughly zero mean and unit variance distributed, we know its exact length due to the constraint on the mean update in Equation \ref{eq:cas_more_mean} as long as the mean update is at its bound.
The vector $\bm{\Sigma}_t^{-\frac{1}{2}}(\bm{\mu}_{t+1} - \bm{\mu}_t)$ has in that case length $\sqrt{2 \epsilon_\mu}$ and by scaling it with $(\sqrt{2 \epsilon_\mu})^{-1}$ the resulting length is 1. 
We set
\begin{align*}
\delta_{t+1} = \alpha \left(1 - \frac{\norm{\bm{p}_{t+1}}}{\norm{ \bm{p}_{t+1}^{\text{des}}}} \right),
\end{align*}
where $\norm{ \bm{p}_{t+1}^{\text{des}}} $ is a desired length of the evolution path given by the length of a evolution path resulting from fully uncorrelated updates and $\alpha$ influences the magnitude of entropy change which in our experiments was set to 1.
This rule for determining the change in entropy is almost identical to the one in CMA-ES.
If the norm of the current evolution path $\bm{p}_{t+1}$ is smaller than $\bm{p}_{t+1}^{\text{des}}$, i.e., the mean updates were dithering around a constant location, we reduce entropy (i.e. $\delta_{t+1} > 0)$ while entropy is increased if the length of the evolution path is longer than desired.

In order to determine the length of an uncorrelated path $\norm{ \bm{p}_{t+1}^{\text{des}}}$, we first look at the length of two vectors
\begin{align*}
\norm{\bm{a} + \bm{b}} = \sqrt{\norm{\bm{a}}^2 + \norm{\bm{b}}^2 + 2 \norm{\bm{a}} \norm{\bm{b}}\cos \theta},
\end{align*}
where $\theta$ is the angle between vectors $\bm{a}$ and $\bm{b}$.
If these two vectors are uncorrelated (i.e. $\theta = \pi/2$), the length becomes $\sqrt{\norm{\bm{a}}^2 + \norm{\bm{b}}^2}$. 
Under the assumption that the mean update is always at the bound, i.e. $(2 \epsilon_\mu \bm{\Sigma}_t)^{-\frac{1}{2}}(\bm{\mu}_{t+1} - \bm{\mu}_t) = 1$, the length of the evolution path $p_{t+1}$ given the previous length $p_{t}$ is then 
\begin{align*}
		p_{t+1} = \sqrt{(1 - c_\sigma)^2 p_t^2 + c_\sigma (2 - c_\sigma)}.
	\end{align*}
In practice, we set the desired angle $\theta = \frac{3 \pi}{8} $ and compute $\norm{\bm{p}_{t+1}^{\text{des}}}$ accordingly to account for unavoidable correlations between iterations due to sample reuse and constrained updates.
Finally, the adapted policy is
\begin{align*}
\pi_{t+1}^\sigma = \mathcal{N}(\bm{x} \mid \bm{\mu}_{t+1}, \sigma_{t+1}^2 \bm{\Sigma}_{t+1}).
\end{align*}

\subsection{Illustrative Example}
\begin{figure}
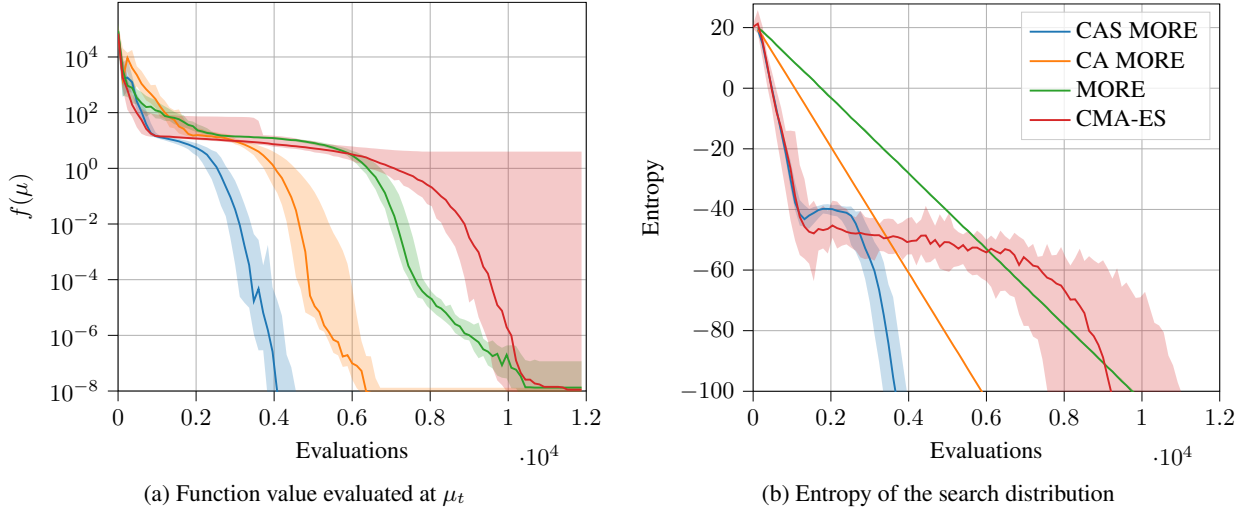

    \centering
    \begin{subfigure}[b]{0.49\textwidth}
        \centering
        \includestandalone[width=\linewidth]{figures/rosenbrock/rb_cas_ca_more_cma_dist_to_opt_median} 
        \caption{Function value evaluated at $\mu_t$}
        \label{fig:rb_more_comp_a}
    \end{subfigure}
    \hfill
    \begin{subfigure}[b]{0.49\textwidth}
        \centering
        \includestandalone[width=\linewidth]{figures/rosenbrock/rb_cas_ca_more_cma_entropy_median}
        \caption{Entropy of the search distribution}
        \label{fig:rb_more_comp_b}
    \end{subfigure}
    \caption{This figure shows the function value (left) and the entropy of the search distribution (right) over the course of optimization of a 15-dimensional Rosenbrock function using different variants of MORE. The original MORE is shown in green, while coordinate ascent versions of more are orange and blue where CA-MORE indicates a fixed entropy reduction schedule and CAS-MORE indicates the adaptive entropy schedule using the step-size adaptation. For comparison, CMA-ES is plotted in red.}
    \label{fig:rb_more_comp}
\end{figure}

We demonstrate the characteristics of CAS-MORE by comparing it to the original formulation of MORE 
and Coordinate Ascent MORE without adaptive entropy control (CA-MORE). %
We therefore use a 15 dimensional Rosenbrock function as defined in \cite{hansen:inria-00362633} and run each variant of MORE 20 times with different seeds.
The optimization is stopped once a target function value of \SI{1e-8} is reached, or \num{12000} function evaluations are exceeded.
Figure \ref{fig:rb_more_comp} shows median and 5\% / 95\% quantiles of the function value at the mean (Figure \ref{fig:rb_more_comp_a}) and the entropy of the search distribution (Figure \ref{fig:rb_more_comp_b}) over the optimization.
The hyper-parameters for the KL bounds (and entropy schedule for MORE) are chosen individually for each algorithm based on a grid search.
We observe that decoupling the mean and covariance update already significantly improves convergence speed as it allows for a higher learning rate for the mean and therefore a quicker entropy reduction.
Note, that reducing entropy even quicker leads to premature convergence while increasing $\epsilon$ for MORE leads to divergence of the algorithm.
This problem is alleviated by introducing the adaptive entropy schedule that first quickly reduces entropy, then plateaus, and, finally, quickly reduces entropy again, resulting in an accelerated optimization process.

\section{Model Learning}
\label{sec:model_learning}
In this section, we introduce a simple but effective approach of learning a quadratic model based on polynomial ridge regression using the method of least squares as replacement for the more complex Bayesian dimensionality reduction technique introduced in the original paper \citep{NIPS2015_36ac8e55}.
We design the model learning with two objectives in mind.
First, it should be data- and time-efficient to allow for quick execution of the algorithm.
Additionally, it should be robust towards outliers and samples from very low entropy regimes.
To this end, we apply a series of data pre-processing techniques, re-use old samples, and start learning a model with a lower complexity and increase it once sufficient data is available.

\subsection{Least Squares Model Fitting}
In each iteration, we generate a fixed number $K$ of samples $\bm x_i \sim \pi$, $i = 1, \dots, K$, evaluate them on the objective function to obtain $y_i = f(\bm x_i)$ and add the tuple $(\bm x_i, y_i)$ to a data-set $\mathcal{D} = \{(\bm x_q, y_q) \mid q = 1 \dots Q\}$ of length $Q$\footnote{Once the number of data-points $Q$ in $\mathcal{D}$ exceeds a maximum queue-size $Q_{\text{max}}$, we discard old samples in a first-in, first-out manner.}.
The goal of the model fitting process is to find the parameters $\bm{A}$, $\bm{a}$, and $a$ of a quadratic model of the form
\begin{align*}
	f(\bm{x}) \approx \hat{f}(\bm{x}) = - 1 / 2 \, \bm{x}^\mathrm{T} \bm{A} \bm{x} + \bm{x}^\mathrm{T} \bm{a} + a.
\end{align*}
These parameters can be found by solving a regularized least squares problem
\begin{align*}
	\min_{\bm \beta} \, \norm{(\bm y - \bm \Phi \bm \beta)}_2^2 + \lambda \norm{\bm{\beta}}_2^2,
\end{align*}
where $\bm \Phi$ is the design matrix whose rows are given by a feature transformation $\phi(\bm x)$, i.e $\bm \Phi = \vecT{[\phi(\bm x_1), \dots, \phi(\bm x_\queueSize)]}$, $\bm y$ is a vector containing the fitness evaluations $y_i = f(\bm x_i)$ and $\lambda$ is the regularization factor. 
The solution is given by the well known ridge regression estimator
\begin{align*}
	\hat{\bm{\beta}} = (\vecT{\bm \Phi} \bm \Phi + \lambda \bm I)^{-1} \vecT{\bm \Phi}\bm{y}.
\end{align*}

\subsection{Adaptive Model Complexity}
A full quadratic model has in the order of $\mathcal{O}(\dimF^2)$ parameters that need to be estimated.
Compared to model-free algorithms, this can be a disadvantage if we initially need to sample enough parameters for the first model to be built.
Instead, the complexity of the feature function $\phi(\bm x)$ will be gradually increased from a linear to a diagonal and, finally, a full quadratic model depending on the number of obtained samples. 
The simplest model is a linear model where 
\begin{align*}
	a &= \beta_1 & 
	\bm a &= \vecT{[\beta_2, \dots, \beta_{\dimF + 1}]} & 
	\bm A &= \bm 0.
\end{align*}
Next, we estimate a diagonal model where
\begin{align*}
	\bm{A} = -\text{diag}([\beta_{\dimF + 2}, \dots, \beta_{2 \dimF + 1}]).
\end{align*}
Once sufficiently many data-points are available, we estimate a full quadratic model where
\begin{align*}
	\bm A &= - (\bm L + \vecT{\bm L})
\end{align*}
with
\begin{align*}
	\bm L &= \begin{bsmallmatrix}
	\beta_{\dimF + 2} & 0 & \dots & \dots & 0 \\
	\beta_{\dimF + 3} & \beta_{\dimF + 4} & 0 & \dots & 0 \\
	\vdots & & \ddots & \\
	\beta_{\dimF (\dimF + 3) / 2 - 1} & \dots & \dots & \beta_{\dimF (\dimF + 3) / 2 - \dimF - 1} & 0 \\
	\beta_{\dimF (\dimF + 3) / 2 - \dimF + 2} & \dots & \dots & \dots & \beta_{\dimF (\dimF + 3) / 2 + 1}
	\end{bsmallmatrix}.
\end{align*}
The feature functions for the models are given by
\begin{align*}
\phi_{\text{lin}}(\bm x) &= \vecT{[1, x_1, x_2, \dots, x_\dimF]} \\
\phi_{\text{diag}}(\bm x) &= \vecT{[\vecT{\phi_{\text{lin}}(\bm x)}, x_1^2, x_2^2, \dots, x_\dimF^2]} \\
\phi_{\text{full}}(\bm x) &= \vecT{[\vecT{\phi_{\text{lin}}(\bm x)}, x_1^2, x_1 x_2, x_1 x_3, \dots, x_1 x_\dimF, x_2^2, x_2 x_3, \dots x_2 x_\dimF, x_3 x_4 , \dots , x_{\dimF}^2]}.
\end{align*}
We empirically found that we require at least 10\% more samples than the model has parameters, i.e. start with a linear model once $\vert\mathcal{D}\vert \geq 1.1 (1 + \dimF)$ and continue with a diagonal model once $\vert\mathcal{D}\vert \geq 1.1 (1 + 2 \dimF)$.
Finally, we switch to estimating a full quadratic model once $\vert\mathcal{D}\vert \geq 1.1 (1 + \dimF (\dimF + 3) / 2)$.

\subsection{Data Pre-Processing}
\begin{figure}
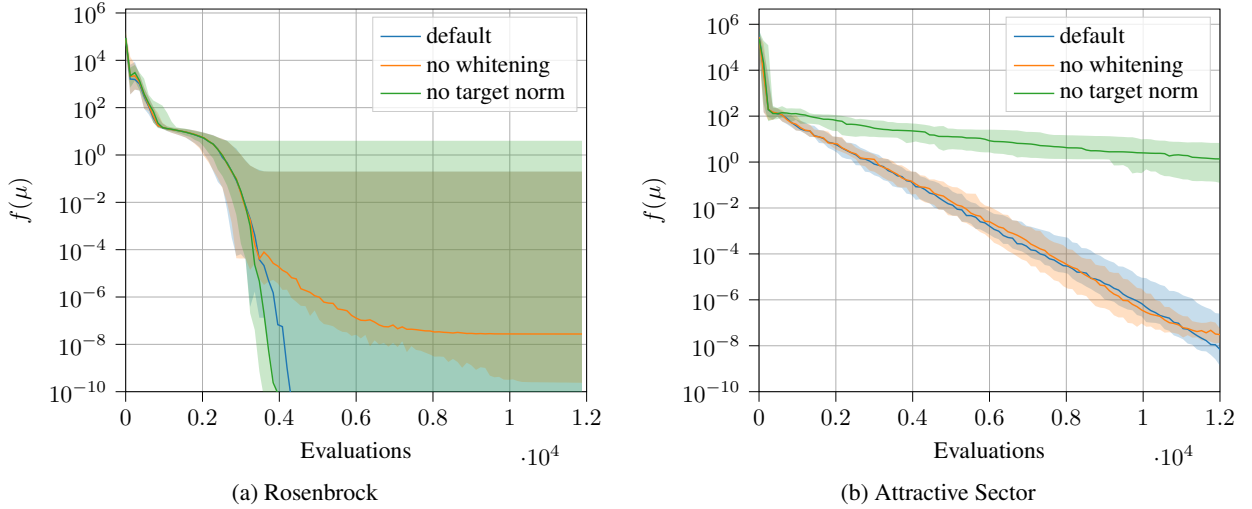

    \centering
    \begin{subfigure}[b]{0.49\textwidth}
        \centering
        \includestandalone[width=\linewidth]{figures/ablations/rb_abl_default_rb_abl_1_1_rb_abl_1_3_dist_to_opt_median} 
        \caption{Rosenbrock}
        \label{fig:rb_alation}
    \end{subfigure}
    \hfill
    \begin{subfigure}[b]{0.49\textwidth}
        \centering
        \includestandalone[mode=buildnew,width=\linewidth]{figures/ablations/as_abl_default_as_abl_1_1_as_abl_1_3_dist_to_opt_median}
        \caption{Attractive Sector}
        \label{fig:as_ablation}
    \end{subfigure}
    \caption{This figure shows the effects of data pre-processing on the optimization. We plot the function value over the course of optimization for a Rosenbrock (a) and Attractive Sector (b) function in 15 dimensions and turn off whitening and robust target normalization with clipping, respectively. We see that whitening becomes important for low entropy regimes near the optimum, while target normalization is especially important in case of outliers in the function values as can be seen for the Attractive Sector function.}
    \label{fig:more_model_ablations}
\end{figure}

Estimating the model parameters can be numerically difficult with function values spanning several orders of magnitude, outliers, and very low variances of input and output values near the optimum.
Therefore, we perform data pre-processing on the input values $\bm x$, as well as the design matrix $\bm \Phi$ and the target values $y$ before solving the least squares problem.
In particular, we perform whitening of the inputs $\bm x$, standardization of the design matrix $\bm \Phi$ and a special form of standardization of the target values $y$ and construct the normalized data set $\mathcal{D}_w$.
We demonstrate the effect of data pre-processing in Figure \ref{fig:more_model_ablations}.

\subsubsection{Data Whitening}
Before estimating $\hat{\bm{\beta}}$, we whiten the input data and obtain
\begin{align*}
	\bm x_w = \bar{\bm{C}}_{\mathcal{D}}^{-1} (\bm x - \bar{\bm x}_{\mathcal{D}}) \; \forall \; \bm x \in \mathcal{D}
\end{align*}
where $\bar{\bm x}_{\mathcal{D}}$ is the empirical mean and $\bar{\bm{C}}_{\mathcal{D}}$ is the Cholesky-factorization of the empirical covariance of the data-set.
The parameters $\hat{\bm \beta}_w$ learned with the whitened data-set to be transformed back into unwhitened parameters.
This transformation is given by
\begin{align*}
    \bm A &= \bar{\bm{C}}_{\mathcal{D}}^{-\mathrm{T}} \bm A_w \bar{\bm{C}}_{\mathcal{D}}^{-1}, \\
    \bm a &= \bm A \bar{\bm x}_{\mathcal{D}} + \bar{\bm{C}}_{\mathcal{D}}^{-\mathrm{T}} \bm a_w, \\
    a &= a_w + \vecT{\bar{\bm x}_{\mathcal{D}}} (\bm A \bar{\bm x}_{\mathcal{D}} - \bar{\bm{C}}_{\mathcal{D}}^{-\mathrm{T}} \bm a_w),
\end{align*}
where $\bm A_w$, $\bm a_w$ and $a_w$ are the model parameters in the whitened space.

\subsubsection{Target Normalization}
Normalization of function values is beneficial as it stabilizes the model learning process.
Depending on how well behaved the objective function is, we propose two target normalization methods.
The first one allows for exact natural gradient updates, while the second one is more robust towards outliers.

\paragraph{Standard Target Normalization}
The first target normalization method is based on a simple standardization
\begin{align*}
	y_w = (y - \bar{y}_{\mathcal{D}}) / s_{\mathcal{D}} \; \forall \; y \in \mathcal{D}
\end{align*}
of the target values where $\bar{y}_{\mathcal{D}}$ is the empirical mean and $s_{\mathcal{D}}$ is the empirical standard deviation of the target values in $\mathcal{D}$.

\begin{figure}
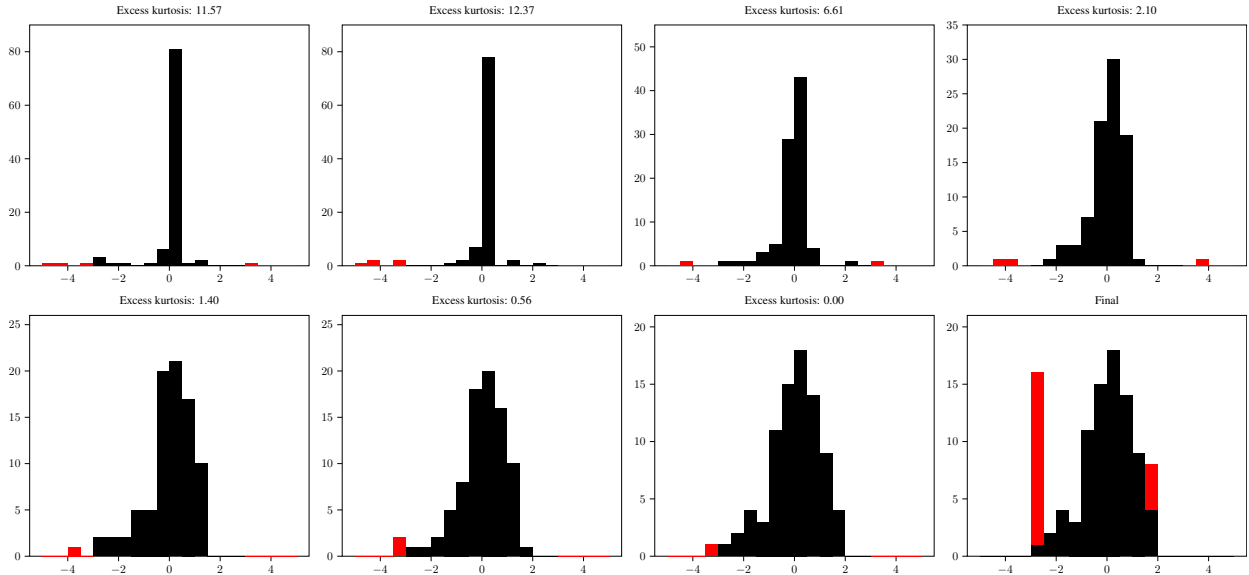

    \centering
    \includestandalone[mode=buildnew, width=\linewidth]{figures/norm_robust/norm_robust_all} \label{fig:norm_robust_0}
	\caption{This figure illustrates the robust target normalization scheme on 100 synthetic data points, generated by a reward function $y = -0.5 \vecT{\bm x} \bm x$. We sample $\bm x$ from a two dimensional multivariate normal distribution with zero mean and unit variance and pick 20 random $y$ values and add Gaussian noise with a standard deviation of 10. Plots from top left to bottom right show histograms of the data after standardization of the input in each level of recursion of the procedure. The excess kurtosis is measured on the black data points in the interval (-3, 3) and only these data points are recursively treated again until the excess kurtosis of the clipped data is below the threshold of $0.55$. Finally, the bottom right histogram shows the output with the red data points clipped to the minimum and maximum values of the remaining data points. Note, that in each plot the standardized data has zero mean and unit variance but the model quality suffers from the present outliers.}
	\label{fig:norm_robust}
\end{figure}

\paragraph{Robust Target Normalization}
Additionally, we propose an adaptive iterative normalization and clipping scheme based on the excess kurtosis of the target values resulting in a normalization process more robust towards outliers.
Given the data-set in a particular iteration, we first perform standardization like above, then treat all values outside the interval $[-v_{\text{clip}}, v_{\text{clip}}]$ as outliers.
Afterwards, we look at the excess kurtosis of the standardized targets in the interval $(-v_{\text{clip}}, v_{\text{clip}})$.
If it is high, the data is still compacted to a small interval while outliers at the borders negatively influence the least squares optimization.
In order to evenly spread the data, we repeat the procedure (normalization and clipping), until the excess kurtosis falls below a threshold or the standard deviation of the remaining values is close to 1.
After the procedure, we replace the negative and positive outliers with the minimum and maximum value of the remaining targets, respectively.
We illustrate the procedure in Figure \ref{fig:norm_robust}.
While this effectively changes the optimization objective, it is not as severe as replacing all function values with a ranking and, thus, the update direction still corresponds to an approximated natural gradient.
An exact quantification of the error is subject to future work.

\section{Experiments}
In this section, we evaluate the performance of CAS-MORE\footnote{An implementation of the algorithm can be found under \url{https://github.com/ALRhub/cas-more}} and compare it to the original formulation of MORE, as well as competitors such as CMA-ES \cite{hansen2016cma} and XNES \cite{wierstra2014natural}.
We first look at the performance in terms of its sample efficiency on a set of black-box optimization benchmark functions.
Additionally, we run our algorithm on a suite of different episodic reinforcement learning tasks from the domain of robotics where the fitness function evaluation is inherently noisy and we also care about the quality of the generated samples (e.g., the sample evaluations should not damage the robot) instead of plainly looking at the fitness evaluation at the mean of the search distribution.

\subsection{Black-box Optimization Benchmarks}
The COCO framework \citep{hansen2020cocoplat} provides continuous benchmark functions to compare the performance of optimizers on problems with different problem dimensions.
In these deterministic problems, it is only important to find a good point estimate $\bm{x}^\ast$.

\subsubsection{Experimental Setup}
We evaluate MORE on the 24 functions of the Black-box Optimization Benchmark (BBOB) \citep{hansen2010fun} suite in dimensions 2, 3, 5, 10, 20 and 40.
We run the experiments without explicit optimization of the algorithm's hyper parameters for individual functions.
Table \ref{tab:default_params} shows a set of default parameters which we found to robustly perform well on a wide variety of benchmark functions in terms of the problem dimensionality $\dimF$.
Additionally, we allow restarts of the algorithm with a larger population size whenever the optimization process fails (see \citep{auger2005restart}) until a budget of $\num{10000} \dimF$ function evaluations is exceeded or the final target of \SI{1e-8} is reached.
Since MORE maximizes, we multiply the function by -1 in order to minimize the objective.

\subsubsection{Black-box Optimization Benchmarks}
\begin{figure*}
	\begin{tabular}{c@{\hspace*{0.01\textwidth}}c@{\hspace*{0.01\textwidth}}c}
		{\small \sffamily separable fcts}\hspace{1cm} & {\small \sffamily moderate fcts}\hspace{1cm} & \hspace{-1cm}{\small \sffamily ill-conditioned fcts}\\
		\includegraphics[width=0.3\textwidth]{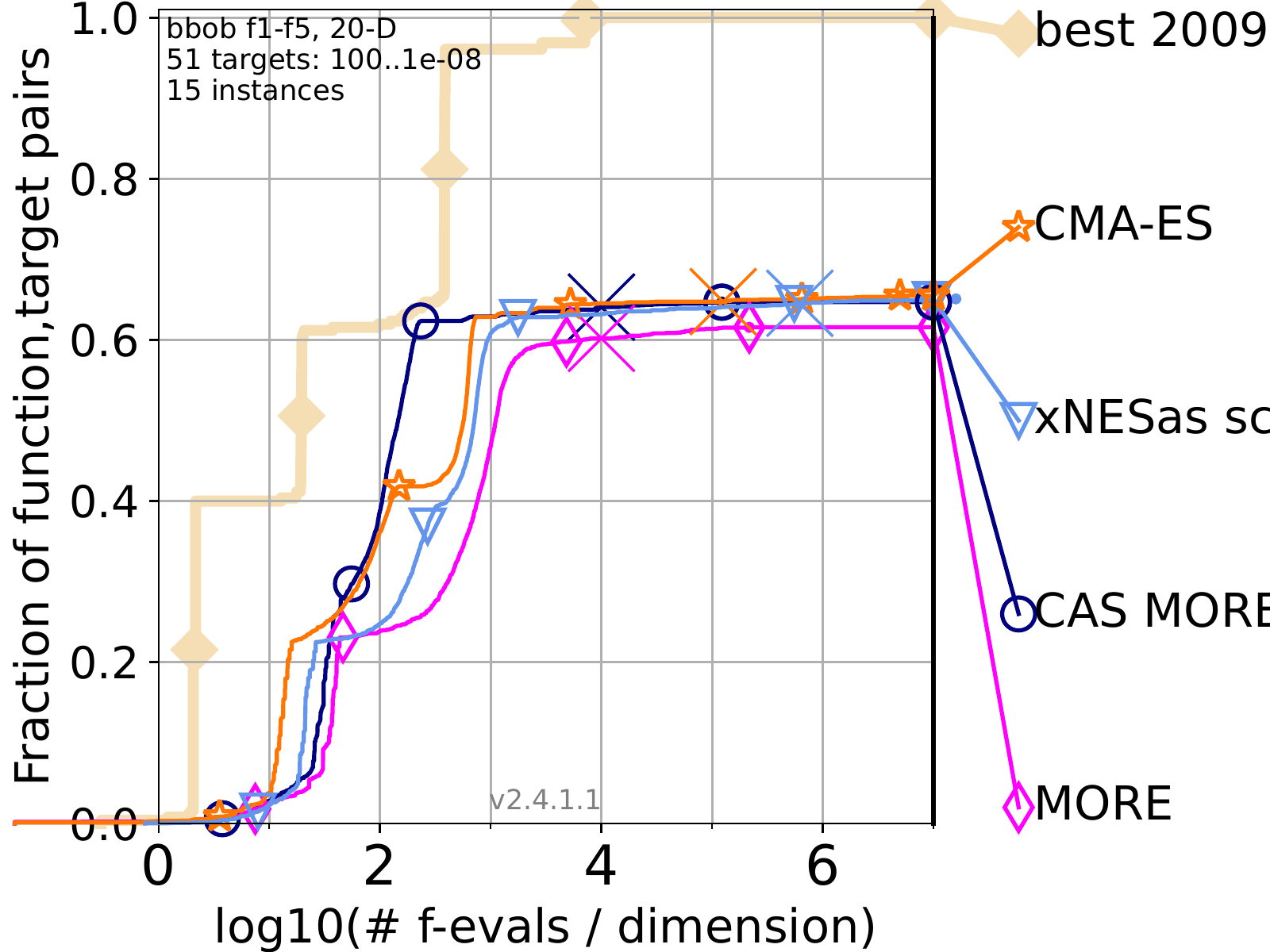}&
		\includegraphics[width=0.3\textwidth]{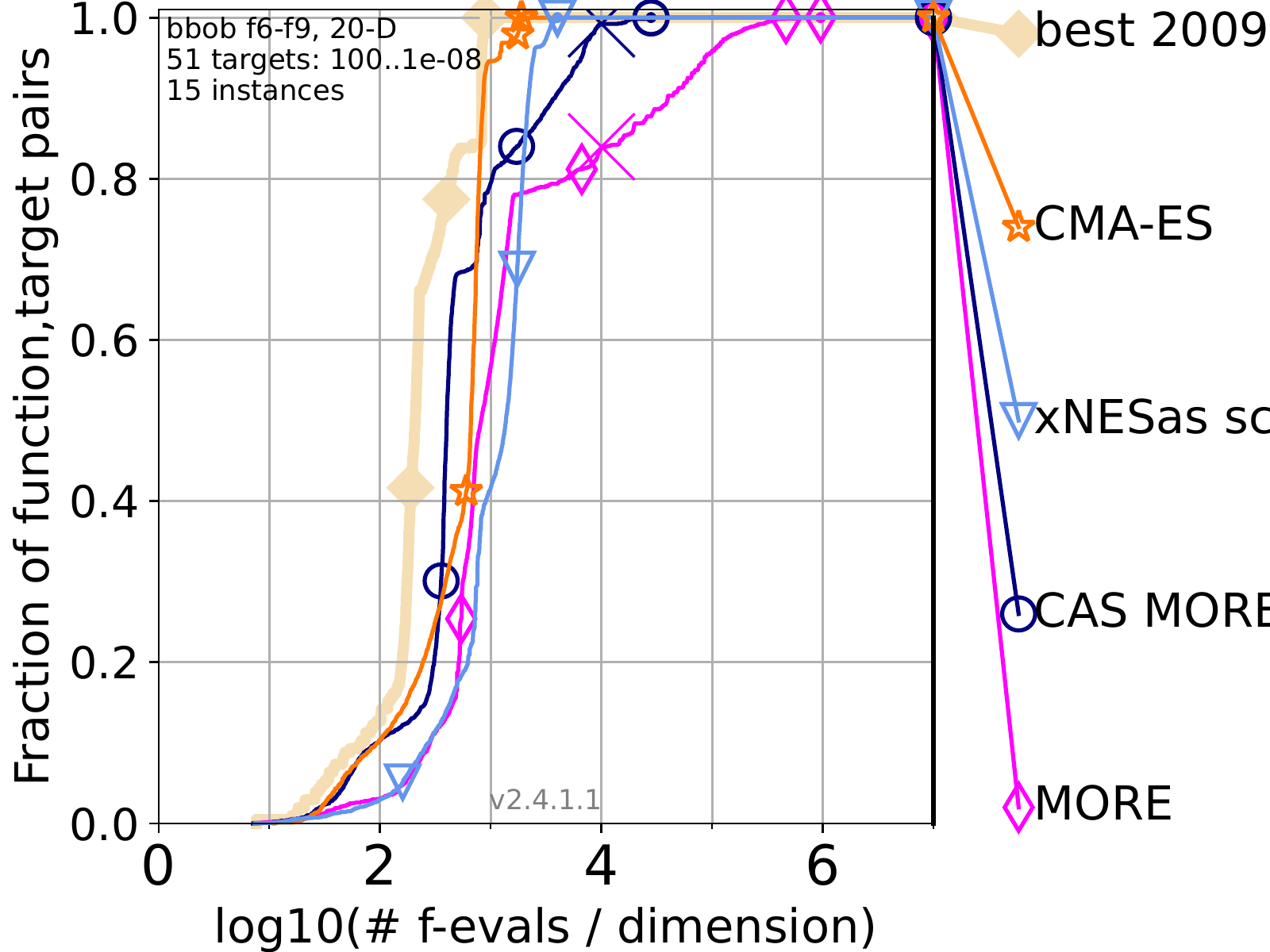}&
		\includegraphics[width=0.3\textwidth]{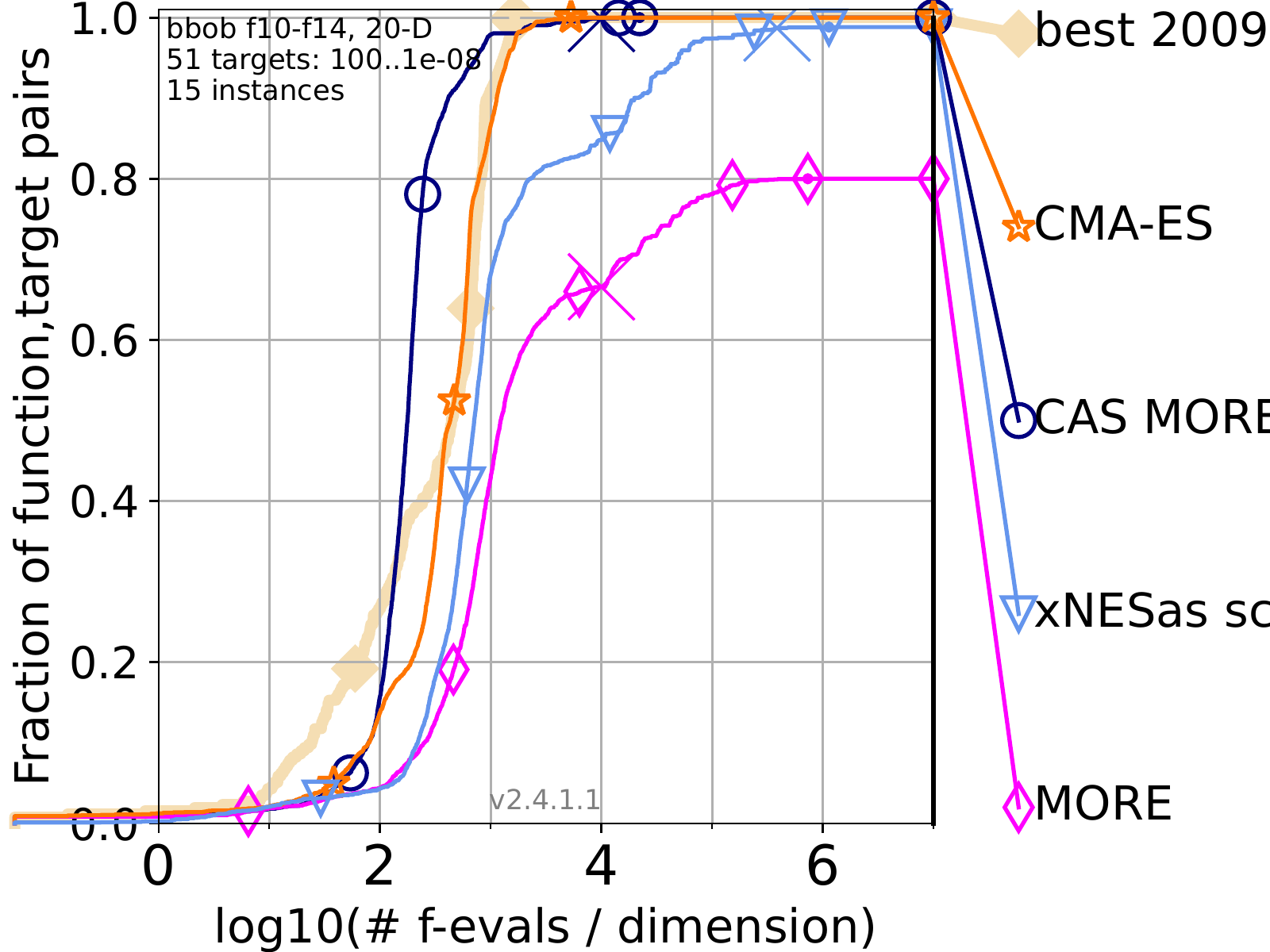}\\[-0.2em]
		{\small \sffamily multi-modal fcts}\hspace{1cm} & {\scriptsize \sffamily weakly structured multi-modal fcts}\hspace{1cm} & \hspace{-1cm}{\small \sffamily all fcts}\\
		\includegraphics[width=0.3\textwidth]{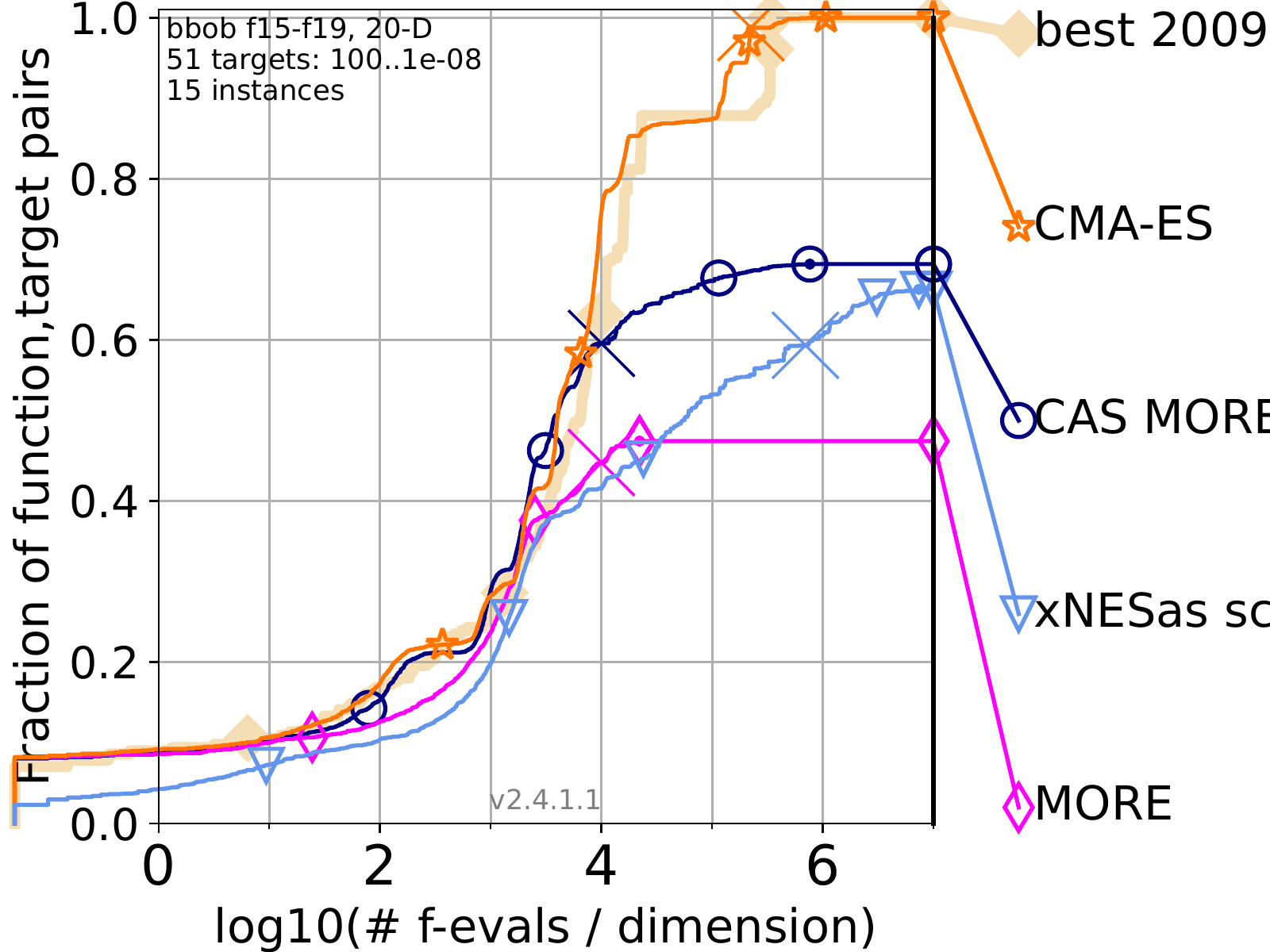}&
		\includegraphics[width=0.3\textwidth]{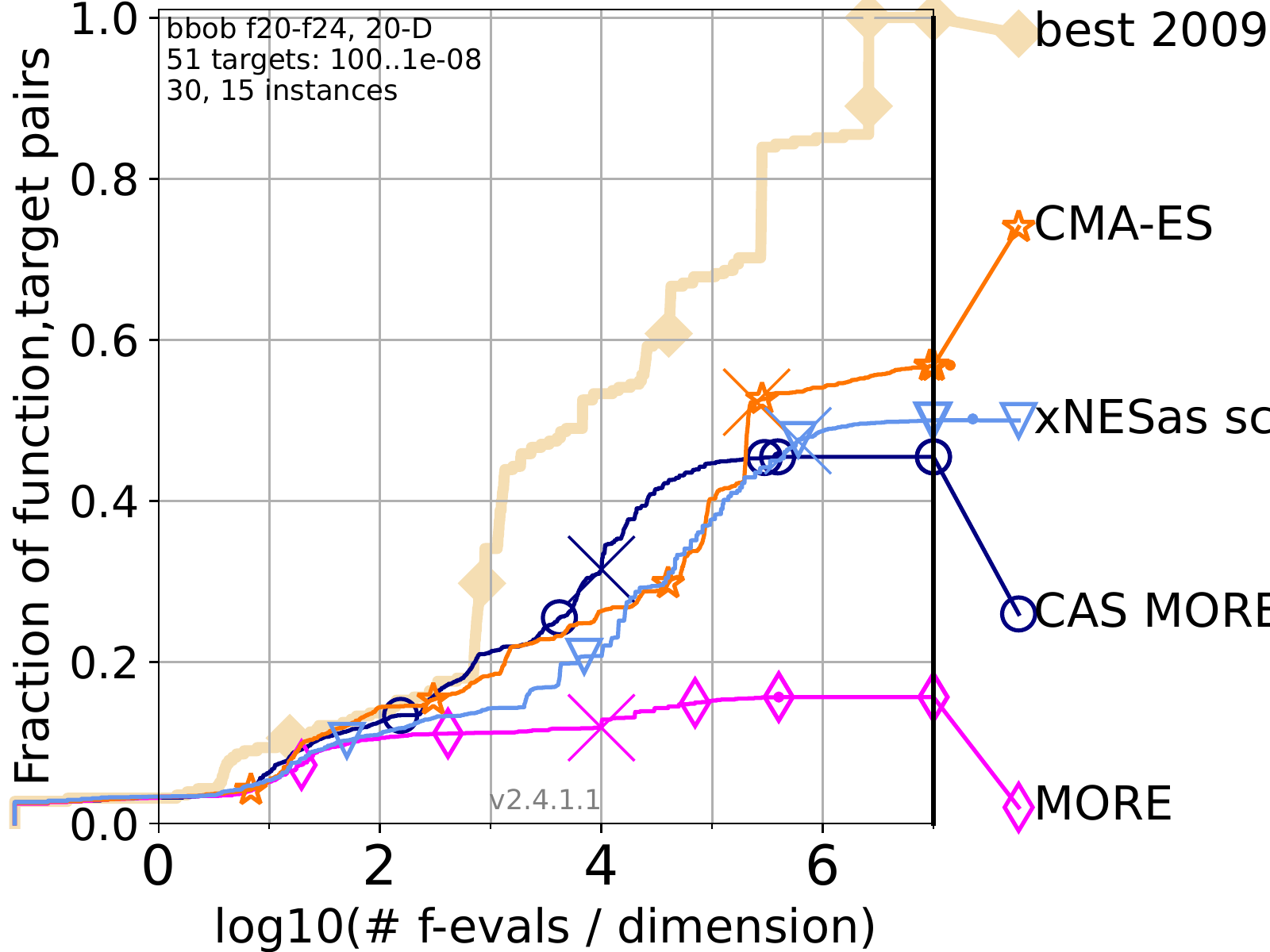}&
		\includegraphics[width=0.3\textwidth]{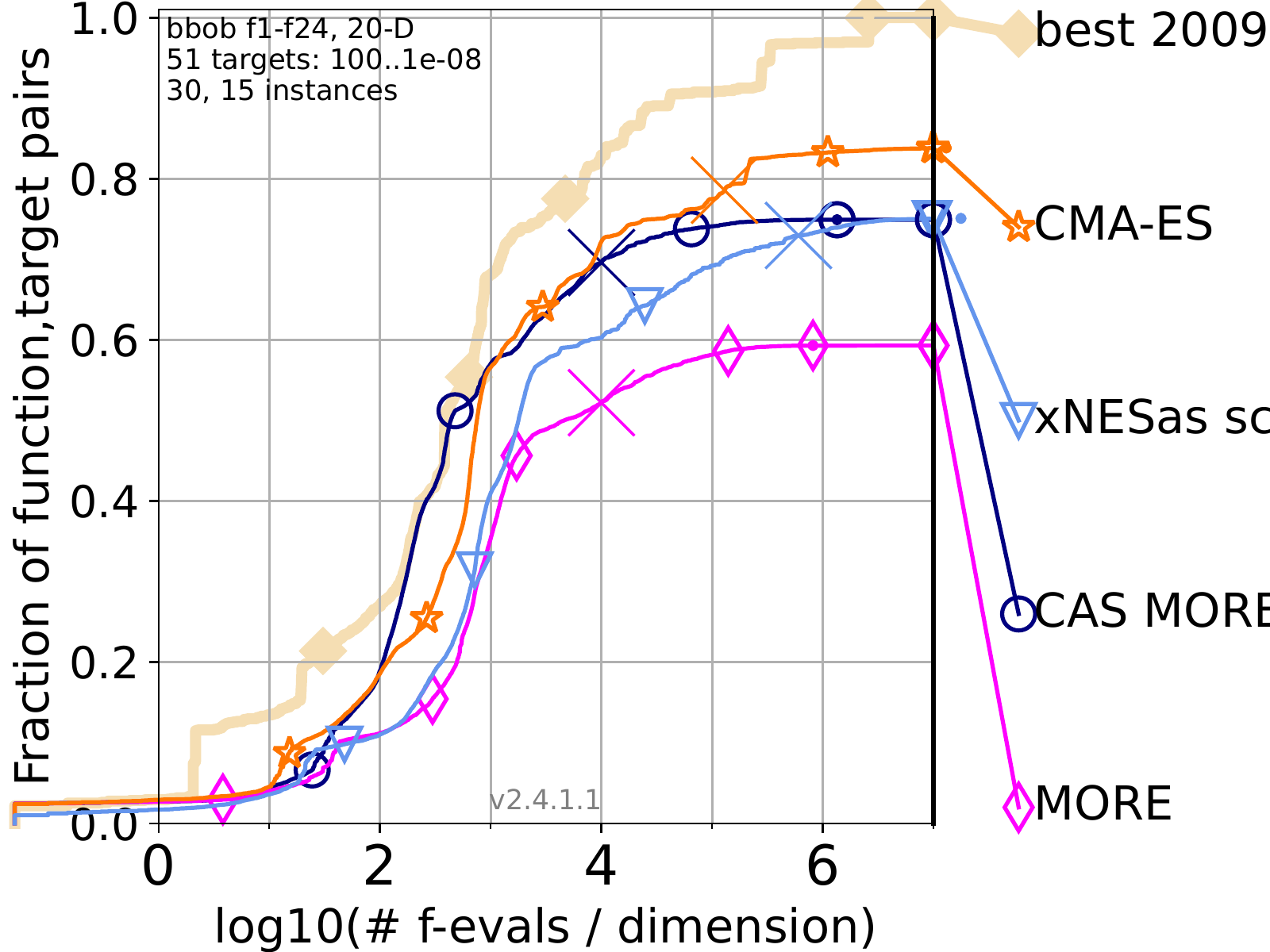}\\[-0.2em]
		\includegraphics[width=0.3\textwidth]{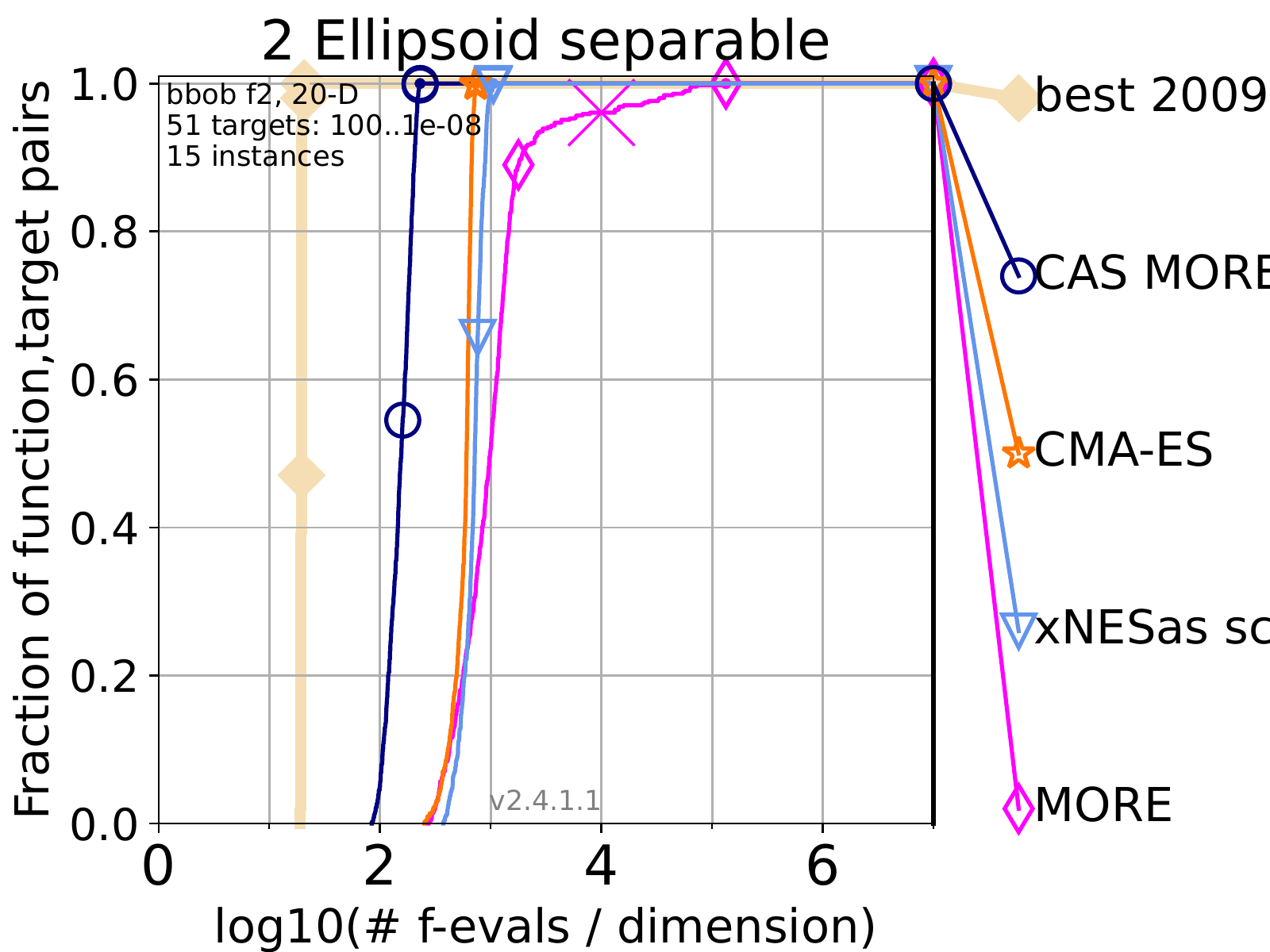}&
		\includegraphics[width=0.3\textwidth]{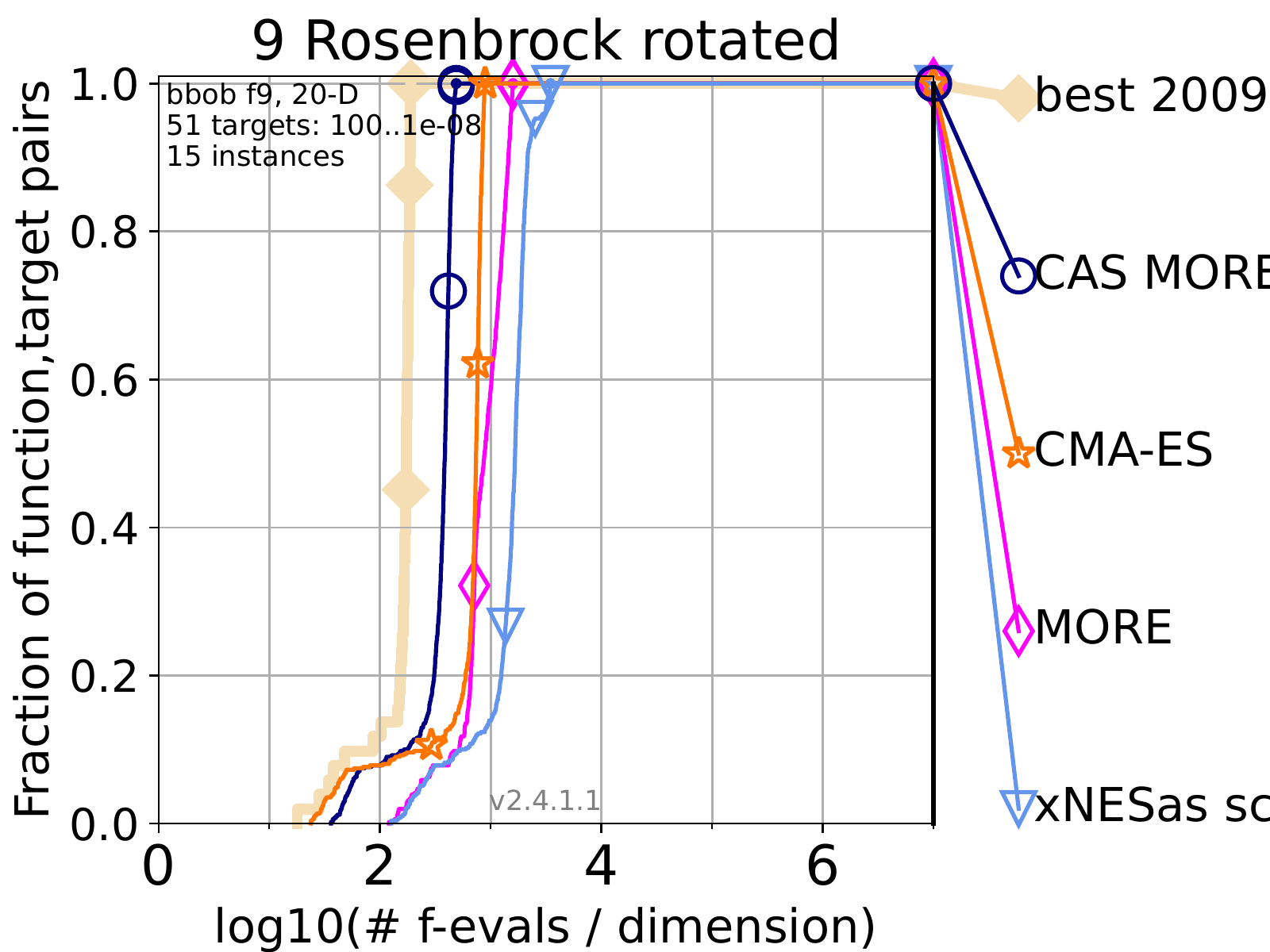}&
		\includegraphics[width=0.3\textwidth]{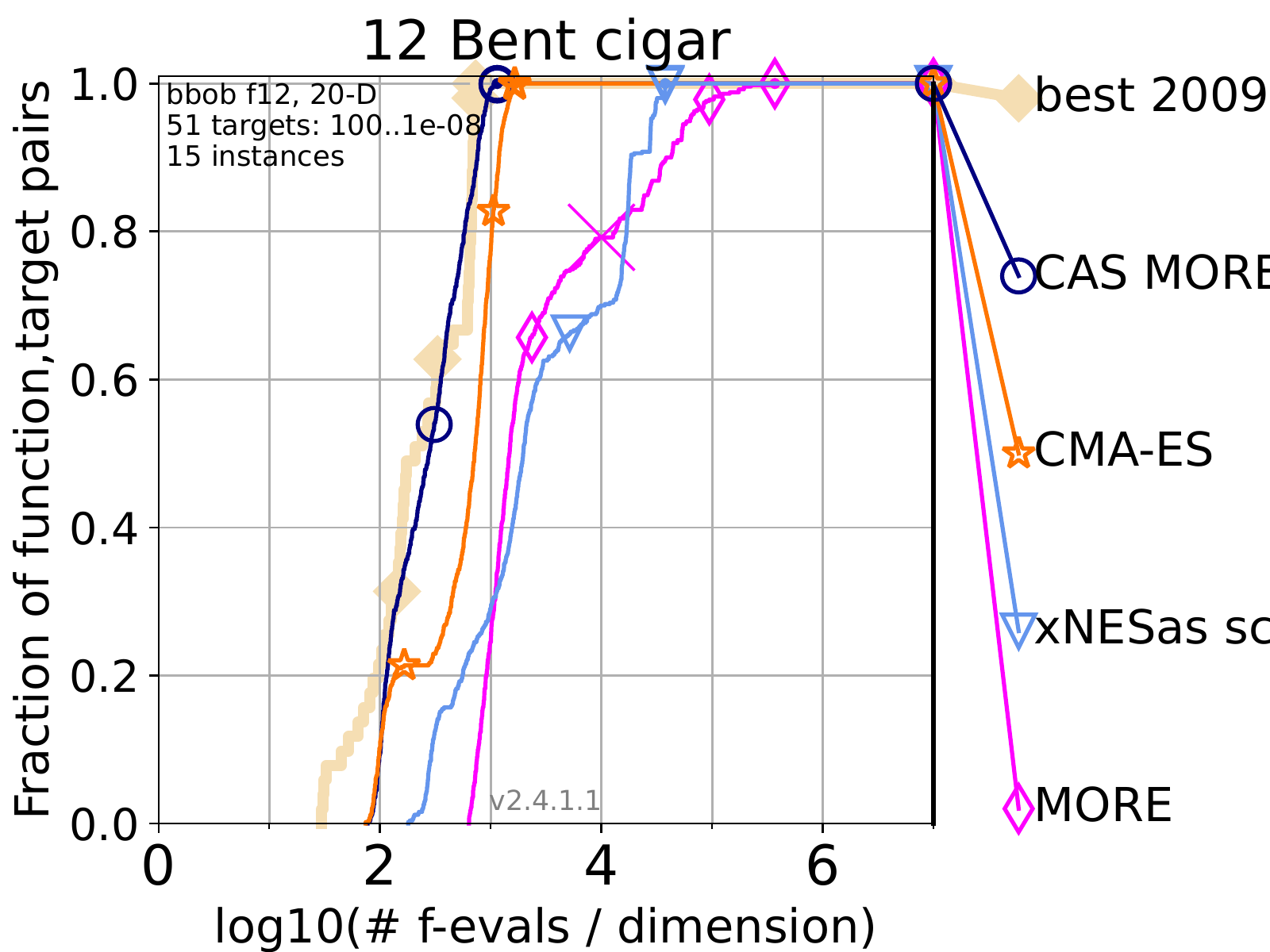}\\[-0.2em]
		\includegraphics[width=0.3\textwidth]{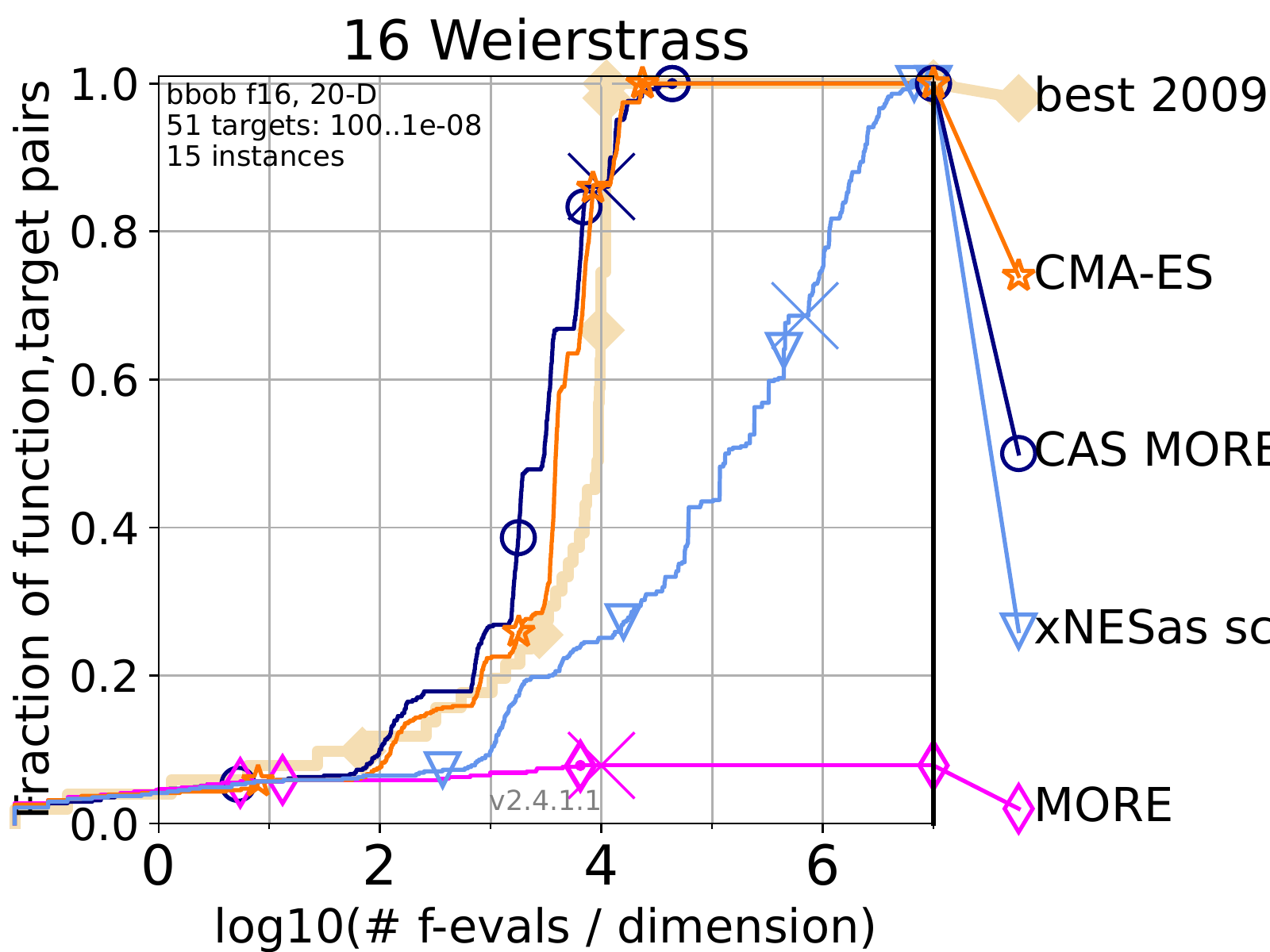}&
		\includegraphics[width=0.3\textwidth]{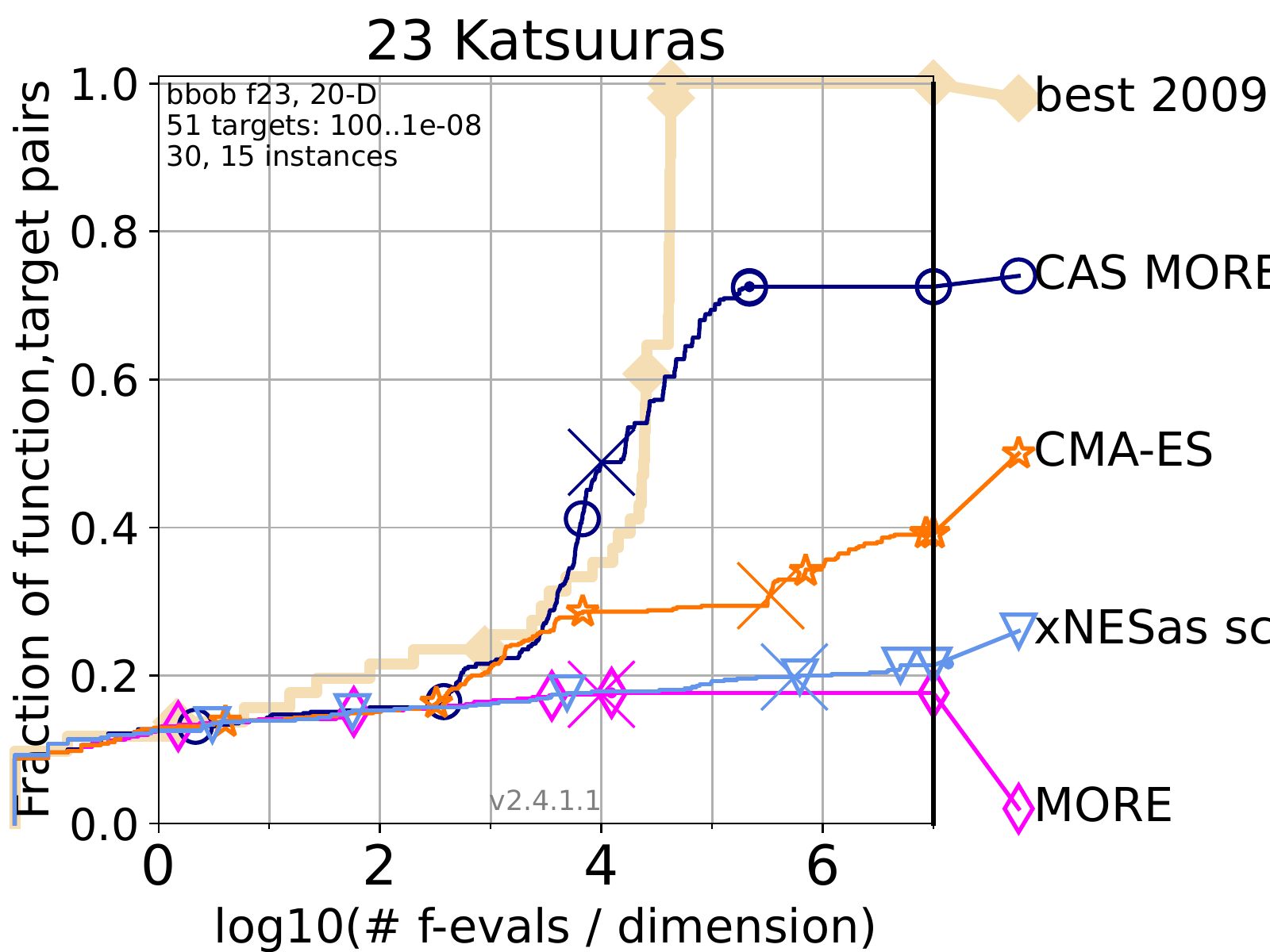}&
		\includegraphics[width=0.3\textwidth]{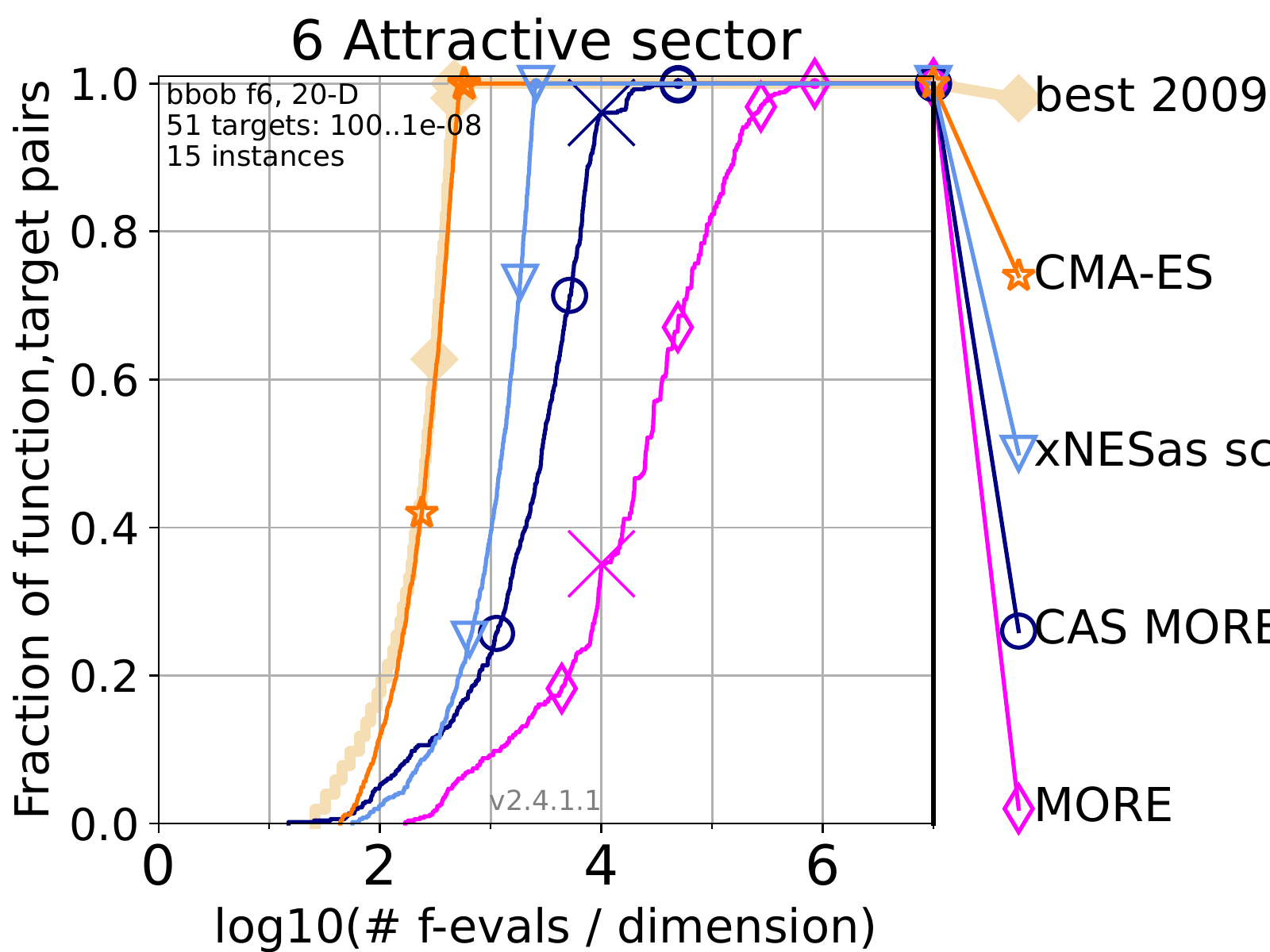}
		\vspace*{-1ex}
	\end{tabular}
	\caption{
		\label{fig:bbob20dresults}
		This figure shows the bootstrapped empirical cumulative distribution of the number of objective function evaluations divided by dimension (FEvals/DIM) for $51$ targets with target precision in $10^{[-8..2]}$ in 20-D representing the percentage of targets achieved over the number of function evaluations. The results are averaged over 15 instances of each function. As reference algorithm, the best algorithm from BBOB 2009 is shown as light thick line with diamond markers.
		The first two rows show the aggregated results for all functions and subgroups of functions.
		Additionally, we show the results of individual functions in the third and fourth row. Big thin crosses indicate the used budget median for the respective algorithm which is $\SI{10000} \dimF$ function evaluations for each trial of CAS-MORE. Runtimes to the right of the cross are based on simulated restarts and are used to determine a runtime for unsuccessful runs \citep{hansen2016coco}.
	}
\end{figure*}

Figure \ref{fig:bbob20dresults} shows runtime results aggregated over function groups and on single functions in dimension 20.
Plotted is the percentage of targets reached within the interval \SI{1e2}{} to \SI{1e-8} versus the log of objective function evaluations divided by the problem dimensionality.
The further left a curve is, the quicker it reached a certain target.
The first two rows correspond to combined results on separable, moderate, ill-conditioned, multi-modal and weakly structured multi-modal functions, as well as combined results from all 24 functions.
The third and fourth row shows results on selected individual functions.

We first notice that CAS-MORE has a significant advantage over the original formulation of MORE on almost all functions.
Furthermore, we can see that, especially on functions from the group of moderate and ill-conditioned functions such as Ellipsoid, Rosenbrock or Bent Cigar, CAS-MORE is able to improve state of the art results of competitors like CMA-ES and XNES.
On other functions on the other hand, for example the Attractive Sector function, MORE has slower convergence.
We found this shortcoming is mainly due to difficulties estimating a good model for this objective (see also Figure \ref{fig:as_ablation}).
Other hard functions include multi-modal functions such as the Rastrigin function, where MORE often focuses on a local optimum.
Here, the solution is to draw more samples in each iteration to improve the estimated model.
For more results on the BBOB suite we refer to the appendix. Typically, these black-box optimization benchmarks also focus purely on the fitness evaluation of the mean of the distribution and disregard the quality of the samples. While CAS-MORE is comparable to state-of-the-art black-box optimizers such as CMA-ES in these domains, the real benefits of MORE appear if we consider problems with noisy fitness evaluations and whenever we care also about the quality of the generated samples. This is for example the case in robot reinforcement learning problems which are discussed in the next sub-section.

\subsection{Episodic Reinforcement Learning}
In this section, we evaluate the performance of the improved MORE algorithm in episodic reinforcement learning problems.
Here, we often have to deal with noisy fitness evaluations.
Moreover, in this setup it is often beneficial to optimize the regret, i.e. the expected fitness instead of just the fitness at the mean of the search distribution.  
To this end, we design three problems, a simple planar hole-reaching task, and two tasks for a simulation of a 7 DoF Barret WAM robotic arm in Mujoco \citep{todorov2012mujoco} where the robot has to play table tennis and beer pong.
Common to all problems is that they are influenced by noise which often happens in reality, for example due to imperfect sensors.
As shown in our experiments, such noisy evaluations can have detrimental effects when using ranking based optimizers while MORE-based algorithms still perform well as they optimize the expected fitness. 

\subsubsection{Experimental Setup}
In each task, we use Probabilistic Movement Primitives \citep{NIPS2013_e53a0a29} to plan trajectories.
In noisy tasks, we increase both the samples per episode, as well as the buffer size by a factor of 5 compared to the default parameters and set $\epsilon_\mu = 1$ while keeping all other parameters at their defaults to cope with the added stochasticity.
We also compare classic mean and standard deviation normalization to the robust target normalization approach.
We again compare to CMA-ES and XNES and use the python implementations provided by the authors \citep{hansen2019cma, schaul2010pybrain}.

\begin{figure}
    \centering
    \begin{subfigure}[b]{0.3\textwidth}
        \centering
        \includegraphics[height=3.1cm]{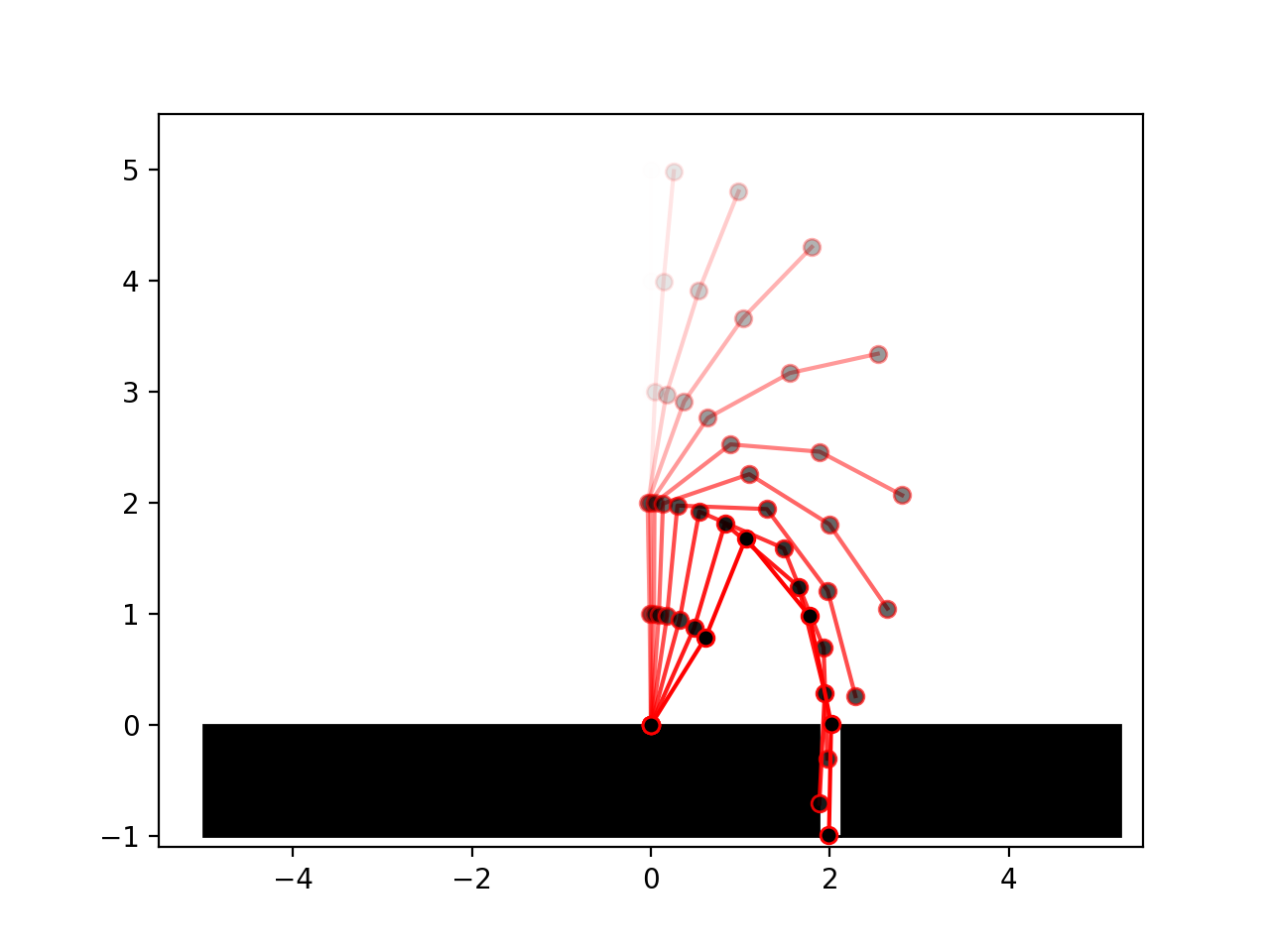}
        \caption{Hole-reaching task}\label{fig:hr_illu}
    \end{subfigure}
    \hfill
    \begin{subfigure}[b]{0.33\textwidth}
        \centering
        \includegraphics[height=2.75cm]{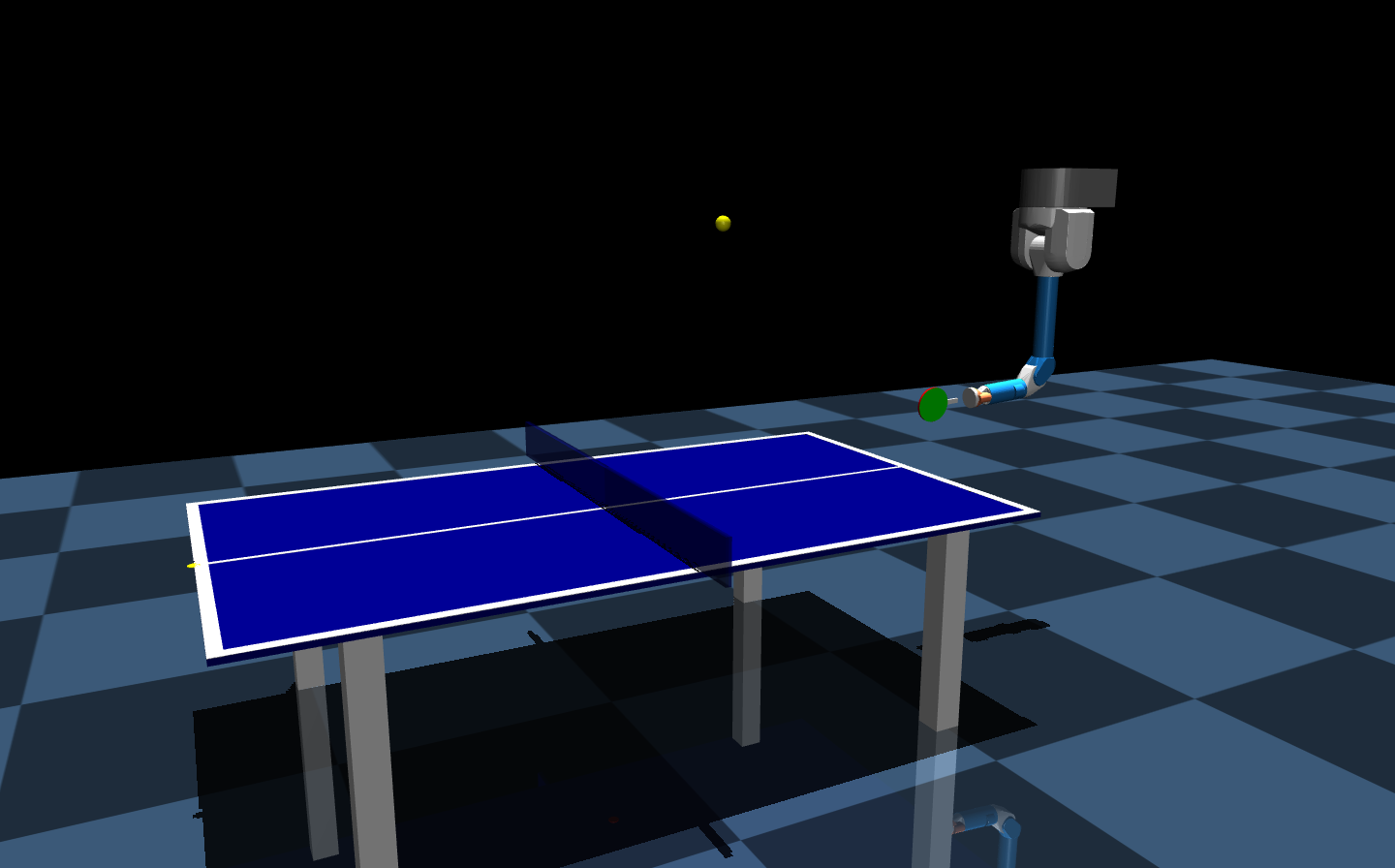}
        \caption{Table tennis task}\label{fig:tt_illu}
    \end{subfigure}
    \hfill
    \begin{subfigure}[b]{0.33\textwidth}
        \centering
        \includegraphics[height=2.75cm]{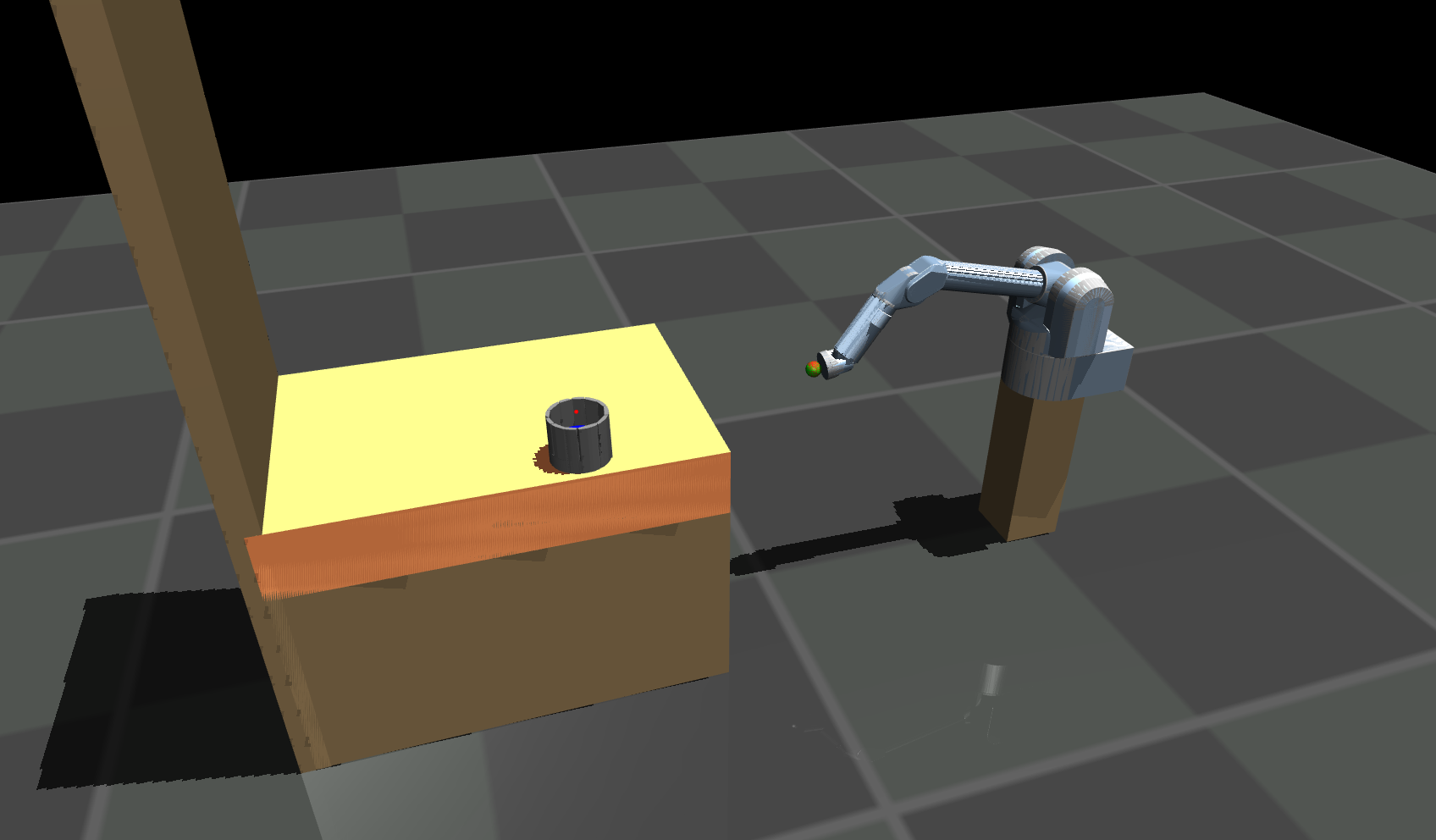}
        \caption{Beer pong task}\label{fig:bp_illu}
    \end{subfigure}
	\caption{Illustrations of all tasks considered for the episodic reinforcement learning evaluation.}\label{fig:task_illus}
\end{figure}

\subsubsection{Hole-Reaching}
\label{sec:hr}
The first task is an advanced version of the holereaching task from \cite{NIPS2015_36ac8e55} that more clearly shows the benefits of maximizing the expected fitness.
The end-effector of a 5-link planar robot arm has to reach into a narrow hole without colliding with the ground or the walls of the hole.
Additionally, the depth perception of the hole is subject to noise.
Each link has a length of \SI{1}{\meter}, the hole has a width of \SI{20}{\centi\meter}, a depth of \SI[separate-uncertainty = true]{1.00 \pm 0.02}{\meter} and is located \SI{2}{\meter} away from the robot's base.
For an illustration of a successful episode, see Figure \ref{fig:hr_illu}.

We learn the parameters of a ProMP with 3 basis functions, resulting in 15 parameters to be optimized.
An episode consists of \num{200} time steps and in each iteration of the learning algorithm, we draw 60 new samples.
The cost function of the task is composed of two stages.
First, the distance of the end-effector to the entrance of the hole is minimized, subsequently, the reward increases the further down the end-effector reaches.
The reward is scaled down with a constant factor if the robot touches the ground.
Additional penalty costs for acceleration ensure a smooth and energy-efficient trajectory.

\begin{figure}
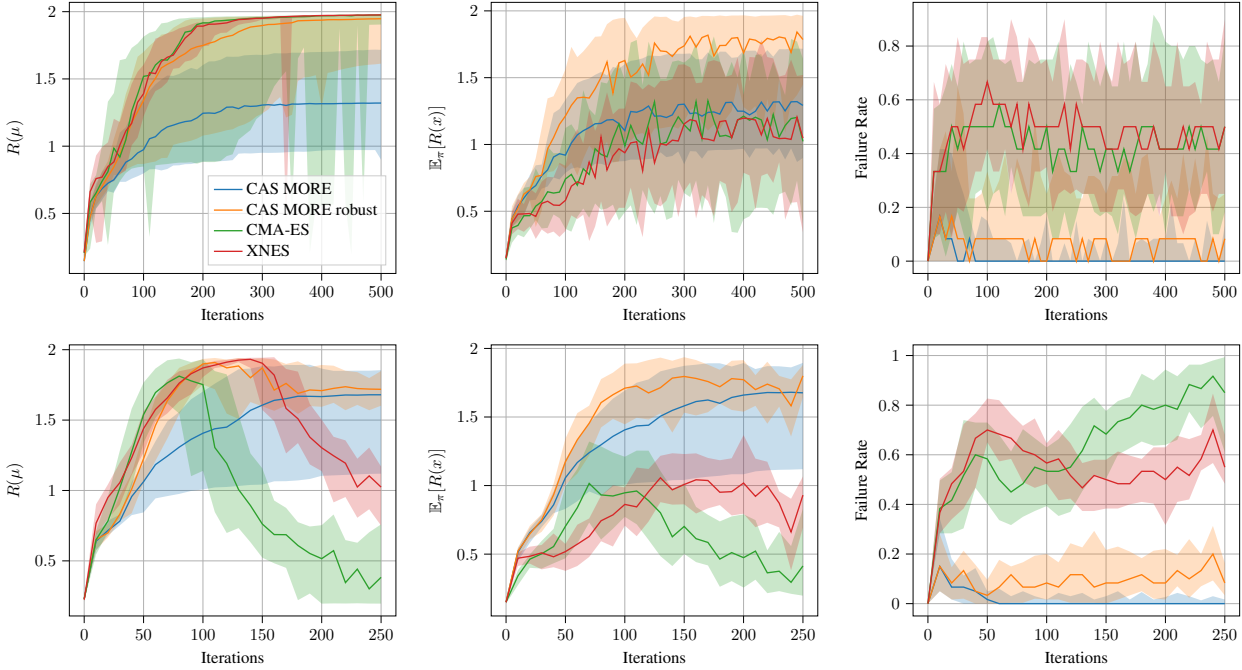

    \centering
    \includestandalone[mode=buildnew, width=\linewidth]{figures/hr/hr_v3_v4_results_all}
	\caption{This figure shows the results of the noise-less hole-reacher experiment in the upper row and the results of the noisy hole-reacher experiment in the lower row. Plotted is the expected performance of the mean in the left column, the mean of episode returns of the samples drawn in each iteration in the center column, and the percentage of trajectories that led to collisions with the ground in the right column. We show the results in terms of the median and 5\% / 95\% quantiles over 25 runs.}
    \label{fig:hr_v3_v4}
\end{figure}

Before we look at the results of the noisy problem, we examine the results of a noise-less version of the task in the upper row of Figure \ref{fig:hr_v3_v4}.
The left plot shows the performance of the mean of the search distribution and the center plot shows the average of the samples produced by the search distribution.
The result is averaged over 25 trials and we plot the median and 5\%/95\% quantiles.
The task specific measure shown in right plot of the upper row in Figure \ref{fig:hr_v3_v4} is given by the percentage of samples that collided with the ground, also denoted as failure rate (lower is better).
Modelling the reward function with few samples per episode is a difficult task as the blue curve for CAS-MORE in the left plot with a model using non-robust standardization of the rewards suggests.
Only when using robust normalization, we are able to solve the task and find parameters that can compete with the rank-based algorithms CMA-ES and XNES.
Unlike CMA-ES and XNES, which in some trials shift the mean towards a solution that touches the ground, indicated by the sharp drops in later iterations, CAS-MORE robustly optimizes the distribution towards a well performing solution.
When looking at the quality of the samples produced by the search distribution during the optimization in the center and right plot, the clearly defined optimization objective together with a model that incorporates actual function values becomes apparent.
While CMA-ES and XNES produce samples with poor quality, the distribution optimized with MORE and a robust model fitting approach produces samples that perform well and only rarely collide with the ground.

Next, we examine the results of the noisy task, shown in the lower row of Figure \ref{fig:hr_v3_v4}.
We again plot the performance of the mean in the left plot, this time averaged over 100 rollouts, the quality of the produced samples in the center plot, and failure rate in the right plot.
We can see that rank based algorithms such as CMA-ES and XNES tend optimize quickly to a well performing mean but, as soon as the end-effector closes in on the ground, can't cope with the stochasticity of the task. 
The reason for this effect is that ranking based algorithms tend to use a high rank for samples where the fitness evaluation had by chance a very good outcome (i.e., reaching this deep worked due to the noise in the depth of the hole), however, on expectation the sample is performing poorly.

In contrast, MORE optimizes towards a distribution that maximizes the expected reward which in this case translates to sampling trajectories that minimize the risk of touching the ground.
The result is a distribution with a more conservative mean but a higher expected reward as can be seen in the bottom center plot of Figure \ref{fig:hr_v3_v4}.
The poor sample quality of CMA-ES and XNES can again be seen in the right plot which shows that these ranking based algorithms produce samples with high probability of colliding with the ground.
The increased sample size allows for a better model estimate and we can achieve a trade-off between a more conservative mean using the standard normalization of targets and an aggressive mean and a slightly worse expected reward by applying the robust reward normalization introduced in Section \ref{sec:model_learning}. 
As the robust target normalization variant of our algorithm is also diminishing the effect of negative punishment in case of colliding with the wall, it obtains a slightly higher collision rate than our algorithm with standard normalization.
Yet, robust target normalization allows for more aggressive updates of the search distribution resulting in a faster learning progress.   

\subsubsection{Table Tennis}
In the second task, we want to teach a 7 DoF Barret WAM robotic arm a fore-hand smash. 
The goal is to return a table-tennis ball and place it as close as possible to the far edge of table on the opponent's side.
In every episode, the ball is initialized at the same position but the velocity in x-direction (approaching the robot) is noisy.
Thus, a robust strategy is to account for the uncertain velocity and aim for a spot that is not too close to the edge.
An illustration of the task can be found in Figure \ref{fig:tt_illu}.

We learn the parameters of a ProMP where we use 2 basis functions for each joint, resulting in 14 parameters to be optimized.
The weights of the movement primitive are initialized to perform a forehand motion without hitting the ball.
The reward function is composed of three stages.
First, the distance between the ball and the racket is minimized to enforce hitting of the ball.
Second, a term is added to minimize the distance between the landing point of the ball and the far edge of the table.
Once the ball lands on the table, the reward increases with the distance of the ball.
If it drops to the bottom, we go back to the second stage.

\begin{figure}
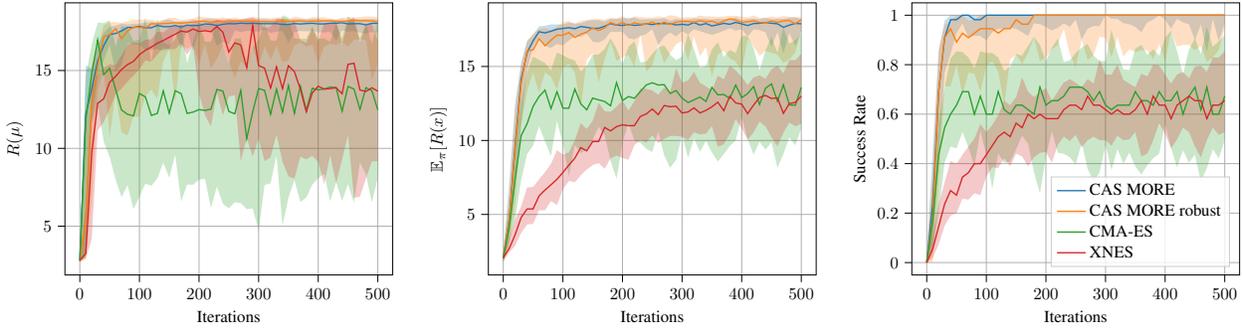

    \centering
    \includestandalone[mode=buildnew, width=\linewidth]{figures/tt/tt_results_all}
	\caption{This figure shows the results of the table tennis experiment. Plotted is the expected performance of the mean (left), the mean of episode returns of the samples drawn in each iteration (center), and the percentage of strokes that successfully landed on the opponent's side of the table (right). We show the results in terms of the median and 5\% / 95\% quantiles over 25 runs.}
	\label{fig:tt_v4}
\end{figure}

The figures in the second row of Figure \ref{fig:tt_v4}
show the learning progress of CAS-MORE compared to CMA-ES and XNES where in each iteration we draw 55 samples.
Again, we examine the performance of the mean of the search distribution, the expected performance under the current distribution, and the success rate of returning the ball to the opponents side of the table.
Results are similar to the previous experiment with CAS-MORE finding policies that are able to return the ball in nearly every case, while CMA-ES and XNES only manage to successfully return roughly \SI{60}{\percent} of the balls as the use of the ranking is biasing the optimization towards "lucky" fitness evaluations.
Using the robust normalization is slightly worse than standard normalization as poor rewards are not penalized as heavy.

\subsubsection{Beerpong}
We use the same simulated robot as before to throw a ball towards a table where it first needs to bounce at least once and then fly into a cup.
This time, the ball release is noisy.
Unlike the previous two experiments where a conservative strategy resulted in successful executions of the task, it is now not possible to find a conservative strategy.
The noise will always cause some trials to miss the cup.

While in this example CMA-ES and XNES converge to a good mean of the search distribution, the sample quality of CAS-MORE is still significantly better in comparison to these methods.

\begin{figure}
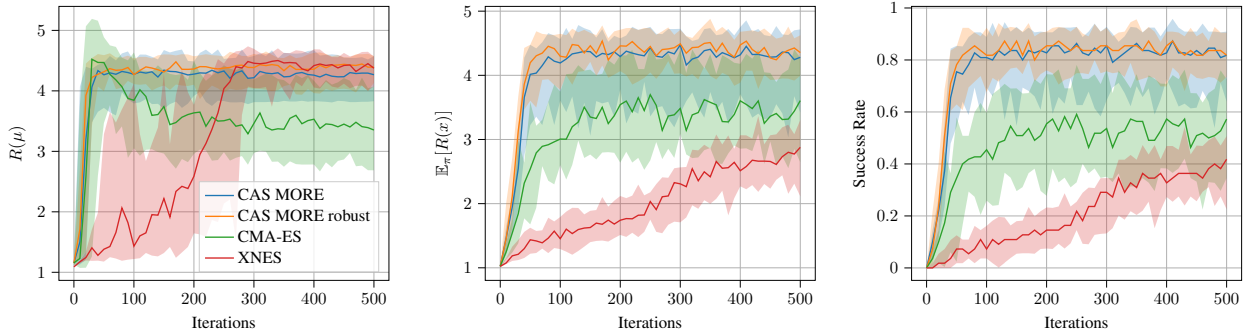

    \centering
    \includestandalone[mode=buildnew, width=\linewidth]{figures/bp/bp_results_all}
	\caption{This figure shows the results of the beer pong experiment. Plotted is the expected performance of the mean (left), the mean of episode returns of the samples drawn in each iteration (center), and the percentage of strokes that successfully landed in the cup (right). We show the results in terms of the median and 5\% / 95\% quantiles over 25 runs.}
	\label{fig:bp_v1}
\end{figure}

\section{Conclusion}
In this paper, we presented CAS-MORE, a new version of Model-based Relative Entropy Stochastic Search, based on a coordinate ascent strategy on the mean and covariance of the search distribution and an adaptive entropy schedule, and an improved model fitting process.
We show how the parameter updates follow the direction of the natural gradient and, as the model estimation is not based on rankings of objective function values, we truly optimize the expected fitness under the search distribution.
The result is a regret-aware optimization which we show to outperform state of the art algorithms such as CMA-ES, especially on noisy episodic reinforcement learning tasks.

\section{Acknowledgements}
The authors acknowledge support by the state of Baden-Württemberg through bwHPC. Research that lead to this work was funded by the Federal Ministry of Education and Research (BMBF) and the state of Hesse as part of the NHR Program.

\appendix

\section{Derivation of CA-MORE Dual}
Before we solve the optimization problems, we will first state the closed-form solution of the objective and KL-divergence using a quadratic model under multi-variate Gaussian distributions.
The solution to the objective is given by 
\begin{equation*}
	\int_{\bm{x}} \pi(\bm{x}) \hat{f}(\bm{x}) \dx = - \frac{1}{2} \vecT{\bm{\mu}} \bm{A} \bm{\mu} - \frac{1}{2} \tr{\bm{A} 
		\bm{\Sigma}} + \vecT{\bm{\mu}}\bm{a} + a_0 
\end{equation*}
and the KL-divergence between two Gaussian distributions is given by 
\begin{align*}
	\KLdiv{\pi(\bm{x})}{\pi_t(\bm{x})} =& \frac{1}{2} \left\{(\bm{\mu}_{t} - \bm{\mu})^\mathrm{T} \bm{\Sigma}_{t}^{-1} (\bm{\mu}_{t} - \bm{\mu}) \right.  \\ 
	& \left. + \mathrm{tr}(\bm{\Sigma}_{t}^{-1} \bm{\Sigma}) - k + \log \vert \bm{\Sigma}_{t} \vert - \log \vert \bm{\Sigma} \vert \right\}.
\end{align*}

\subsection{Mean Update}
Setting $\bm \Sigma = \bm \Sigma_t$, the optimization problem is given by
\begin{maxi*}|l|
	{\bm{\mu}}{- \frac{1}{2} \vecT{\bm{\mu}} \bm{A} \bm{\mu} + \vecT{\bm{\mu}}\bm{a}}
	{}{}
	\addConstraint{\frac{1}{2} \vecT{(\bm{\mu}_{t} - \bm{\mu})} \bm{\Sigma}_{t}^{-1} (\bm{\mu}_{t} - \bm{\mu}) }{\leq \epsilon_\mu}
\end{maxi*}
and the Lagrangian is given by
\begin{align*}
L(\bm{\mu}, \lambda) &= -\frac{1}{2} \bm{\mu}^\mathrm{T} \bm{A} \bm{\mu} + \bm{\mu}^\mathrm{T} \bm{a} \\ %
&+ \lambda \left(\epsilon_\mu - \frac{1}{2} (\bm{\mu}_{t} - \bm{\mu})^\mathrm{T} \bm{\Sigma}_{t}^{-1} (\bm{\mu}_{t} - \bm{\mu})\right) 
\end{align*}
where $\lambda$ is a Lagrangian multiplier.
The optimal solution $\bm{\mu}^\ast$ in terms of the Lagrangian multipliers can be found by differentiating $L$ with respect to $\bm{\mu}$ and setting it to 0, i.e,
\begin{align*}
\frac{\partial L}{\partial \bm{\mu}} &= -\bm{A} \bm{\mu} + \bm{a} + \lambda \bm{\Sigma}_{t}^{-1}( \bm{\mu}_{t} - \bm{\mu}) \overset{!}{=} 0.
\end{align*}
Using the solution $\lambda^\ast$,
\begin{align*}
\bm{\mu}^\ast &= \underbrace{(\lambda^\ast \bm{\Sigma}_{t}^{-1} + \bm{A})}_{\bm{M}_\mu(\lambda^\ast)}{}^{-1} \underbrace{(\lambda^\ast \bm{\Sigma}_{t}^{-1} \bm{\mu}_{t} + \bm{a})}_{\bm{m}_\mu(\lambda^\ast)} = \bm{M}_\mu(\lambda^\ast)^{-1} \bm{m}_\mu(\lambda^\ast) %
\end{align*}

After rearranging terms, the dual problem for the mean is given by
\begin{align*}
    g_\mu(\lambda) &= \lambda \epsilon_\mu + \frac{1}{2} \Big(\bm{m}_\mu(\lambda)^\mathrm{T} \bm{M}_\mu(\lambda)^{-1} \bm{m}_\mu(\lambda) \\
    &- \lambda \bm{m}_{t}^\mathrm{T} \bm{M}_t^{-1} \bm{m}_t\Big).
\end{align*}

\subsection{Covariance Update}
Analogously, we set $\bm \mu = \bm \mu_t$.
The optimization problem for the covariance is given by
\begin{maxi*}|l|
	{\bm{\Sigma}}{- \frac{1}{2} \tr{\bm{A} 
			\bm{\Sigma}}}
	{}{}
	\addConstraint{\frac{1}{2} \left(\tr{\bm{\Sigma}_{t}^{-1} \bm{\Sigma}}  - k + \log \vert \bm{\Sigma}_{t} \vert - \log \vert \bm{\Sigma} \vert \right)}{< \epsilon_\Sigma}
\end{maxi*}
and the Lagrangian for the covariance matrix optimization is given by
\begin{align*}
	L(\bm{\Sigma}, \nu) &= - \frac{1}{2} \tr{\bm{A} \bm{\Sigma}} \label{eq:ca_more_dual_cov2}\\
	&+ \nu \left(\epsilon_\Sigma - \frac{1}{2} \left(\tr{\bm{\Sigma}_{t}^{-1} \bm{\Sigma}}  - k + \log \vert \bm{\Sigma}_{t} \vert - \log \vert \bm{\Sigma} \vert \right) \right) 
	\end{align*}
	where $\nu$ is again a Lagrangian multiplier.
	The optimal solution $\bm{\Sigma}^\ast$ can be found analogously and is given by
	\begin{align*}
	\frac{\partial L_\Sigma}{\partial \bm{\Sigma}} &= -\frac{1}{2} \bm{A} - \frac{1}{2} \nu \bm{\Sigma}_{t}^{-1} + \frac{1}{2} \nu \bm{\Sigma}^{-1} \overset{!}{=} 0.
	\end{align*}
	With the solution $\nu^ast$,
	\begin{align*}
	\bm{\Sigma}^\ast &= \underbrace{\left((\nu^\ast \bm{\Sigma}_{t}^{-1} + \bm{A}) / \nu^\ast\right)}_{\bm{S}(\nu^\ast)}{}^{-1} = \bm{S}(\nu^\ast)^{-1}. %
	\end{align*}

After rearranging terms again, the dual for the covariance is given by
\begin{align*}
    \bm{\Lambda}(\nu) = \frac{\nu \bm{\Sigma}_{t}^{-1} + \bm{A}}{\nu}.
\end{align*}

\section{Robust Target Normalization}
We provide pseudo-code for the robust target normalization technique in Algorithm \ref{alg:targ_norm}.

\begin{figure}
	\begin{algorithmic}[1]
	\Procedure{Normalize}{$\mathcal{Y}$}\Comment{Robust normalization of targets}
		\State $\bar{y}_{\mathcal{Y}} \gets \frac{1}{\vert \mathcal{Y} \vert} \sum_q^{\vert \mathcal{Y} \vert} y_q$
		\State $\sigma_{\mathcal{Y}} \gets \sqrt{\frac{1}{\vert \mathcal{Y} \vert} \sum_q^{\vert \mathcal{Y} \vert} (y_q - \bar{y}_{\mathcal{Y}})^2}$
		\State $y \gets \frac{y - \bar{y}_{\mathcal{Y}}}{\sigma_{\mathcal{Y}}}$ \Comment{Standardize all elements in $\mathcal{Y}$} 
 		\State $\mathcal{I} \gets -v_\text{clip} < y < v_\text{clip}$ \Comment{Boolean mask of all elements in $\mathcal{Y}$ that are in $(-v_\text{clip}, v_\text{clip})$} 
		\State $\mathcal{S} \gets \{y \in \mathcal{Y} \mid -v_\text{clip} < y < v_\text{clip} \}$ \Comment{All elements in $\mathcal{Y}$ that are in $(-v_\text{clip}, v_\text{clip})$} 
		\State $k \gets \text{Kurt}[\mathcal{S}] - 3$\Comment{Excess kurtosis of the elements in $\mathcal{S}$} 
		\If{$k > 0.55$ \algorithmicand{} $\sigma_{\mathcal{Y}} \neq 1$}
			\State $\mathcal{Y}(\mathcal{I}) \gets \Call{Normalize}{\mathcal{S}}$ \Comment{Normalize elements in $\mathcal{S}$} 
		\EndIf
		\State $\mathcal{Y} \gets \text{clip}(\mathcal{Y}, \min(\mathcal{Y}(\mathcal{I})), \max(\mathcal{Y}(\mathcal{I}))$
		\State \textbf{return} $\mathcal{Y}$
	\EndProcedure
	\end{algorithmic}
	\caption{Robust target normalization}\label{alg:targ_norm}
\end{figure}

\section{Impact of Data Pre-processing}
\label{app:data_pre_proc}
\begin{figure}
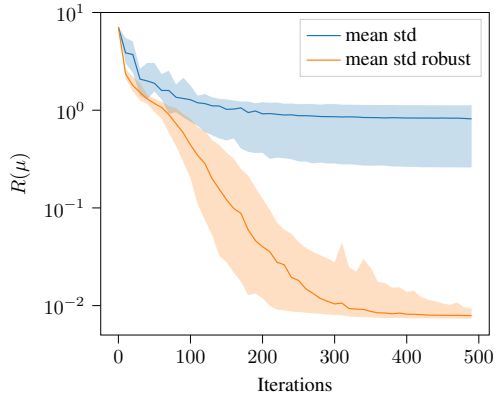

	\centering
	\includestandalone[width=0.4\linewidth]{figures/hr/hr_v2_dist_to_opt_median}
	\caption{Comparison of mean/std normalization and robust normalization on a penalty based reward function for the hole-reaching task.}
	\label{fig:hr_v2}
\end{figure}

In reinforcement learning, a reward function that accurately describes the task and is easy to optimize is not always given.
Here, we want to demonstrate that even with a reward function that includes large jumps, using a robust target normalization scheme leads to good results.
To this end, we use a different reward function formulation for the hole-reacher task from Section \ref{sec:hr}.
The reward in this case is defined as the negative squared distance to a target point at the bottom of the hole minus a large penalty whenever the robot hits the ground.
We leave the depth at -1 to focus on the effects of target normalization.
Figure \ref{fig:hr_v2} shows the negative reward (the cost) on a log-scale and we compare CAS-MORE using mean/std normalization and robust mean/std normalization.
A cost of 1 corresponds to the end-effector of the robot moving close to the entrance of the hole but failing to reach inside.
Only robust normalization is able to capture both, the task reward and the penalty, and guides the robot to reach down the hole.

\section{Hyper-Parameters}
We provide default hyper-parameters in terms of the problem dimensionality $\dimF$ in Table \ref{tab:default_params} which we empirically found to work well over all benchmark functions. For some functions, a higher bound on the mean often leads to quicker convergence but may result in divergence for others.
\begin{table*}
    \renewcommand{\arraystretch}{1.3}
	\caption{Empirically found default hyper-parameters for CAS-MORE based on the problem dimensionality $\dimF$.}
	\label{tab:default_params}
    \centering 
	\begin{tabular}{ c c}
		Parameter & Default Value \\
		\toprule
		$\popSize$: Population size  &  $4 + \floor{3 \log(\dimF)}$\\
		$\queueSize_{\text{max}}$: Maximum queue size & $\max\{\ceil{1.5 (1 + \dimF + \dimF (\dimF + 1) / 2)}, \; 8 (\dimF + 1)\}$\\
		$\epsilon_\mu$: Trust-region for the mean  & 0.5\\
		$\epsilon_\Sigma$: Trust-region for the covariance  & $\frac{1.5}{10 + \dimF^{1.5}}$ \\
		$c_\sigma$: Smoothing factor of evolution path  &  $\frac{1}{2 + \dimF^{0.75}}$\\
		$v_\text{clip}$: Clip value for robust normalization  &  3\\
		Excess kurtosis threshold & 0.55 \\
		\bottomrule
	\end{tabular}
\end{table*}

\section{Black-box Optimization Benchmarks}

\begin{figure*}
	\centering
	\begin{tabular}{@{}c@{}c@{}c@{}c@{}}
		\includegraphics[width=0.24\textwidth]{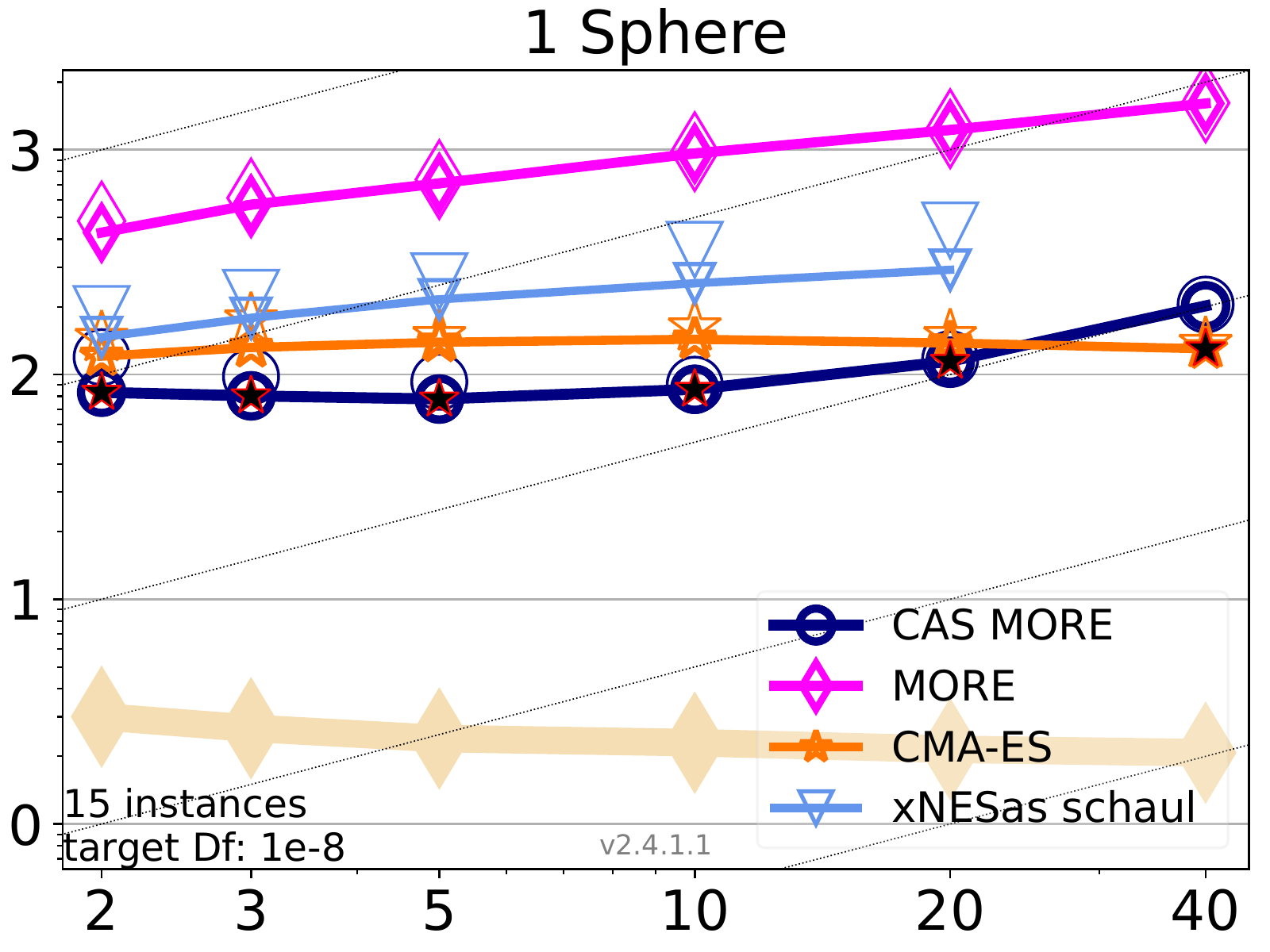}&
		\includegraphics[width=0.24\textwidth]{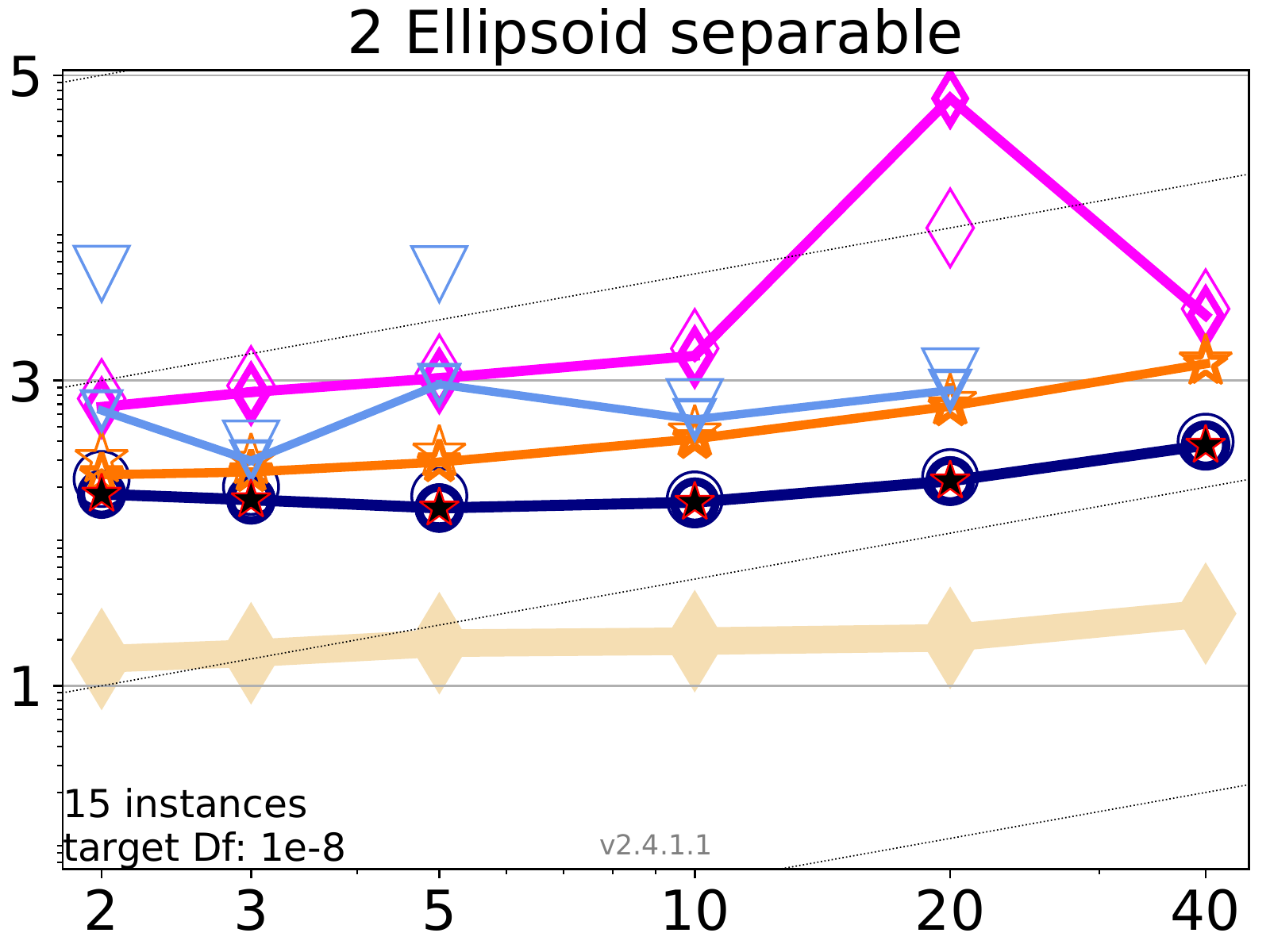}&
		\includegraphics[width=0.24\textwidth]{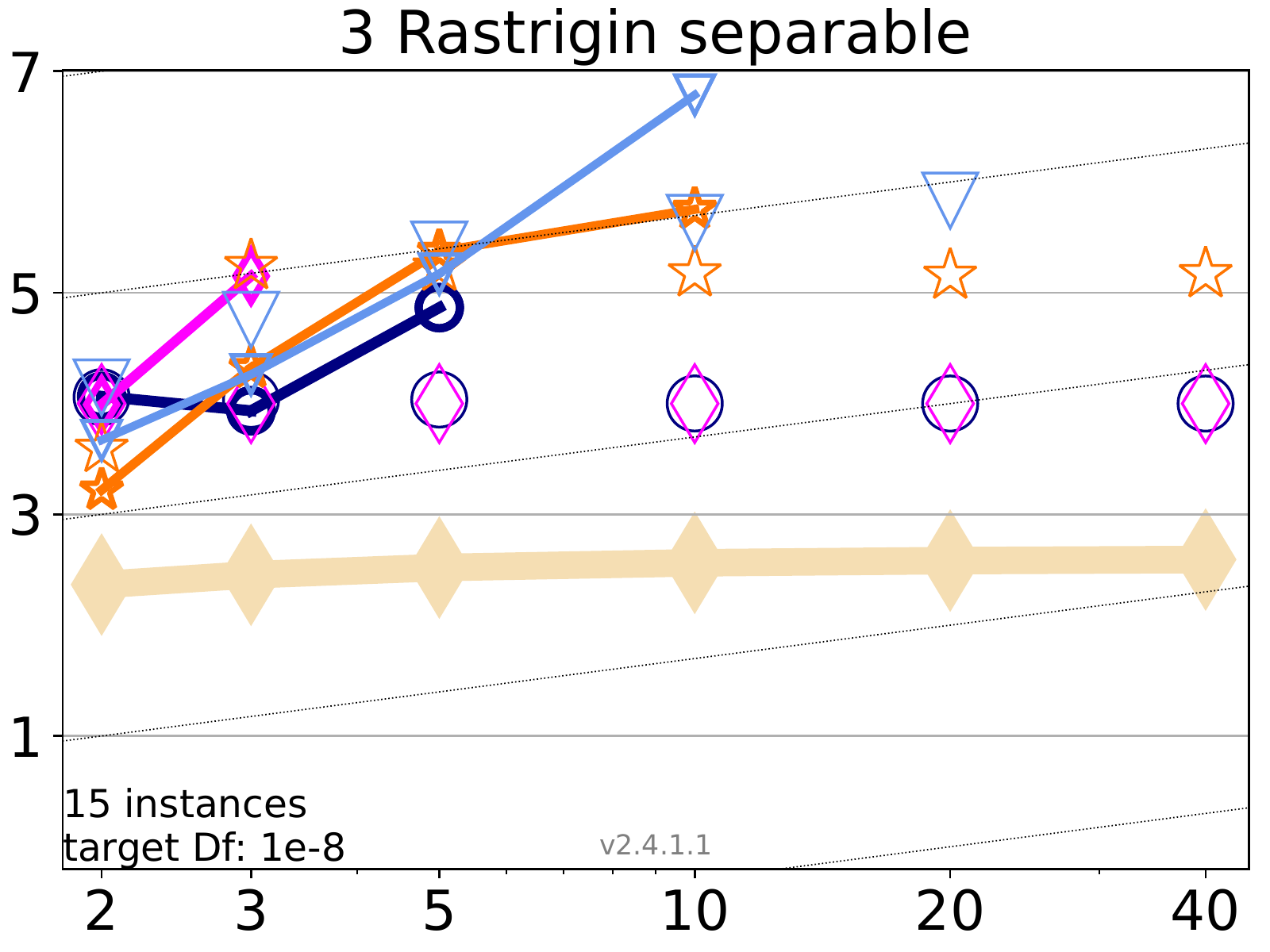}&
		\includegraphics[width=0.24\textwidth]{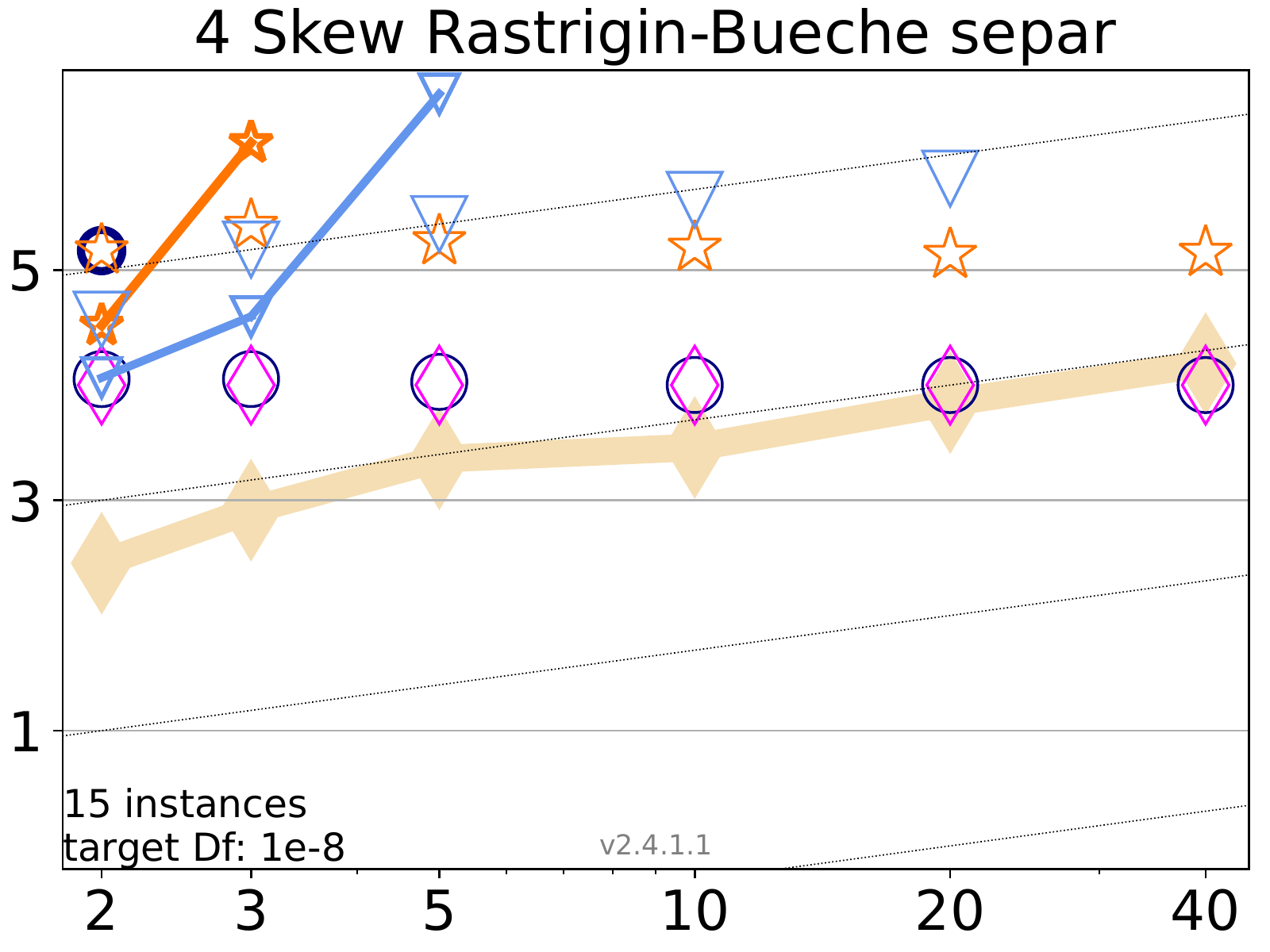}\\[-0.25em]
		\includegraphics[width=0.24\textwidth]{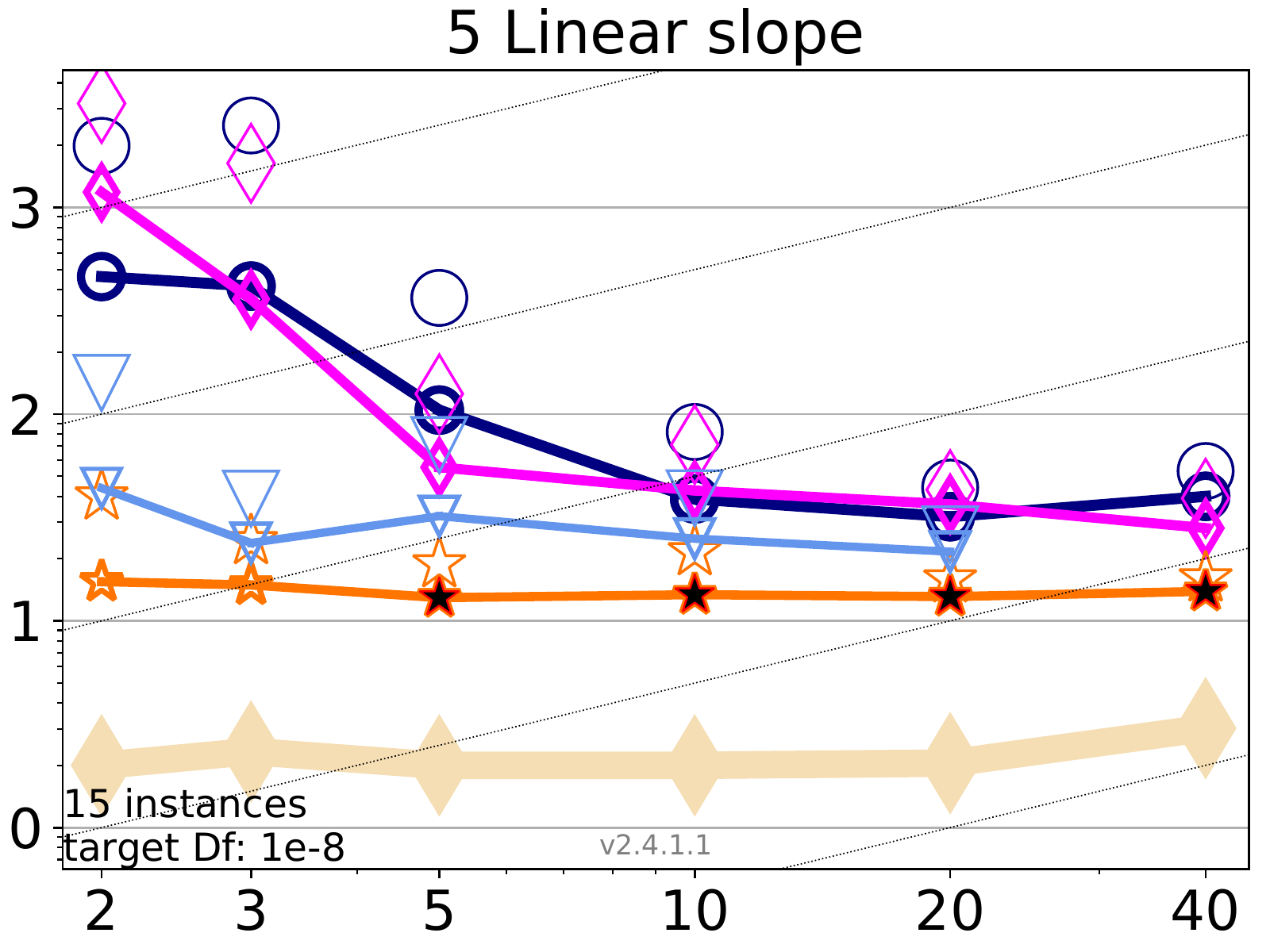}&
		\includegraphics[width=0.24\textwidth]{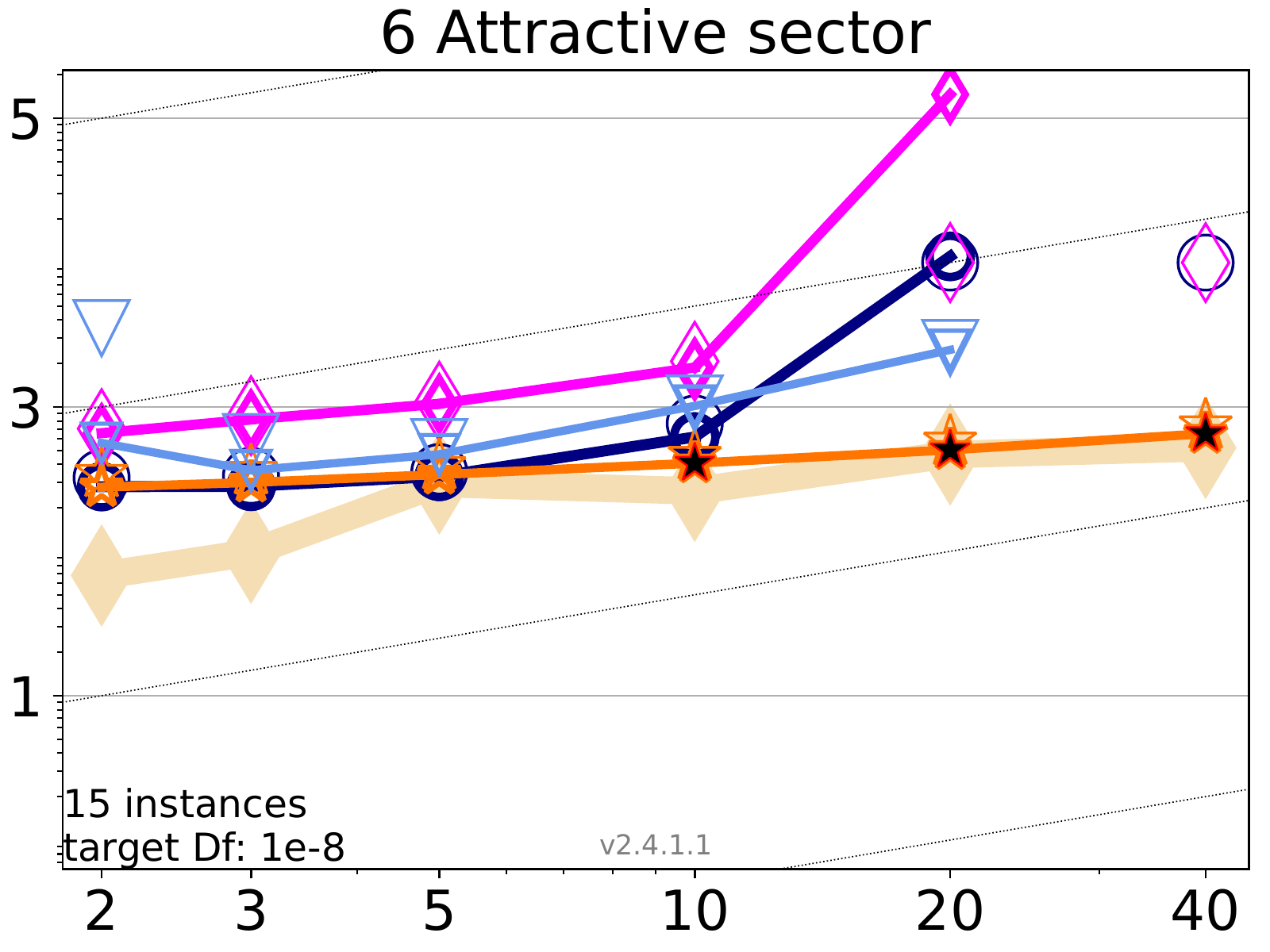}&
		\includegraphics[width=0.24\textwidth]{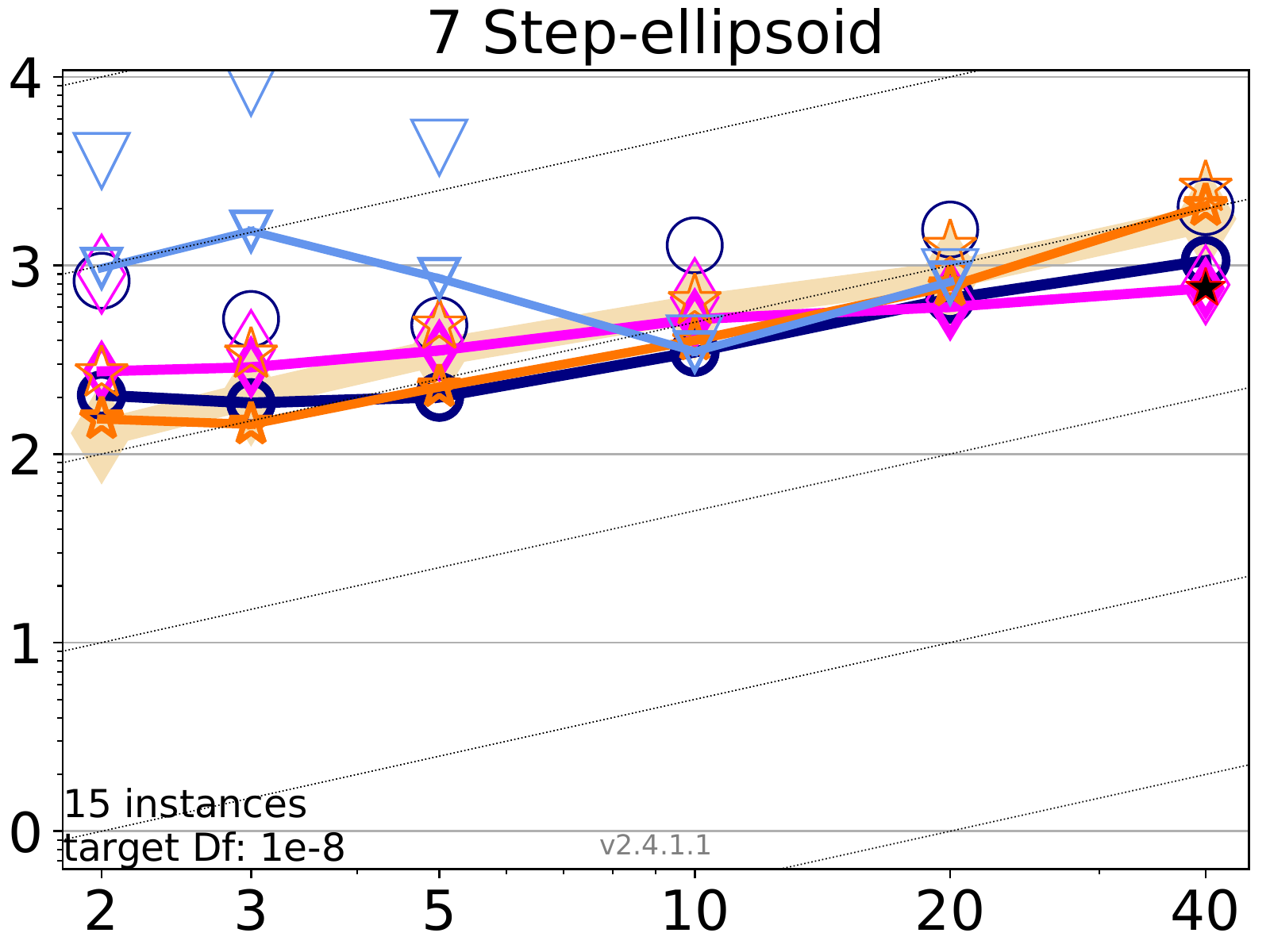}&
		\includegraphics[width=0.24\textwidth]{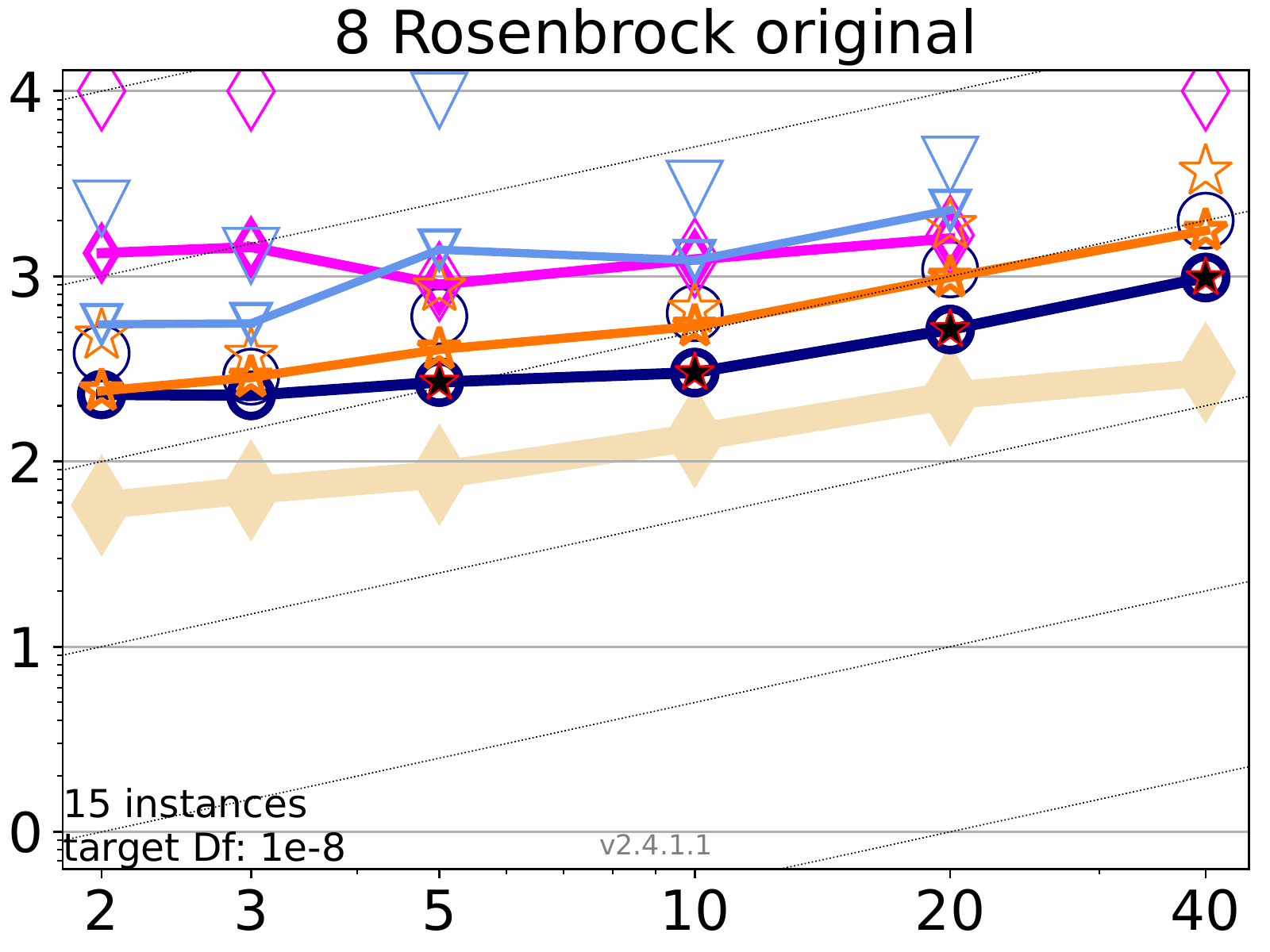}\\[-0.25em]
		\includegraphics[width=0.24\textwidth]{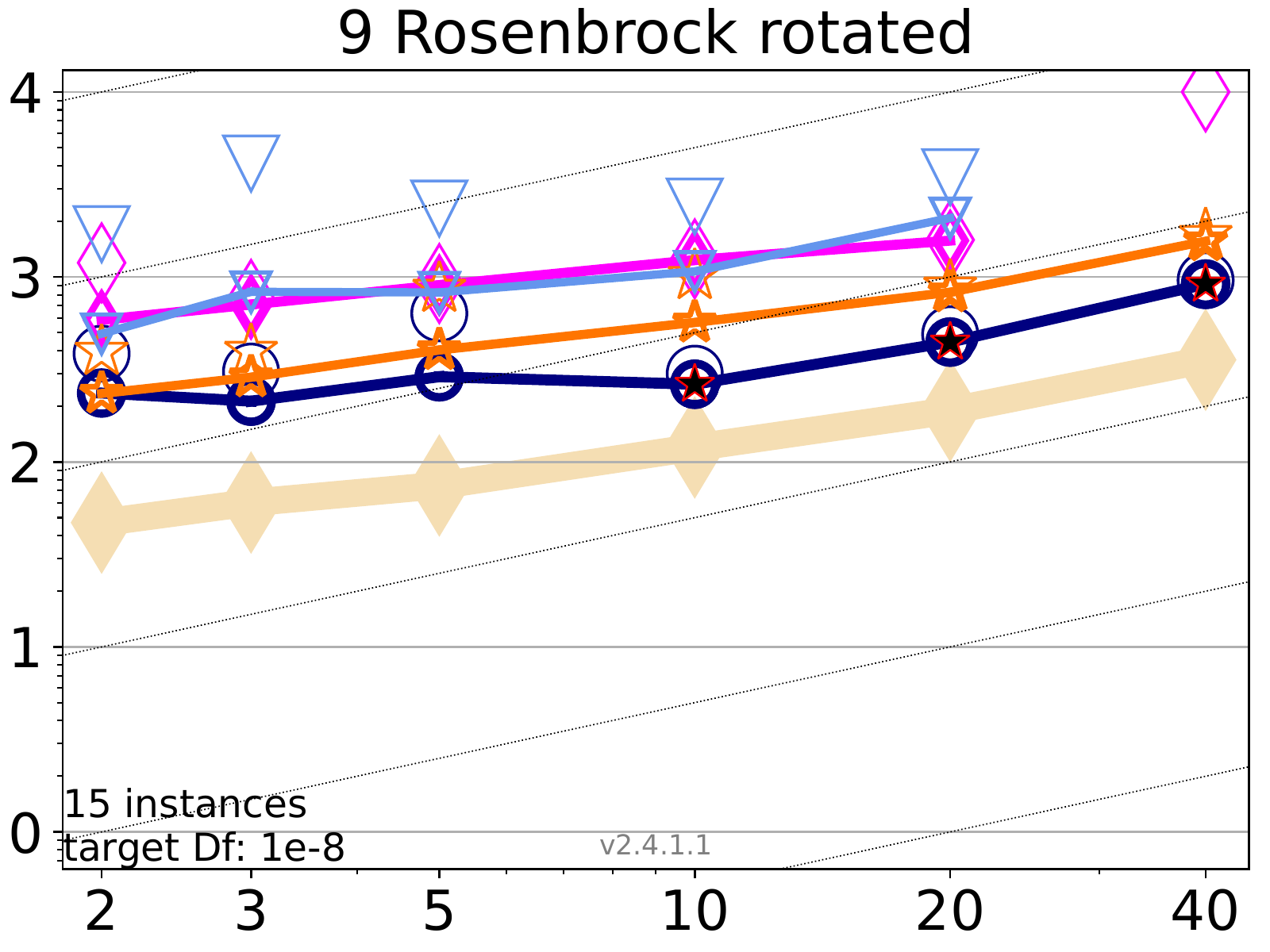}&
		\includegraphics[width=0.24\textwidth]{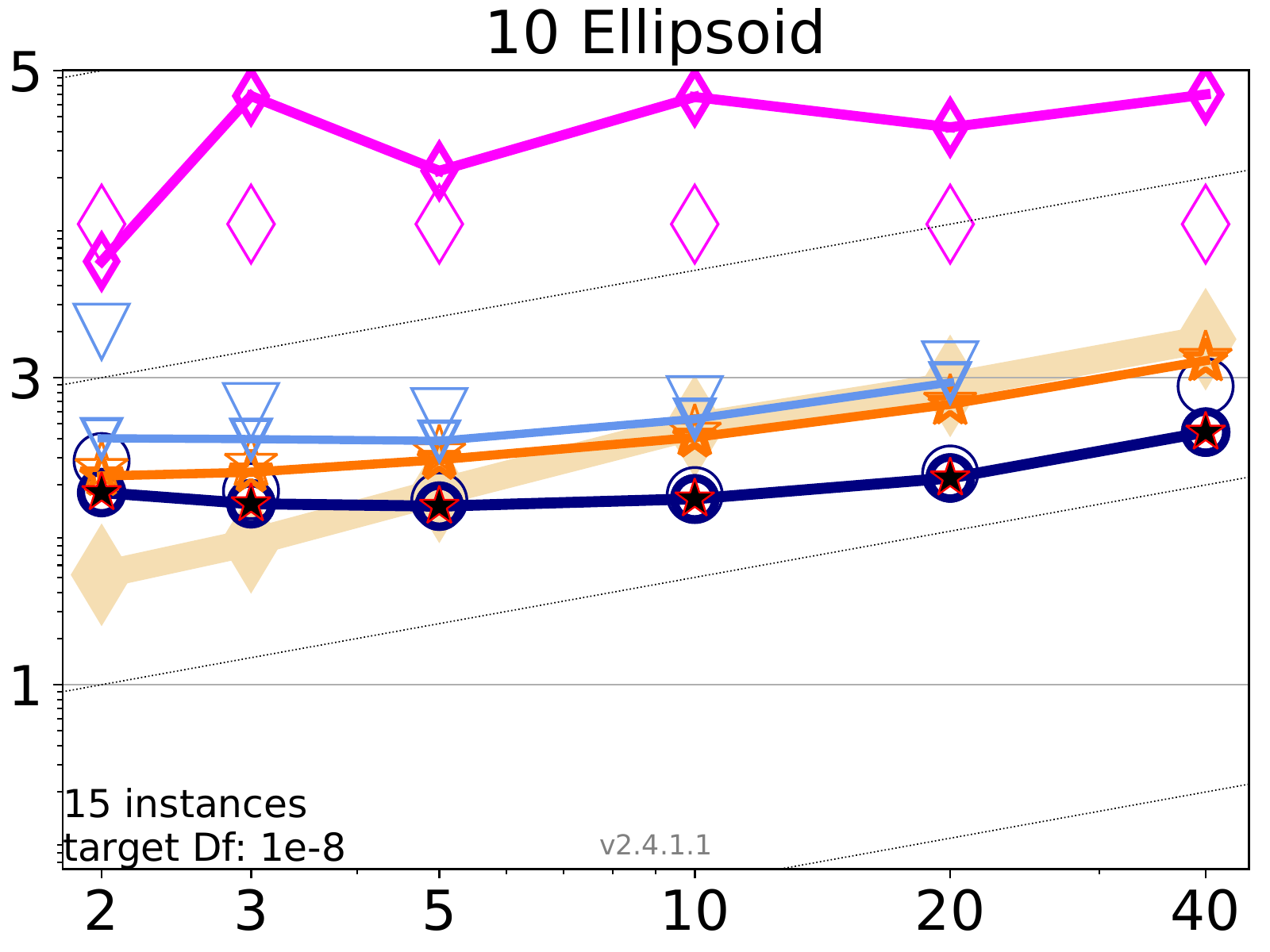}&
		\includegraphics[width=0.24\textwidth]{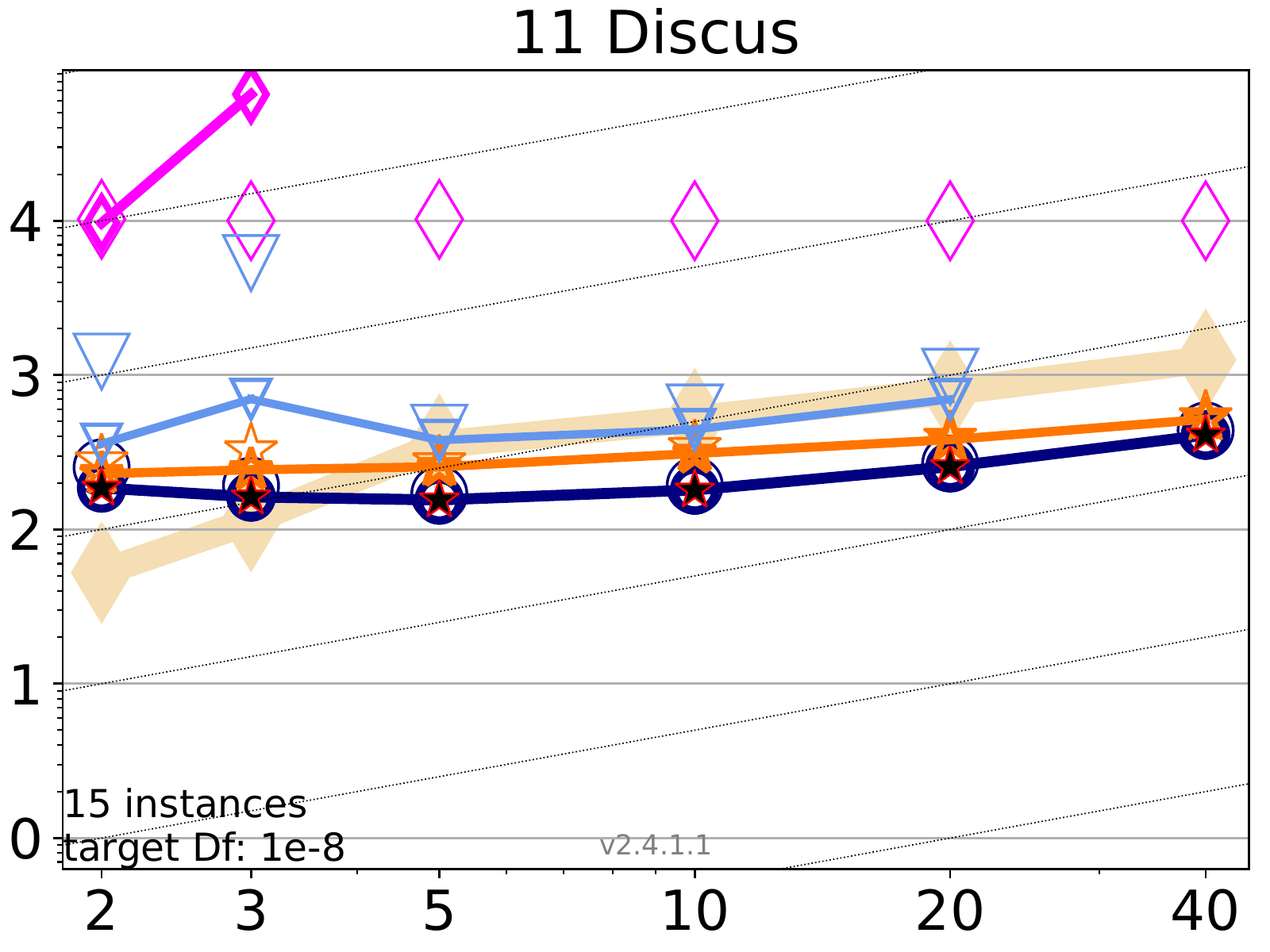}&
		\includegraphics[width=0.24\textwidth]{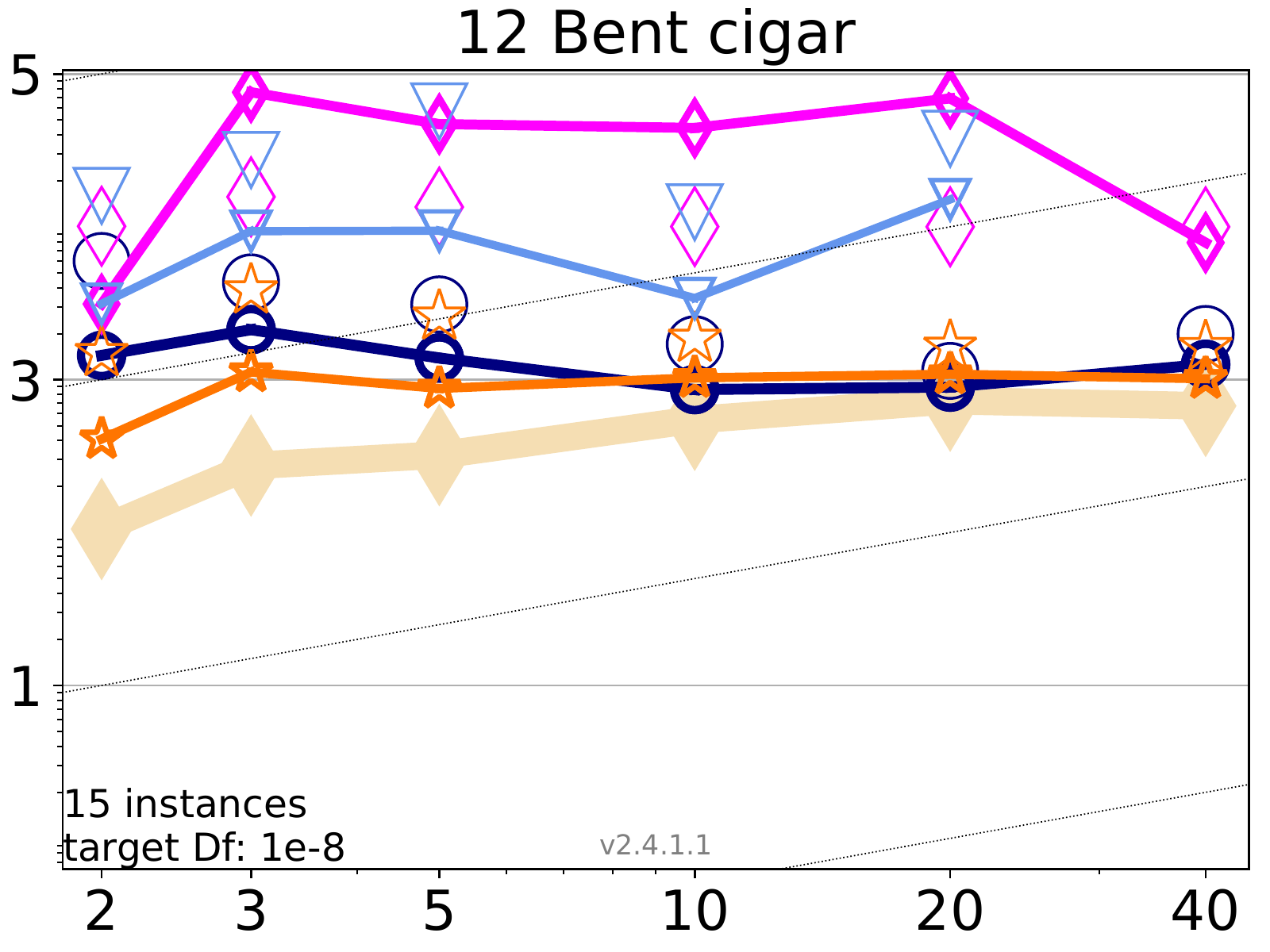}\\[-0.25em]
		\includegraphics[width=0.24\textwidth]{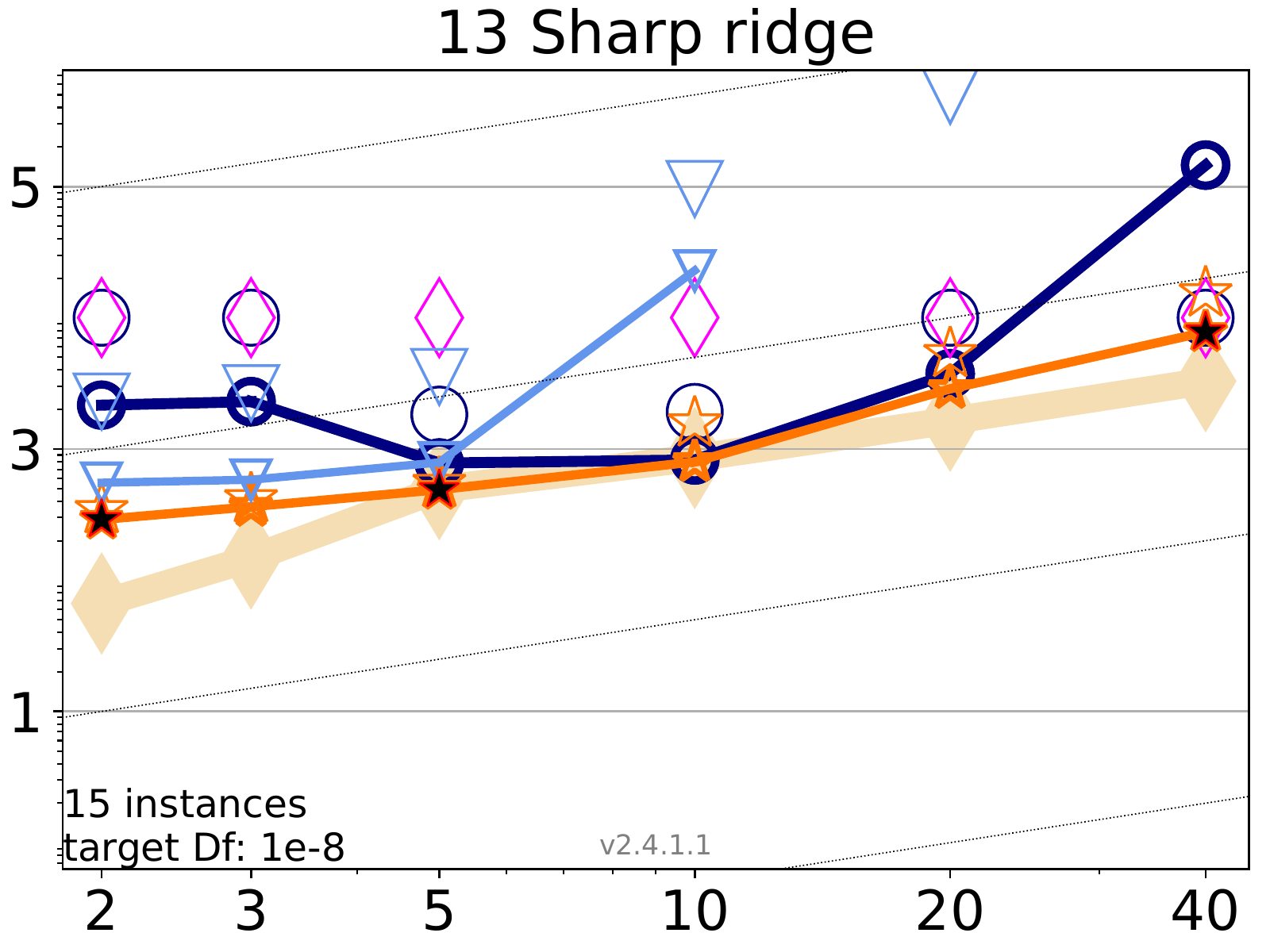}&
		\includegraphics[width=0.24\textwidth]{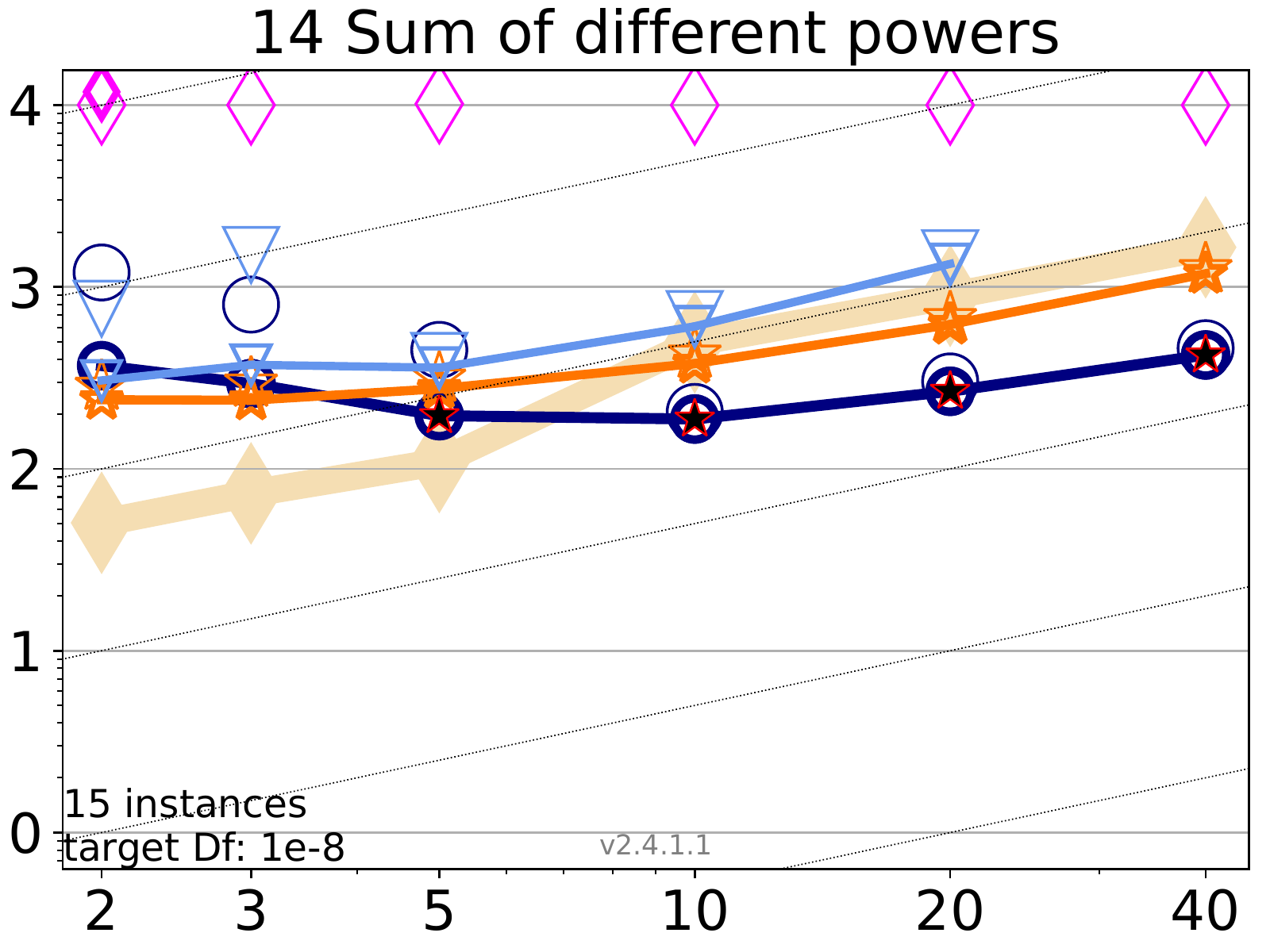}&
		\includegraphics[width=0.24\textwidth]{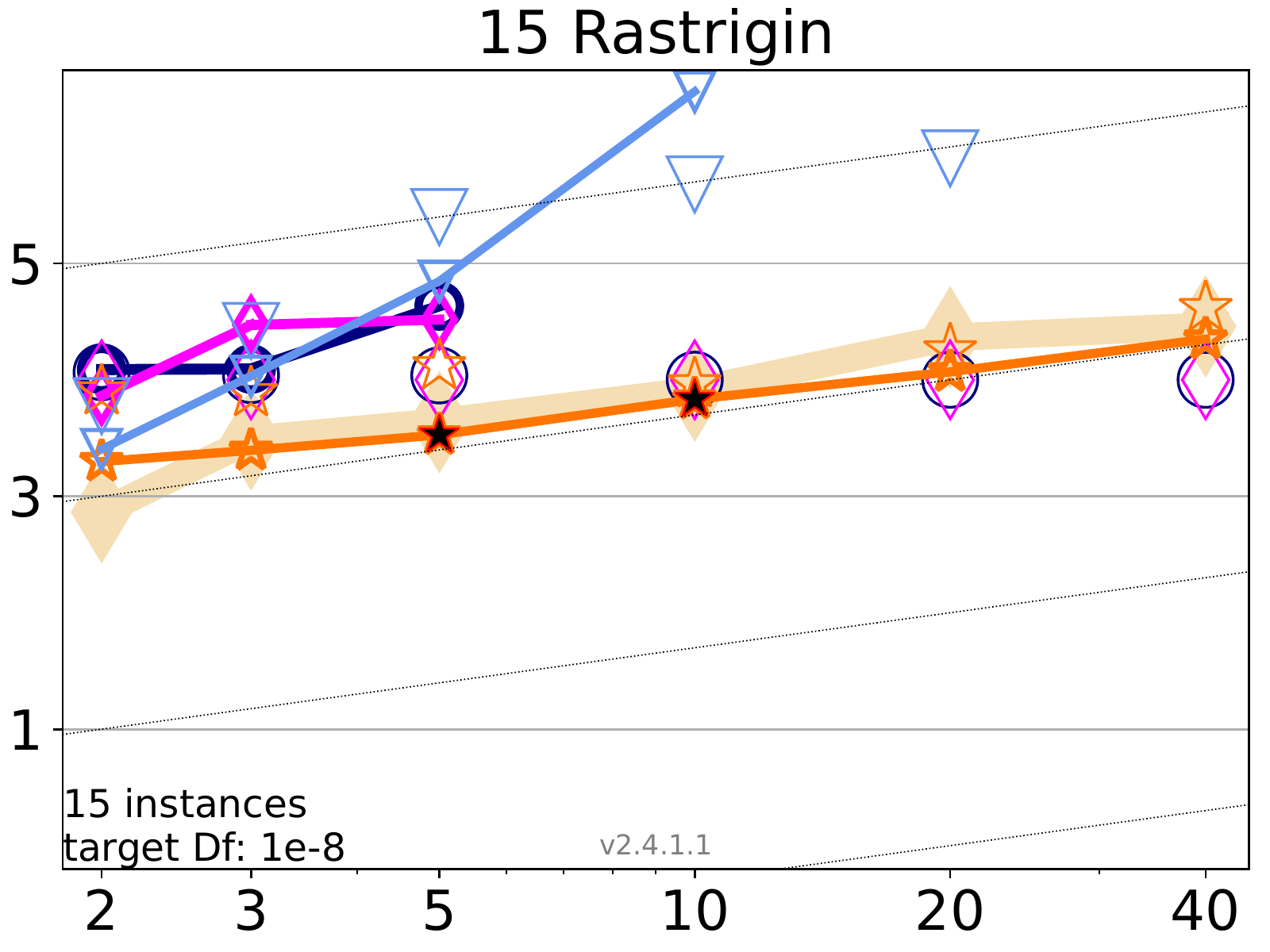}&
		\includegraphics[width=0.24\textwidth]{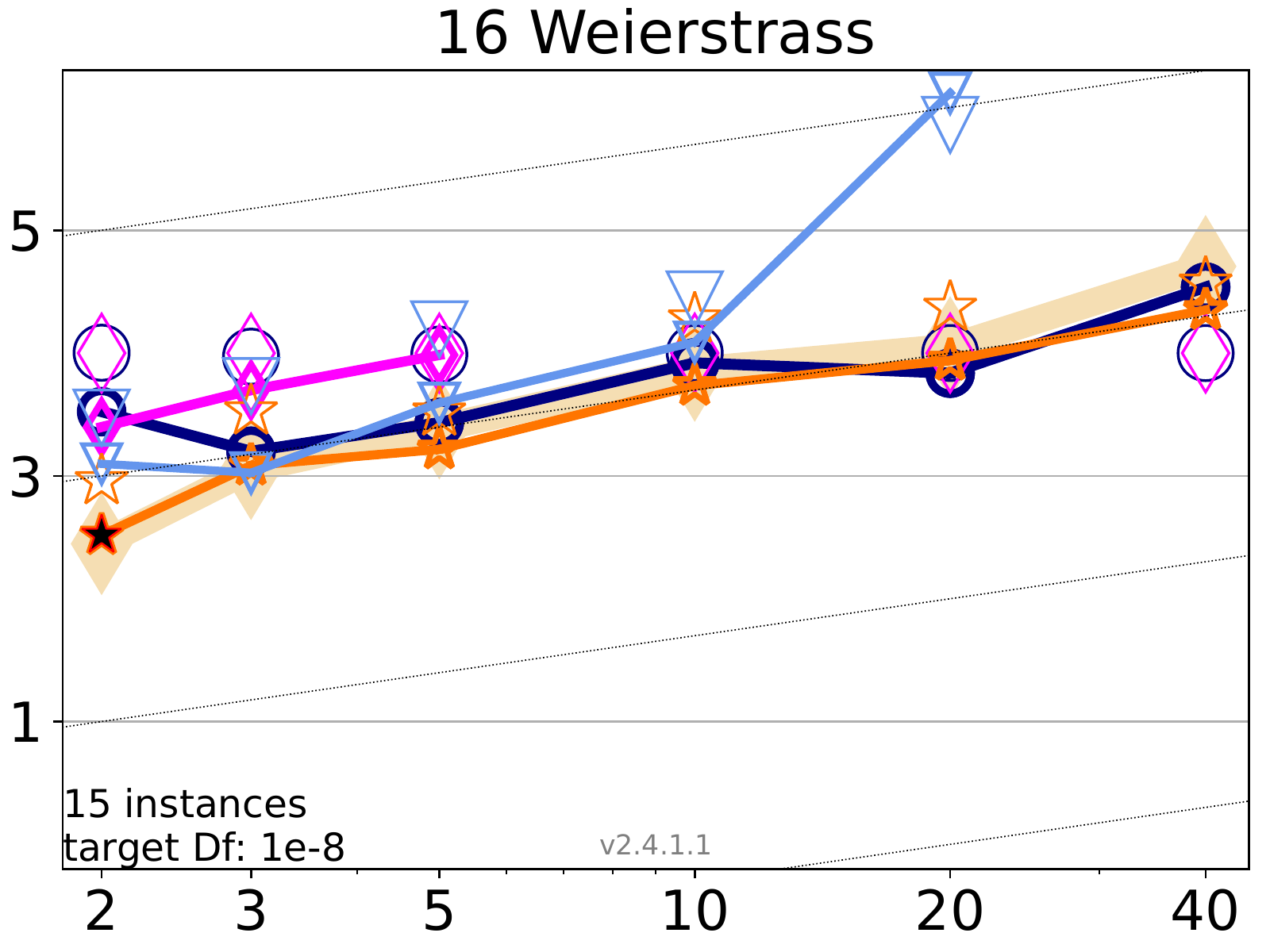}\\[-0.25em]
		\includegraphics[width=0.24\textwidth]{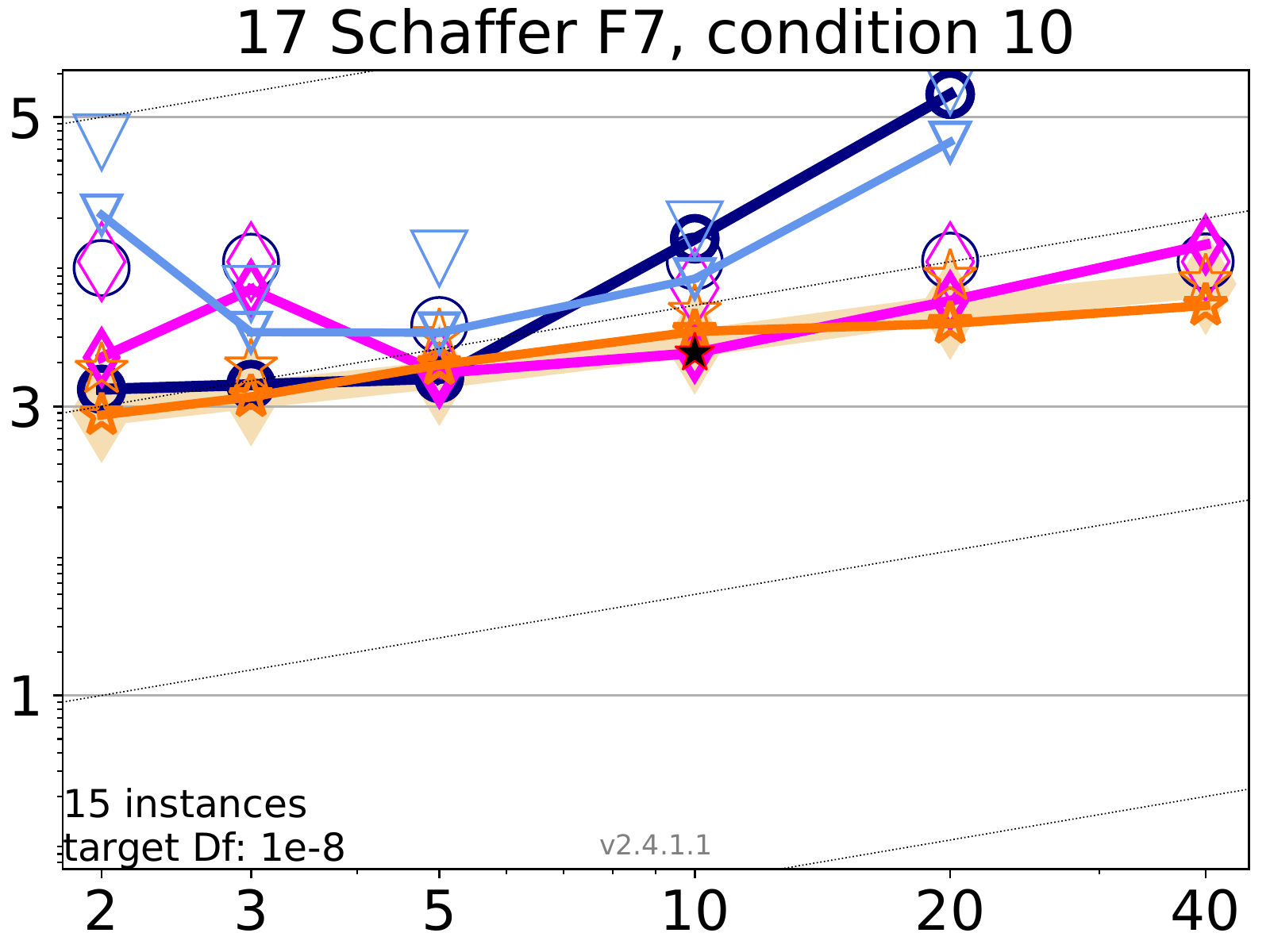}&
		\includegraphics[width=0.24\textwidth]{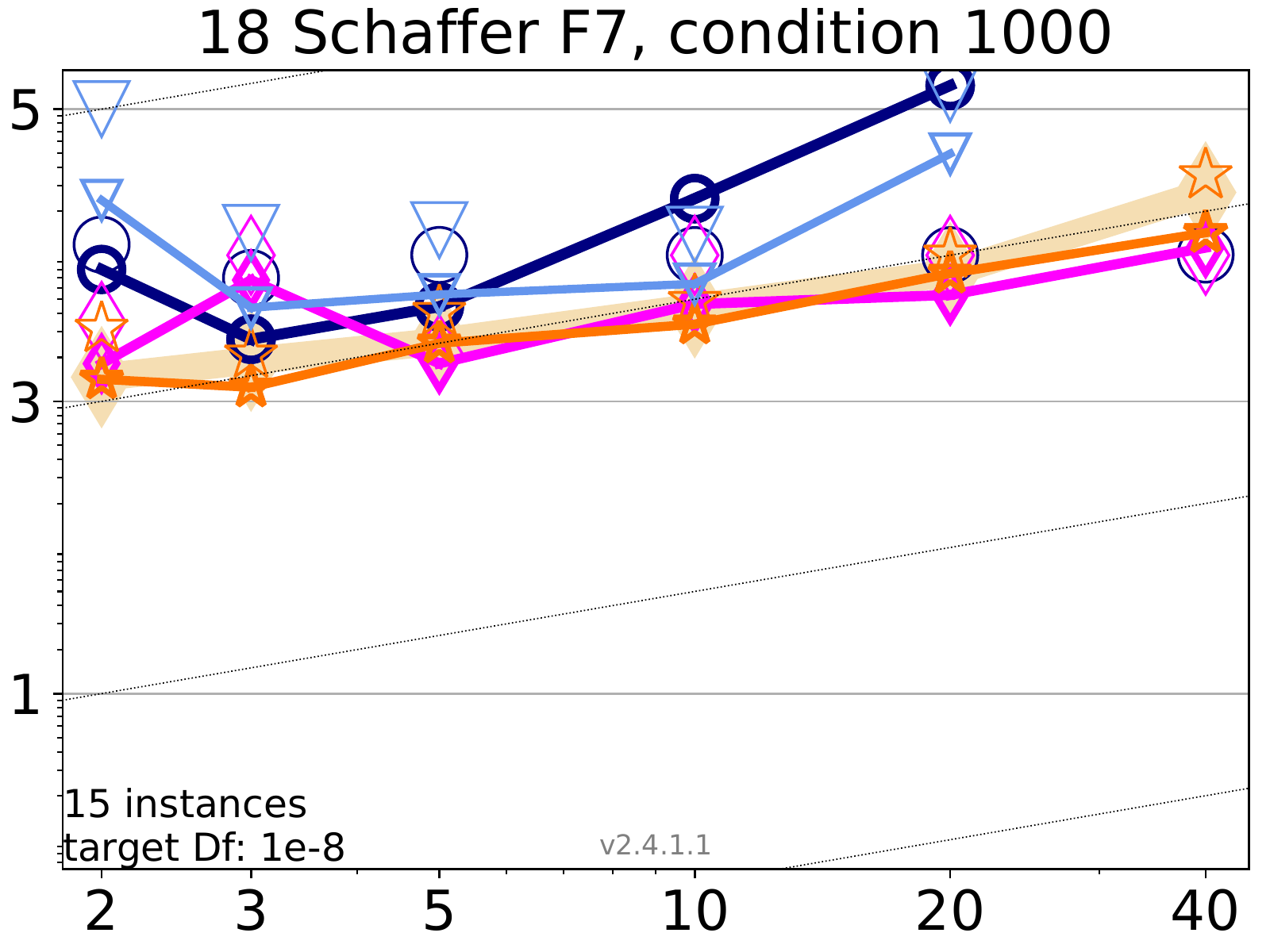}&
		\includegraphics[width=0.24\textwidth]{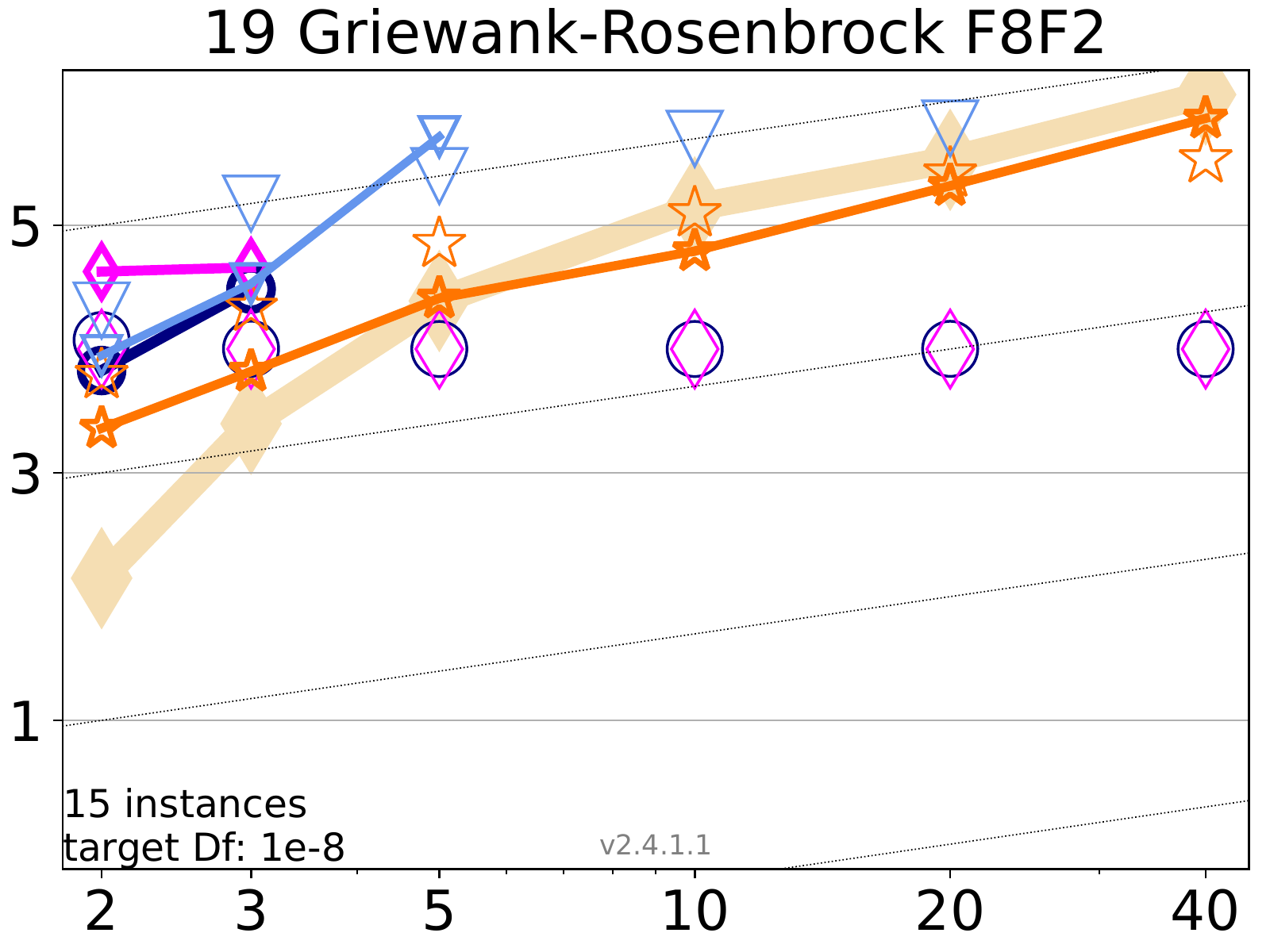}&
		\includegraphics[width=0.24\textwidth]{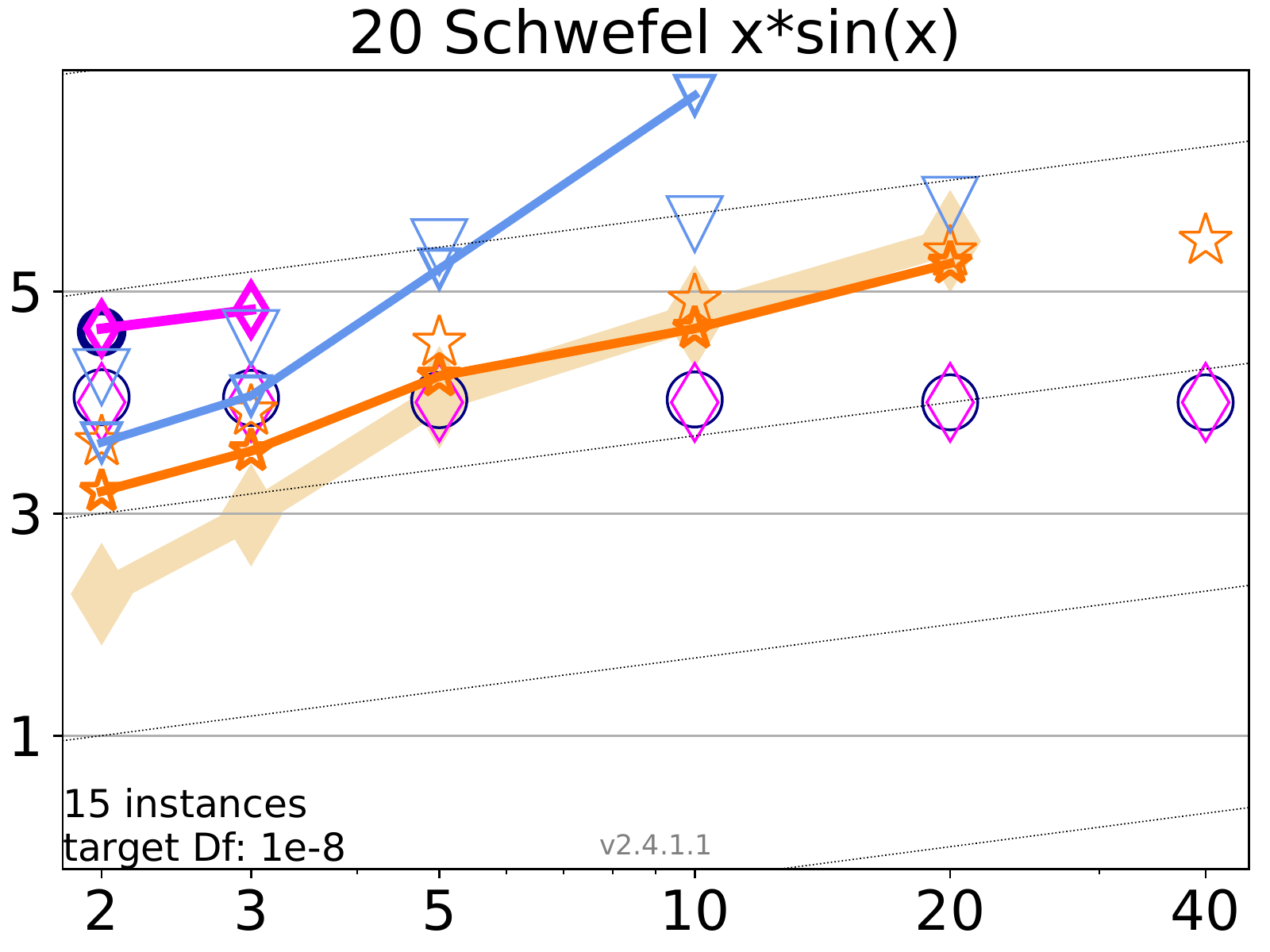}\\[-0.25em]
		\includegraphics[width=0.24\textwidth]{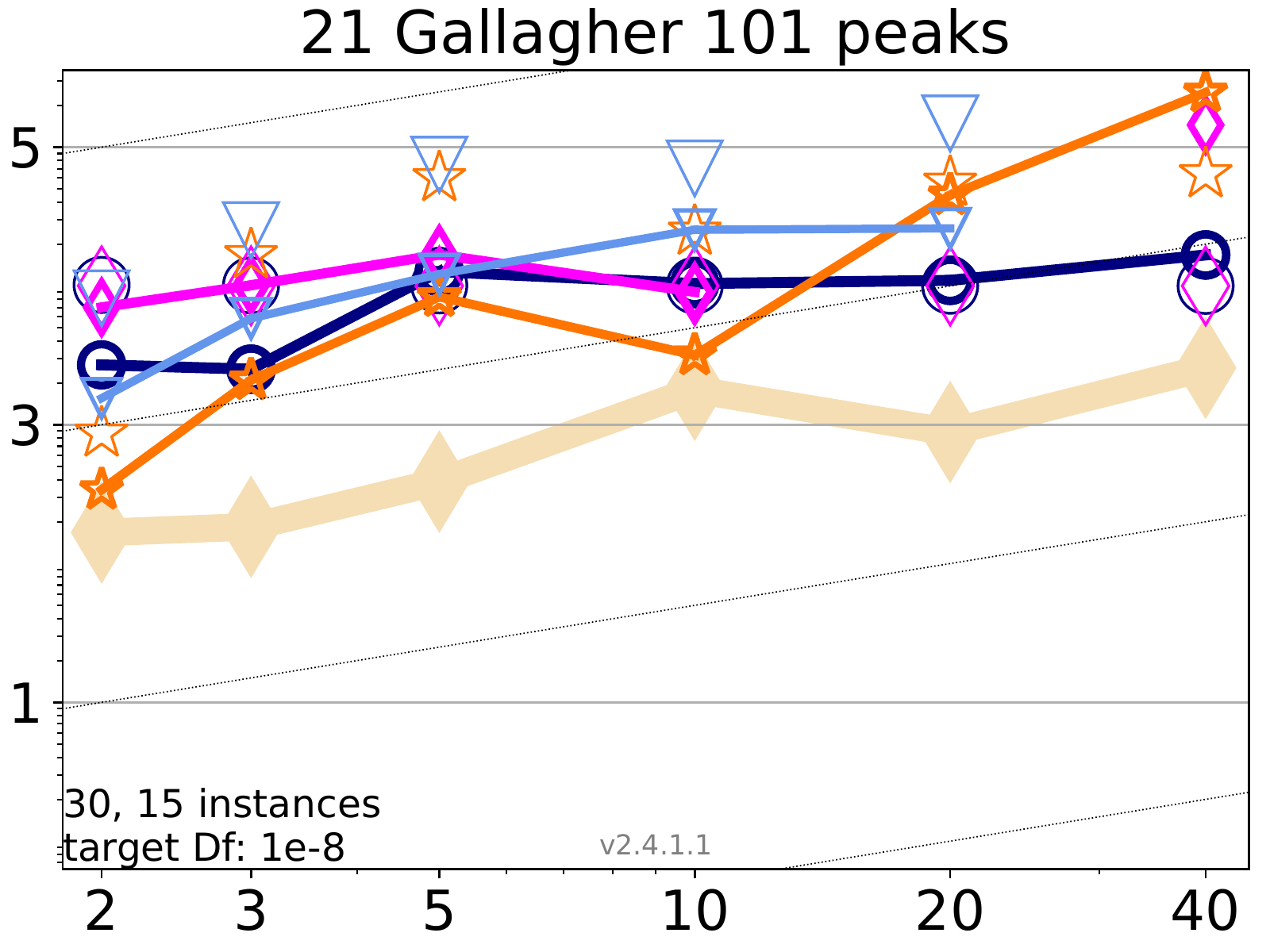}&
		\includegraphics[width=0.24\textwidth]{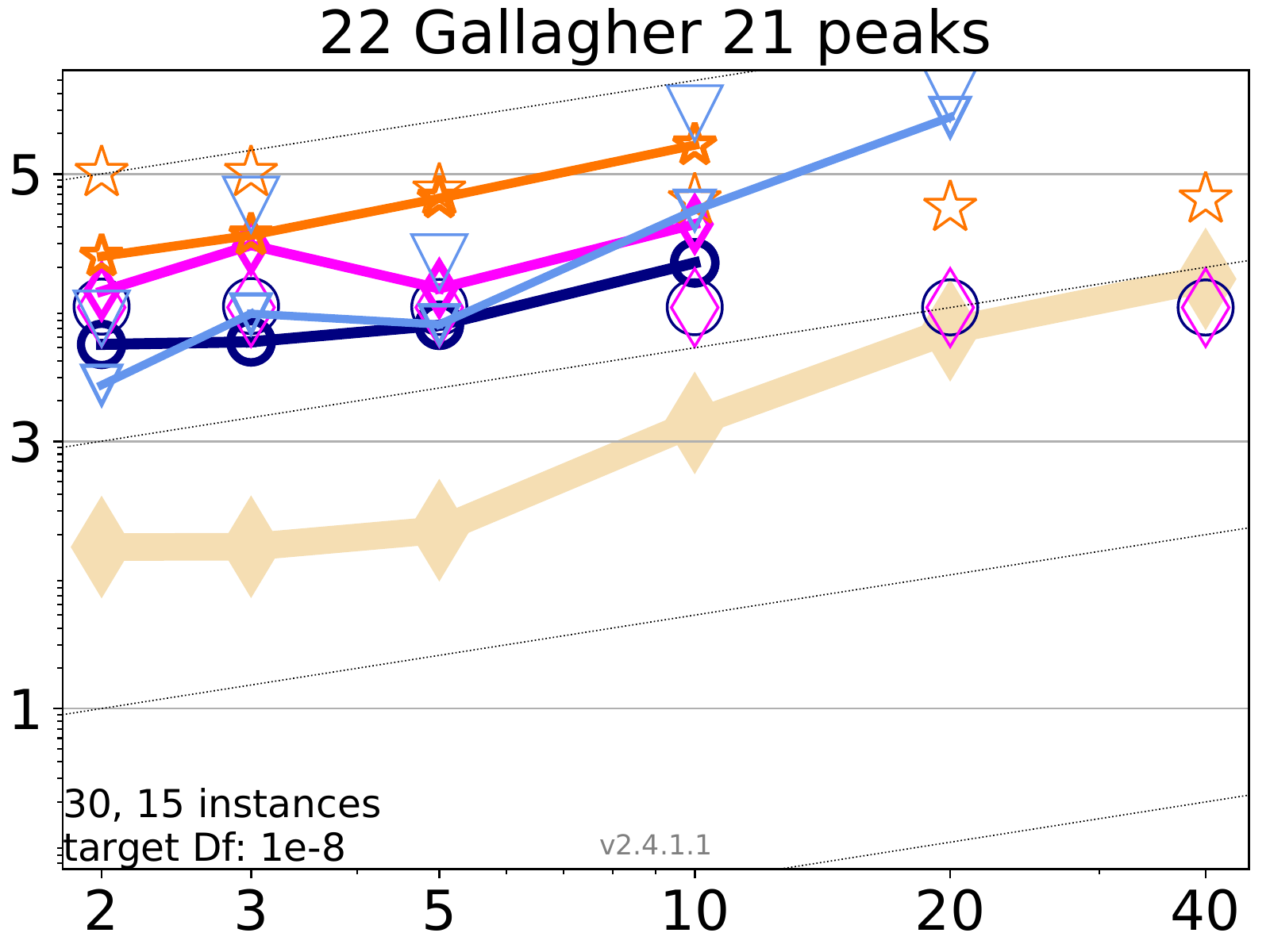}&
		\includegraphics[width=0.24\textwidth]{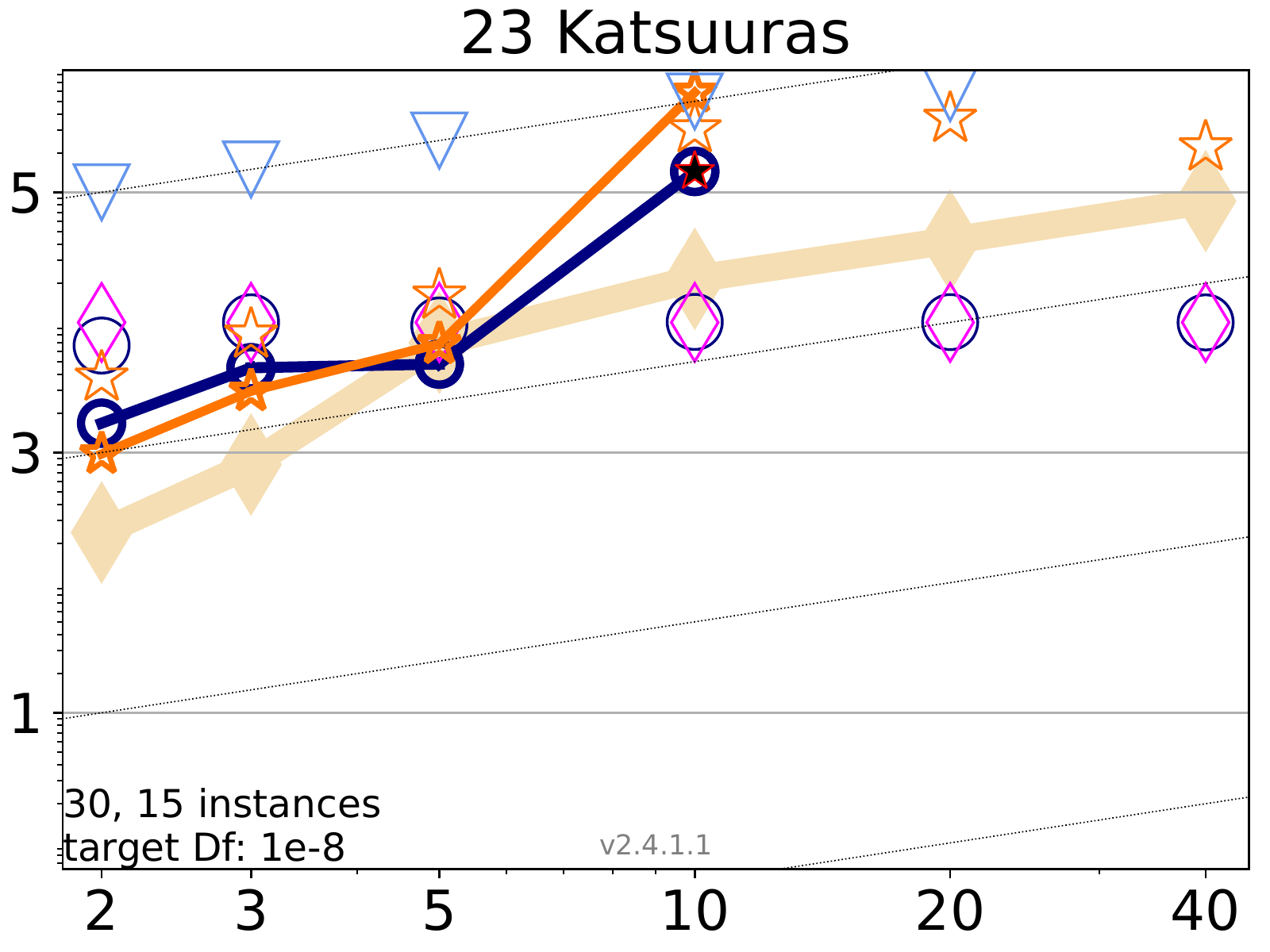}&
		\includegraphics[width=0.24\textwidth]{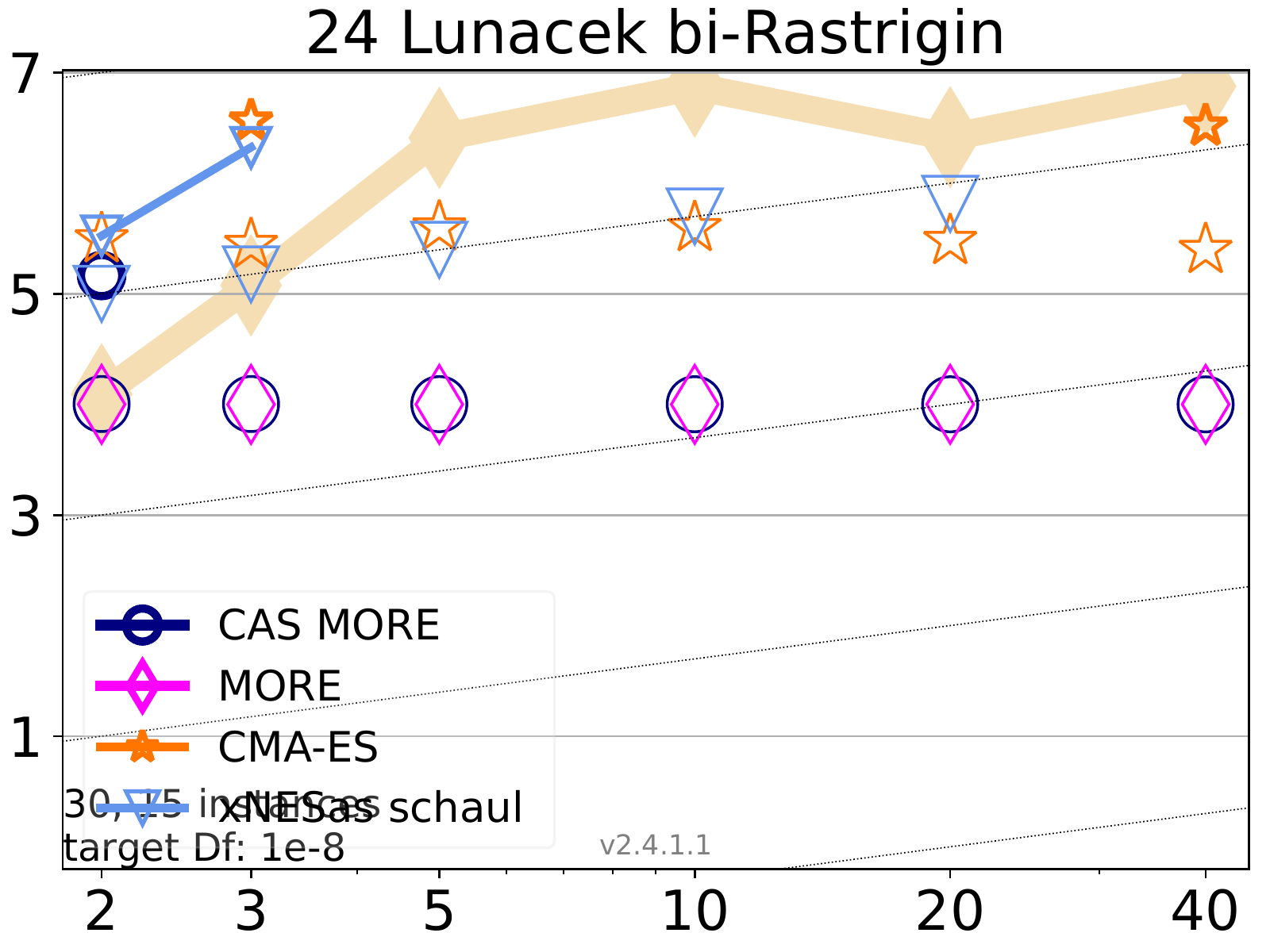}
	\end{tabular}
	\vspace*{-0.2cm}
	\caption[Expected running time (\ERT) divided by dimension
	versus dimension in log-log presentation]{
		\label{fig:scaling}
		\bbobppfigslegend{$f_1$ and $f_{24}$}. 
	}
\end{figure*}

\begin{figure*}
	\begin{tabular}{c@{\hspace*{0.01\textwidth}}c@{\hspace*{0.01\textwidth}}c}
		{\sffamily separable fcts}\hspace{1cm} & {\sffamily moderate fcts}\hspace{1cm} & \hspace{-1cm}{\sffamily ill-conditioned fcts}\\
		\includegraphics[width=0.32\textwidth]{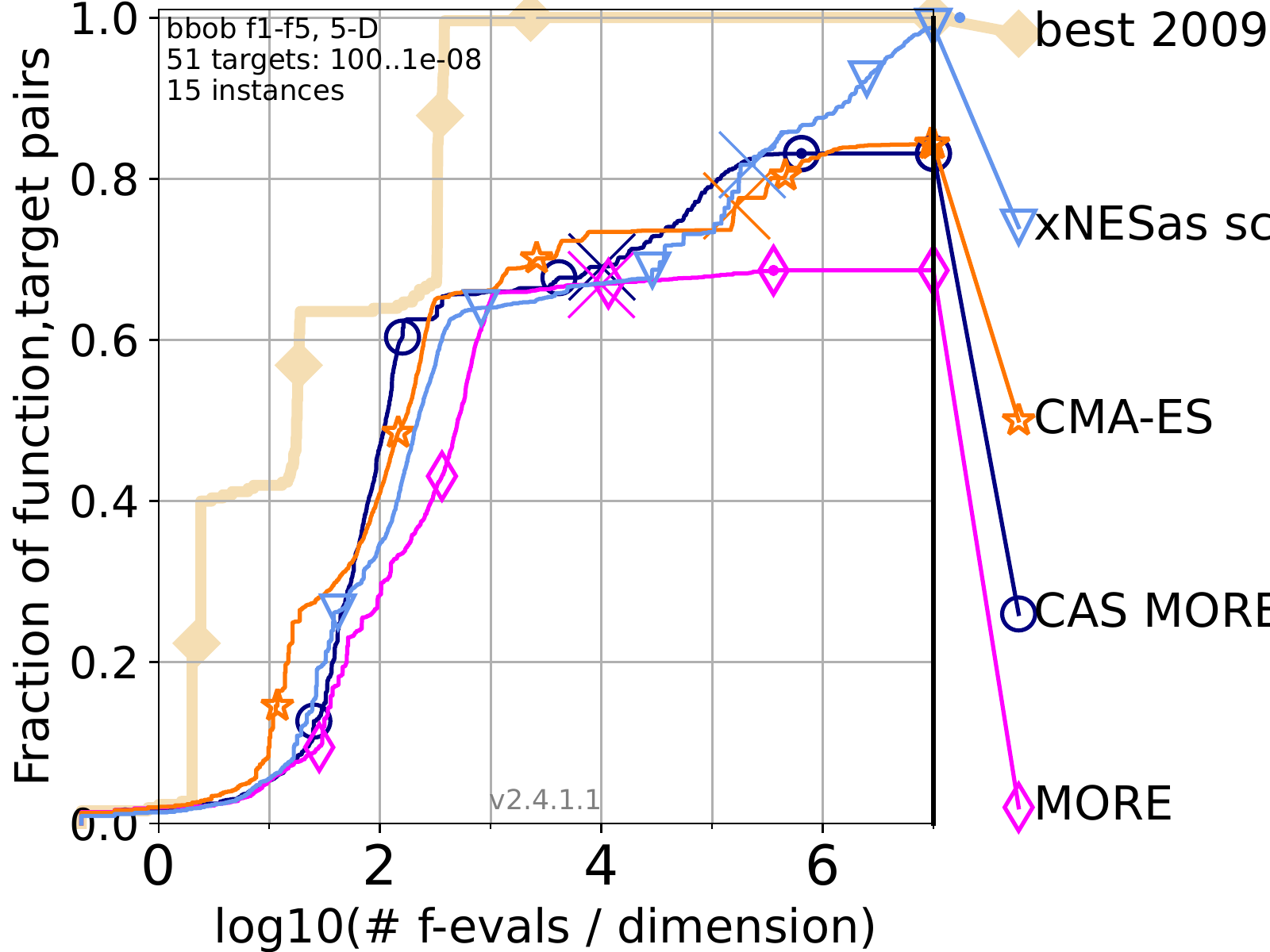}&
		\includegraphics[width=0.32\textwidth]{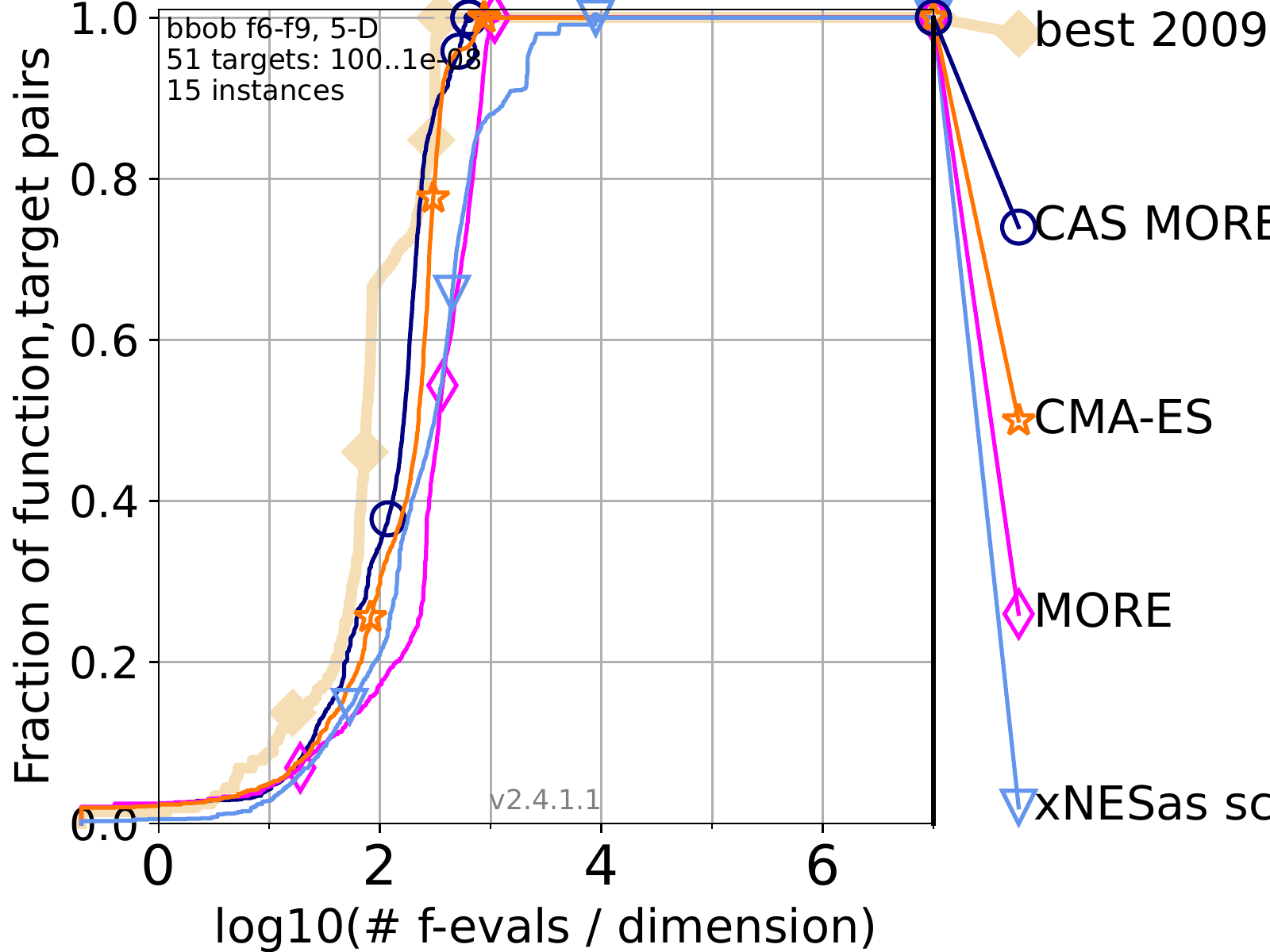}&
		\includegraphics[width=0.32\textwidth]{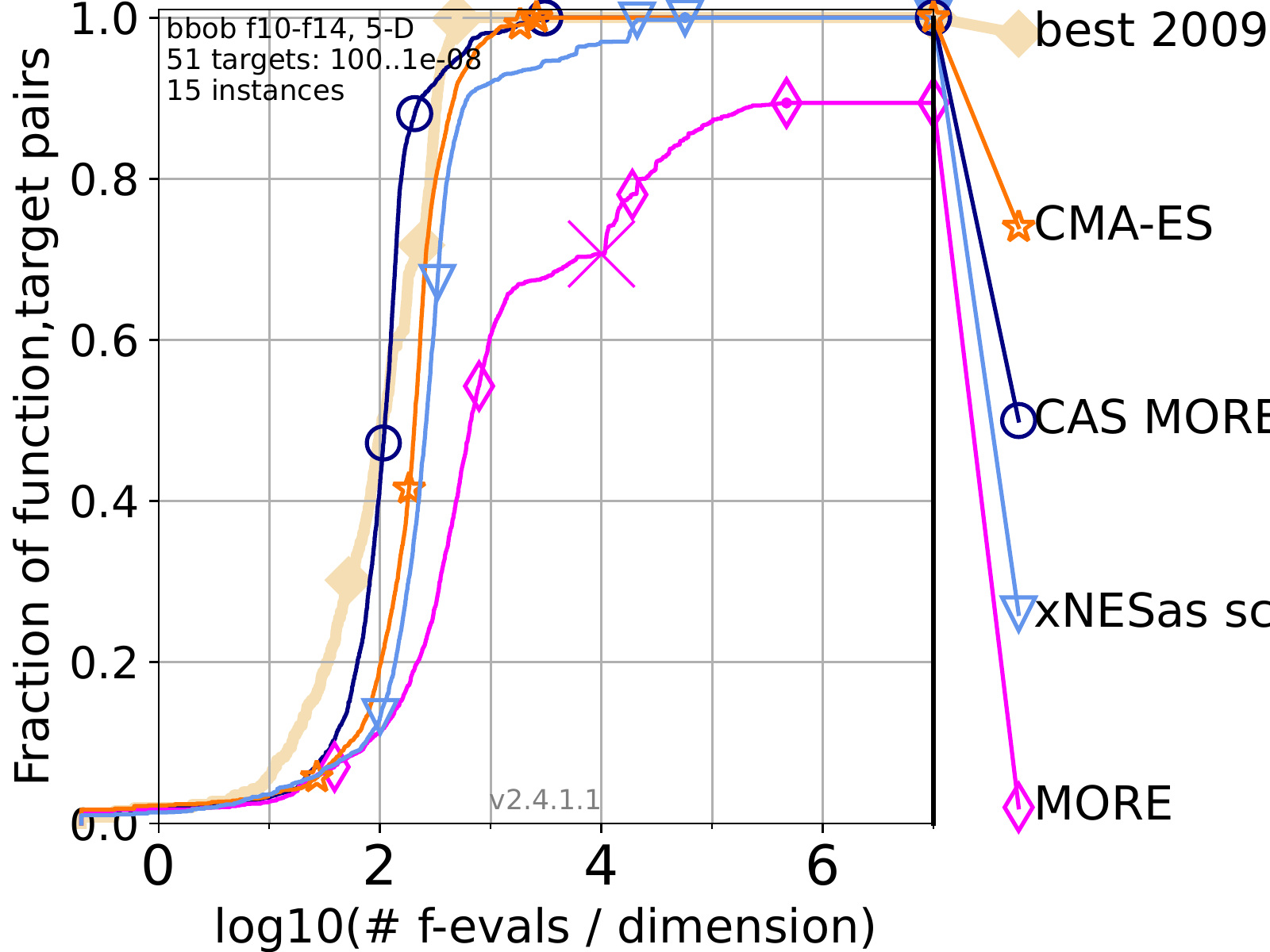}\\[-0.2em]
		{\sffamily multi-modal fcts}\hspace{1cm} & {\sffamily weakly structured multi-modal fcts}\hspace{1cm} & \hspace{-1cm}{\sffamily all fcts}\\
		\includegraphics[width=0.32\textwidth]{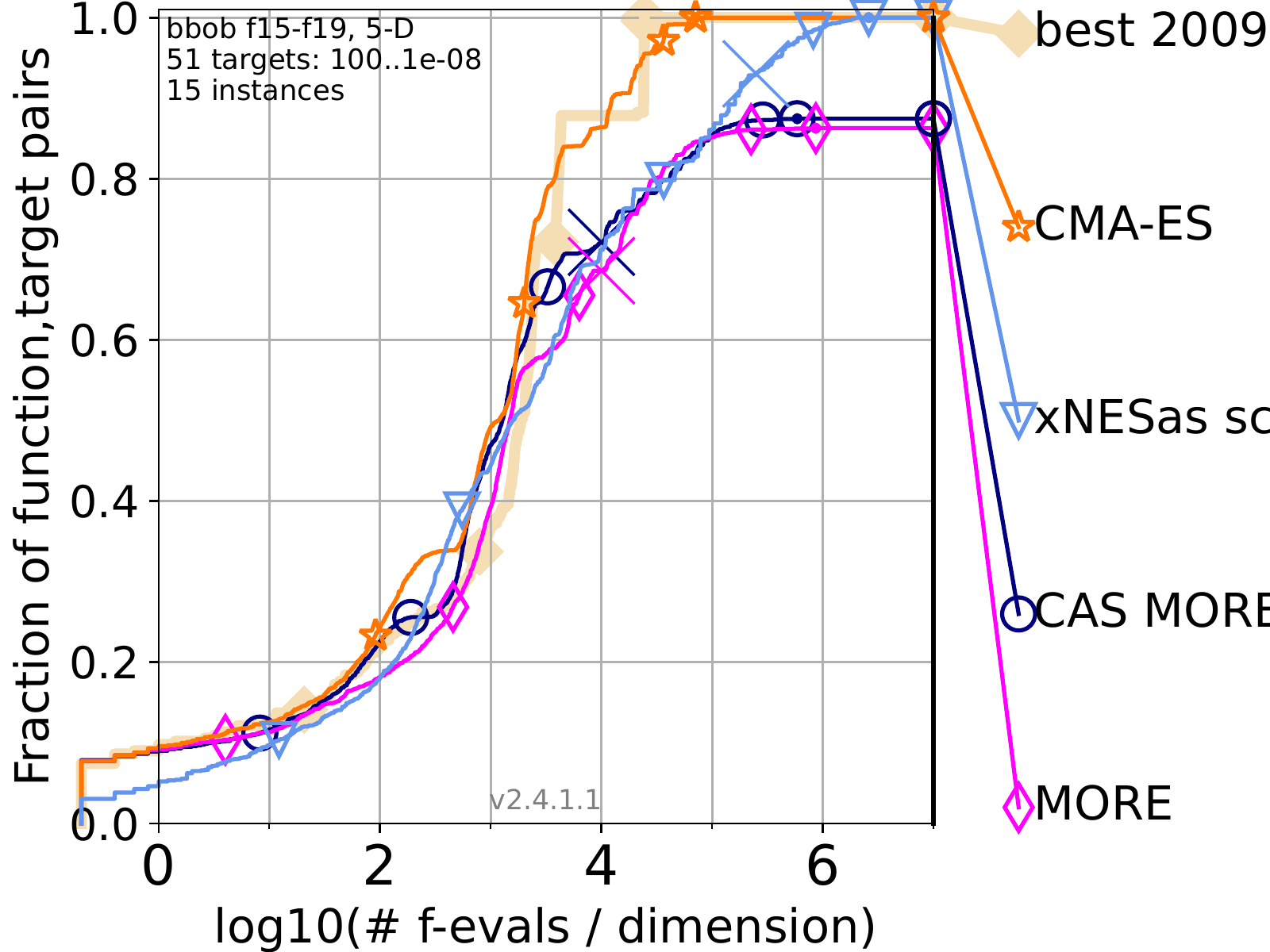}&
		\includegraphics[width=0.32\textwidth]{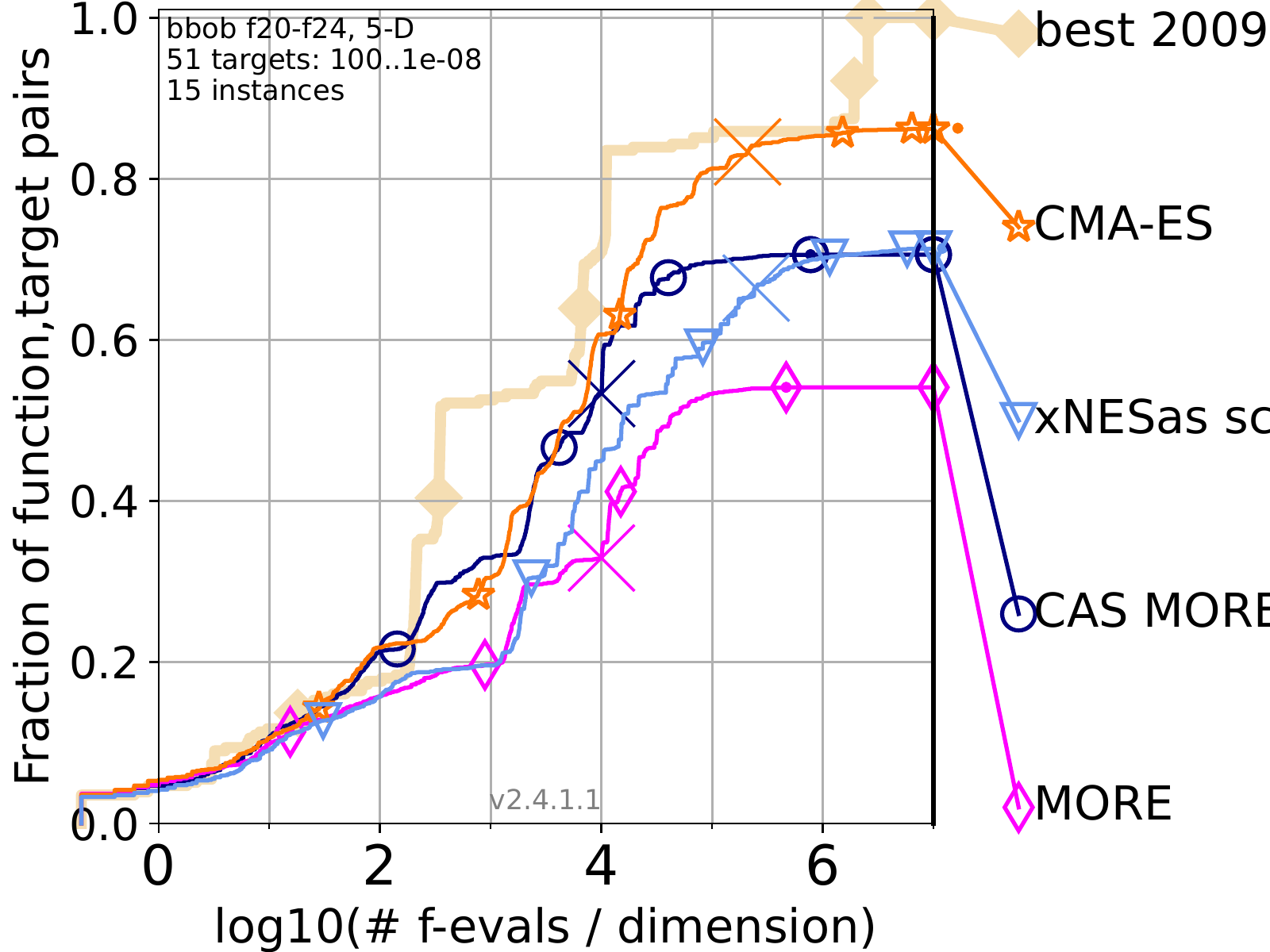}&
		\includegraphics[width=0.32\textwidth]{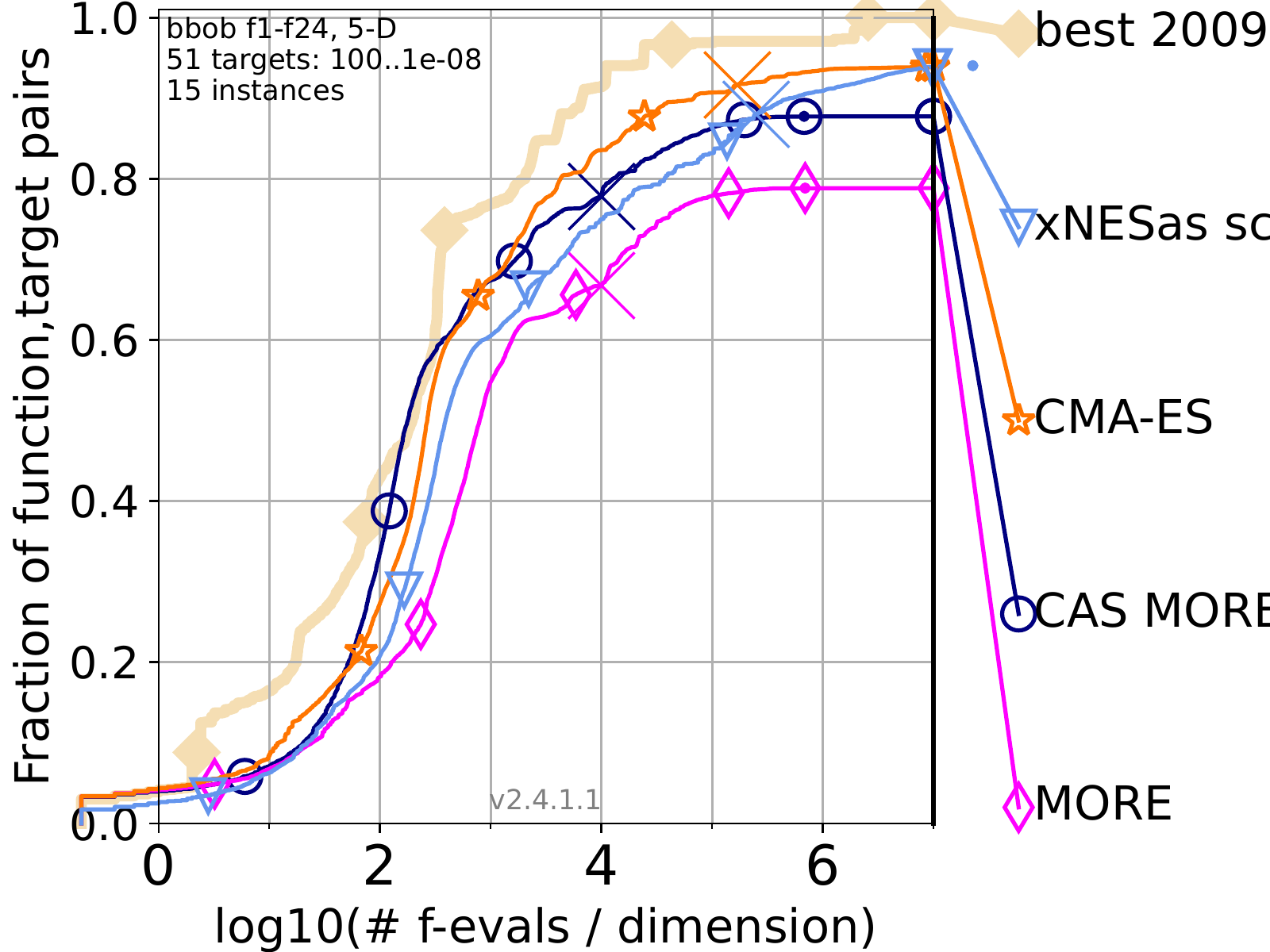}
		\vspace*{-1ex}
	\end{tabular}
	\caption{
		\label{fig:ECDFs05D}
		\bbobECDFslegend{5}
	}
\end{figure*}

\begin{figure*}
	\begin{tabular}{c@{\hspace*{0.01\textwidth}}c@{\hspace*{0.01\textwidth}}c}
		{\sffamily separable fcts}\hspace{1cm} & {\sffamily moderate fcts}\hspace{1cm} & \hspace{-1cm}{\sffamily ill-conditioned fcts}\\
		\includegraphics[width=0.32\textwidth]{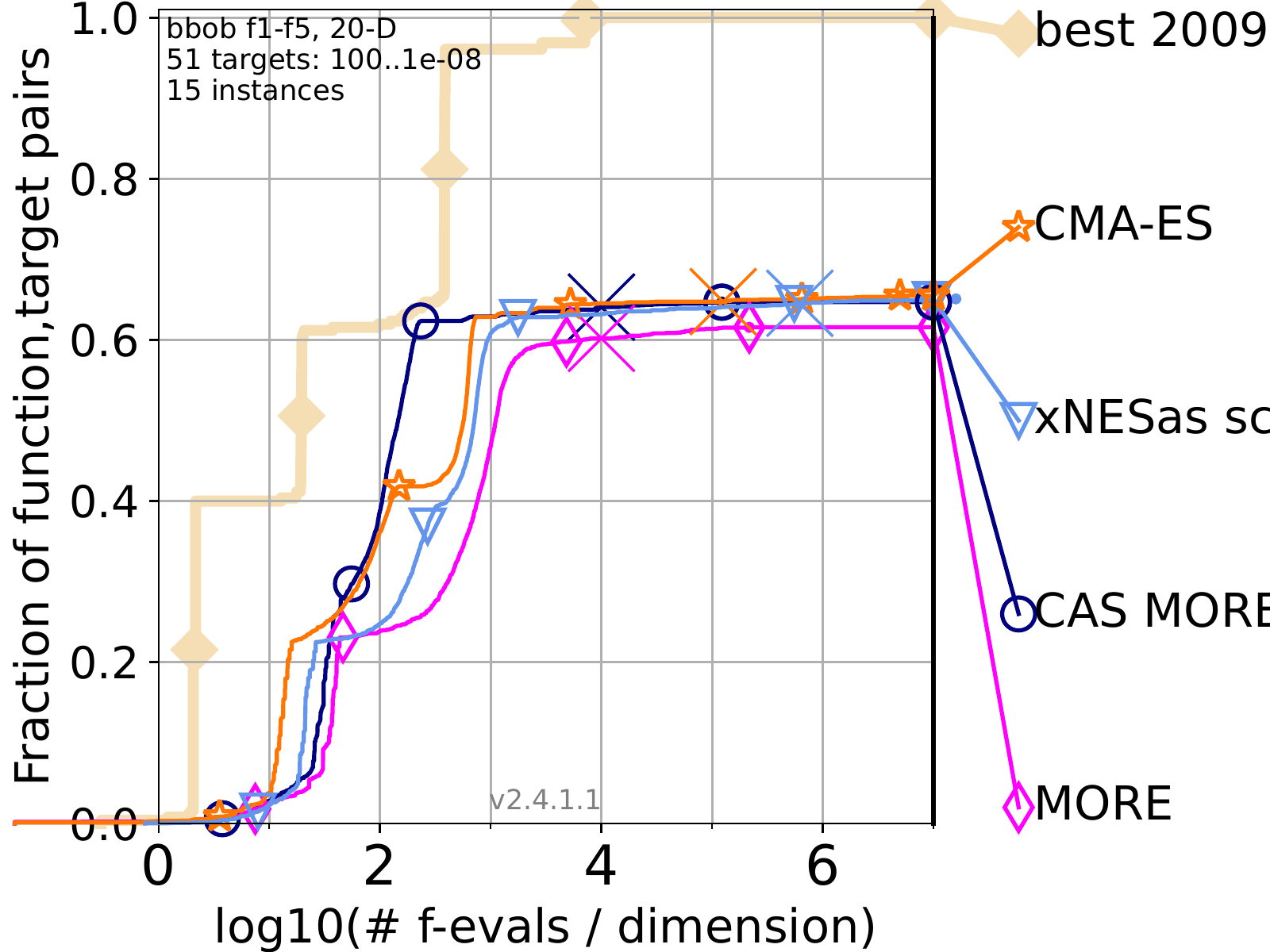}&
		\includegraphics[width=0.32\textwidth]{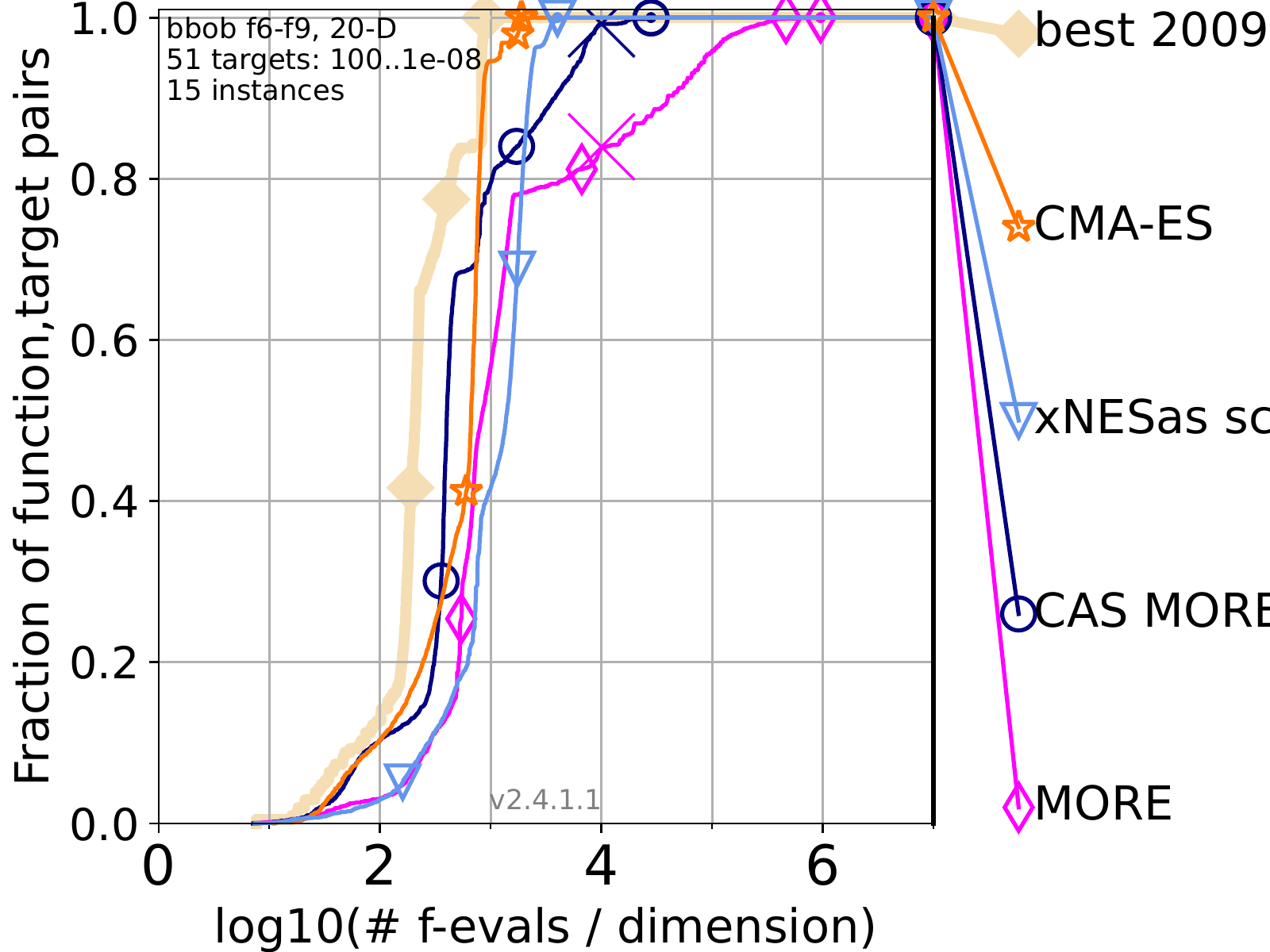}&
		\includegraphics[width=0.32\textwidth]{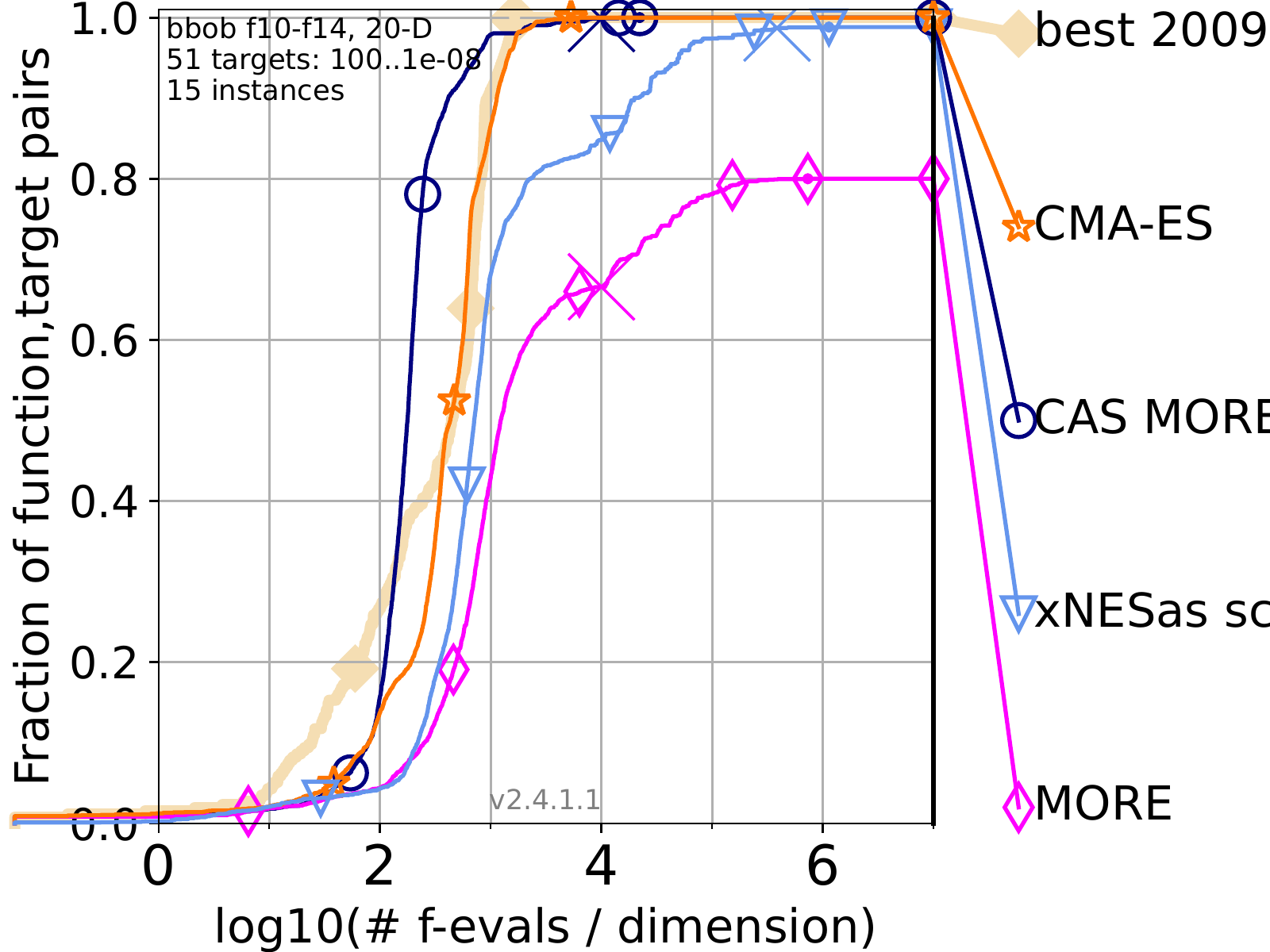}\\[-0.2em]
		{\sffamily multi-modal fcts}\hspace{1cm} & {\sffamily weakly structured multi-modal fcts}\hspace{1cm} & \hspace{-1cm}{\sffamily all fcts}\\
		\includegraphics[width=0.32\textwidth]{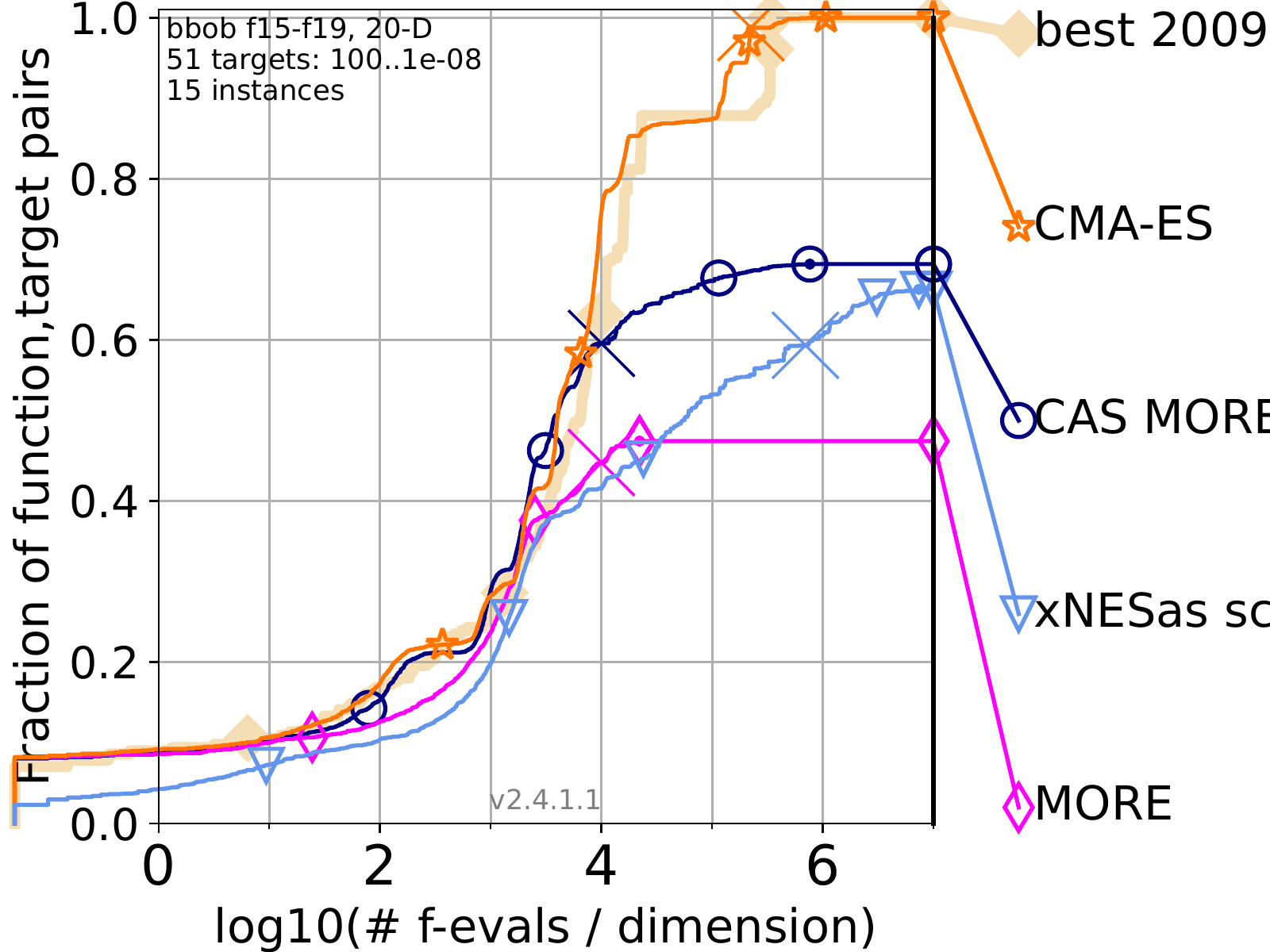}&
		\includegraphics[width=0.32\textwidth]{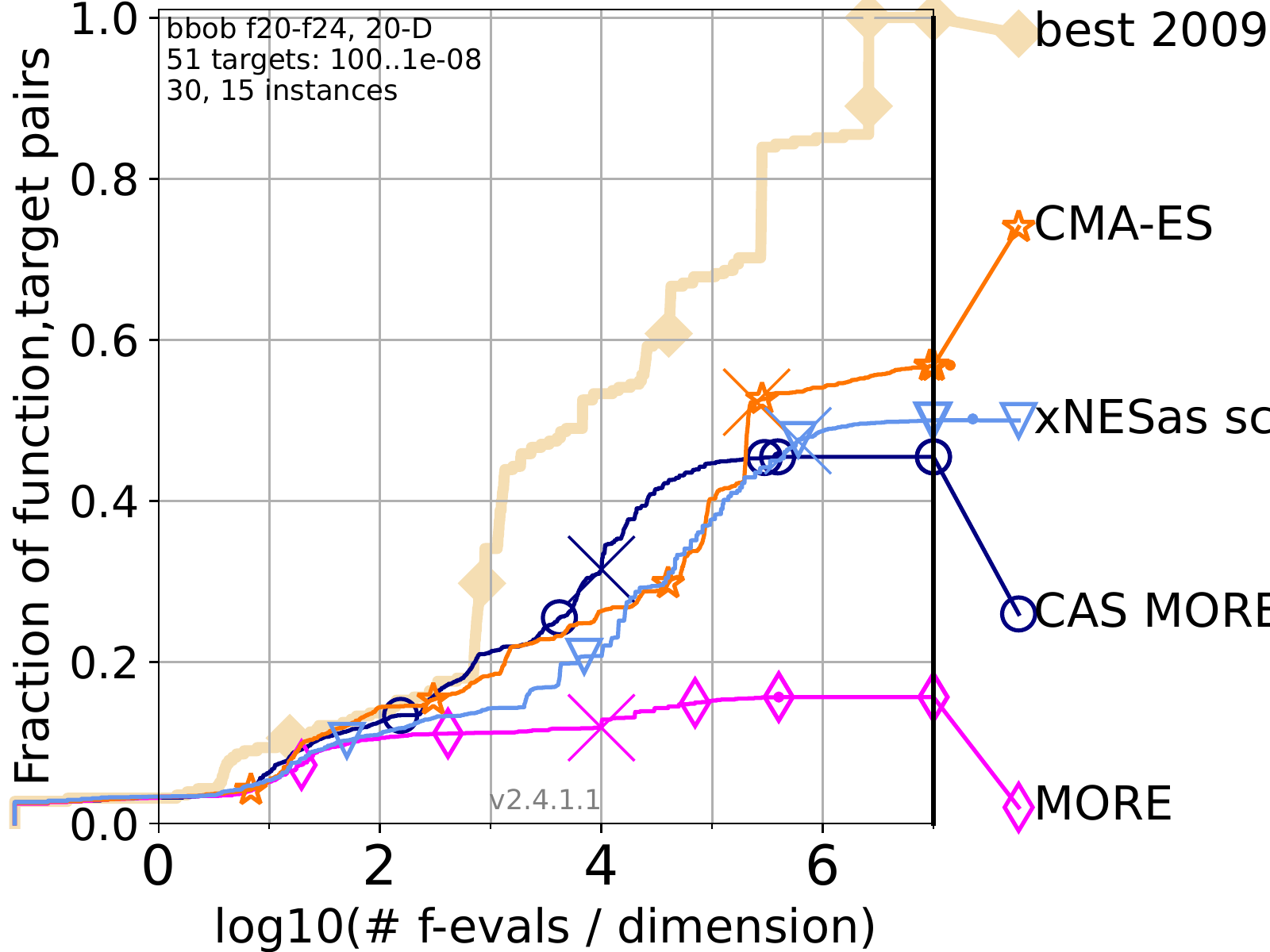}&
		\includegraphics[width=0.32\textwidth]{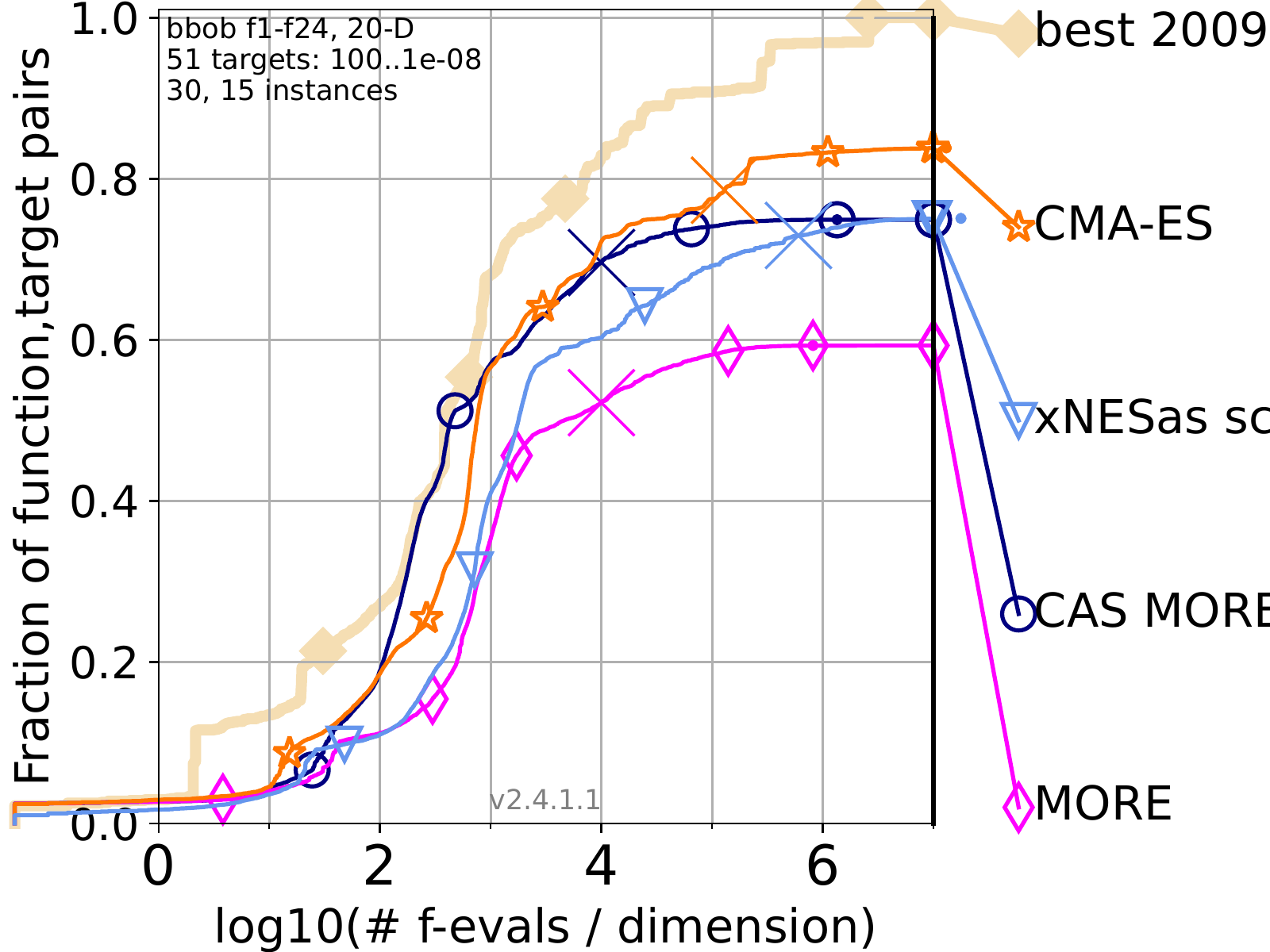}
		\vspace*{-1ex}
	\end{tabular}
	\caption{
		\label{fig:ECDFs20D}
		\bbobECDFslegend{20}
	}
\end{figure*}

Results from experiments according to \cite{hansen2016exp} and \cite{hansen2016perfass} on the benchmark functions given in \cite{wp200901_2010,hansen2010fun} are presented in Figures~\ref{fig:scaling}, \ref{fig:ECDFs05D} and \ref{fig:ECDFs20D}.
The experiments were performed with COCO \cite{hansen2020cocoplat}, version 2.4.1.1, the plots were produced with version 2.4.1.1.
The \textbf{expected runtime (ERT)}, used in the figures and tables,
depends on a given target function value, $\ftarget=\fopt+\Df$, and is
computed over all relevant trials as the number of function evaluations executed during each trial while the best function value did not reach \ftarget, summed over all trials and divided by the number of trials that actually reached \ftarget\ \cite{hansen2012exp,price1997dev}.  \textbf{Statistical significance} is tested with the rank-sum test for a given target $\Delta\ftarget$
using, for each trial, either the number of needed function evaluations to reach $\Delta\ftarget$ (inverted and multiplied by $-1$), or, if the target was not reached, the best $\Df$-value achieved, measured only up to the smallest number of overall function evaluations for any unsuccessful trial under consideration.

\begin{figure*}
	\centering
	\begin{tabular}{@{}l@{}l@{}l@{}l@{}l@{}}
		\includegraphics[width=0.24\textwidth]{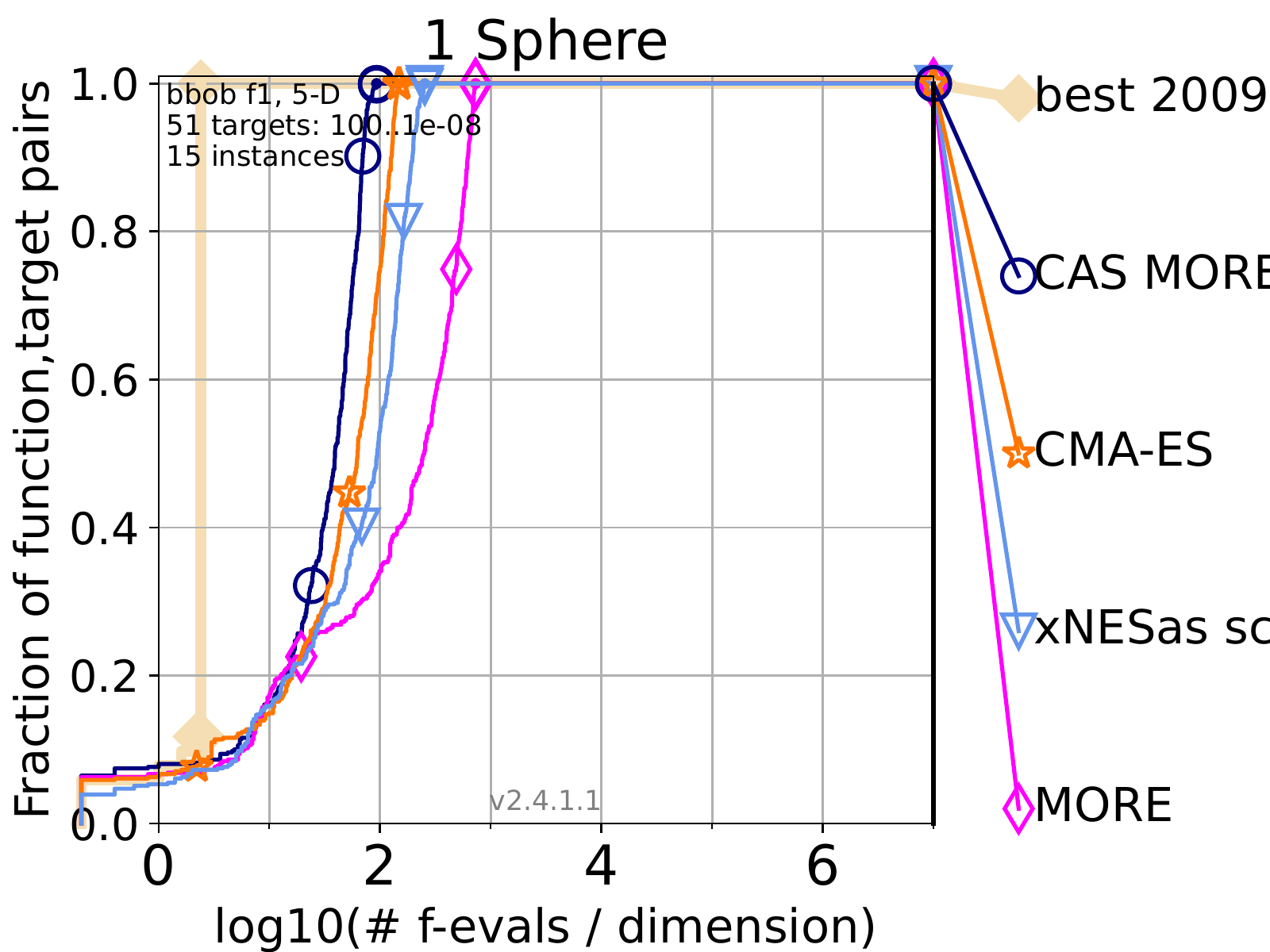}&
		\includegraphics[width=0.24\textwidth]{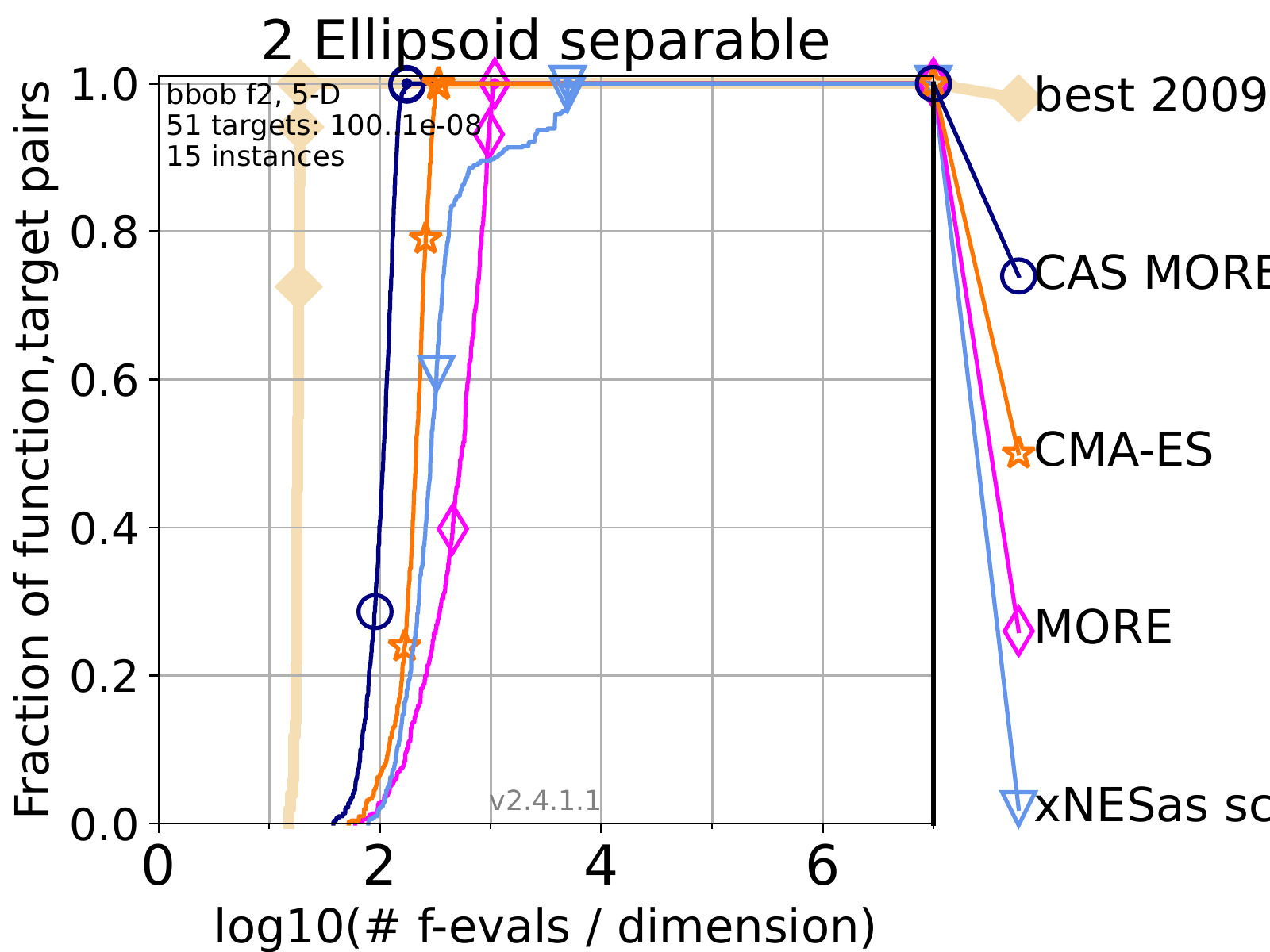}&
		\includegraphics[width=0.24\textwidth]{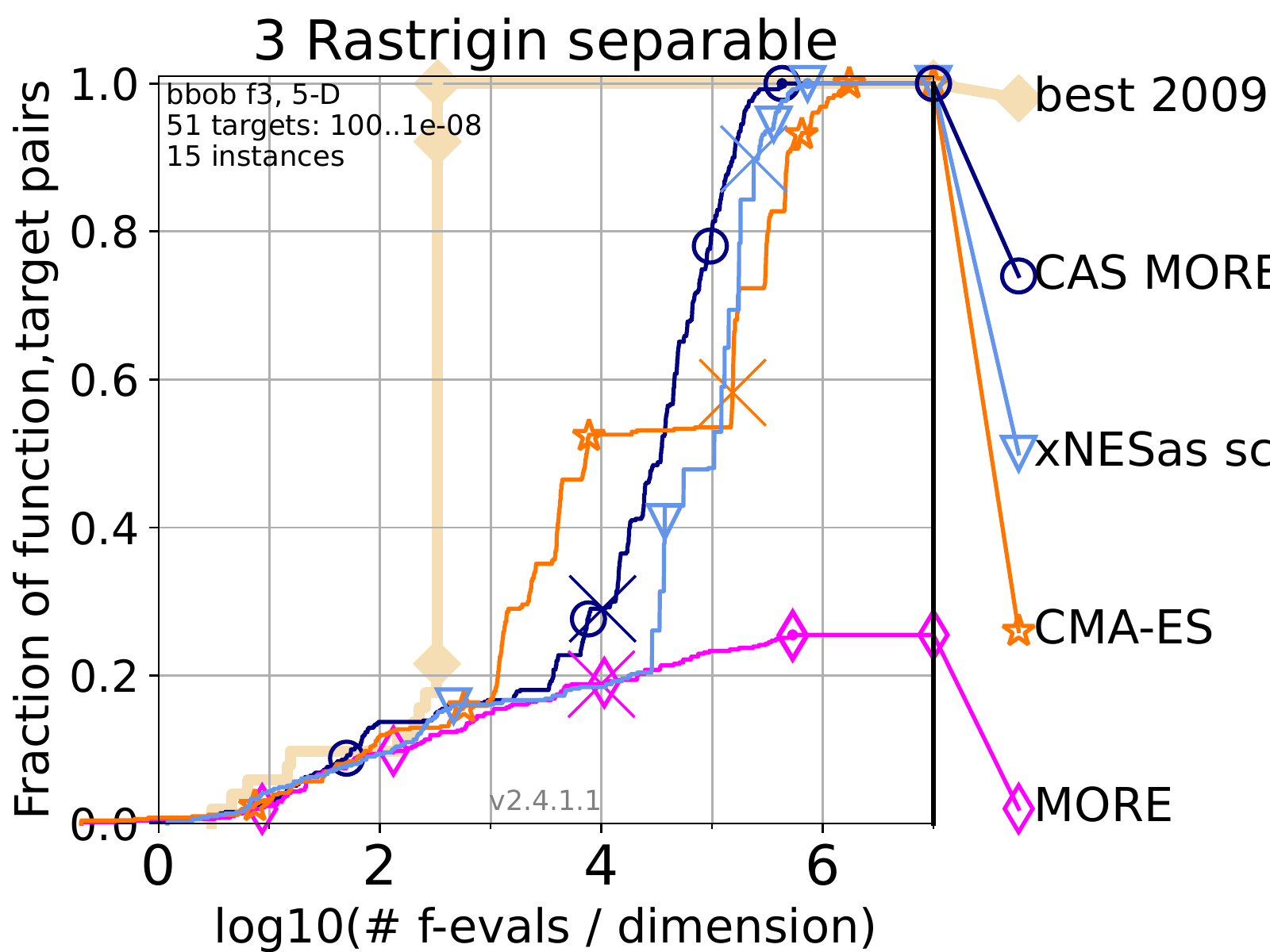}&
		\includegraphics[width=0.24\textwidth]{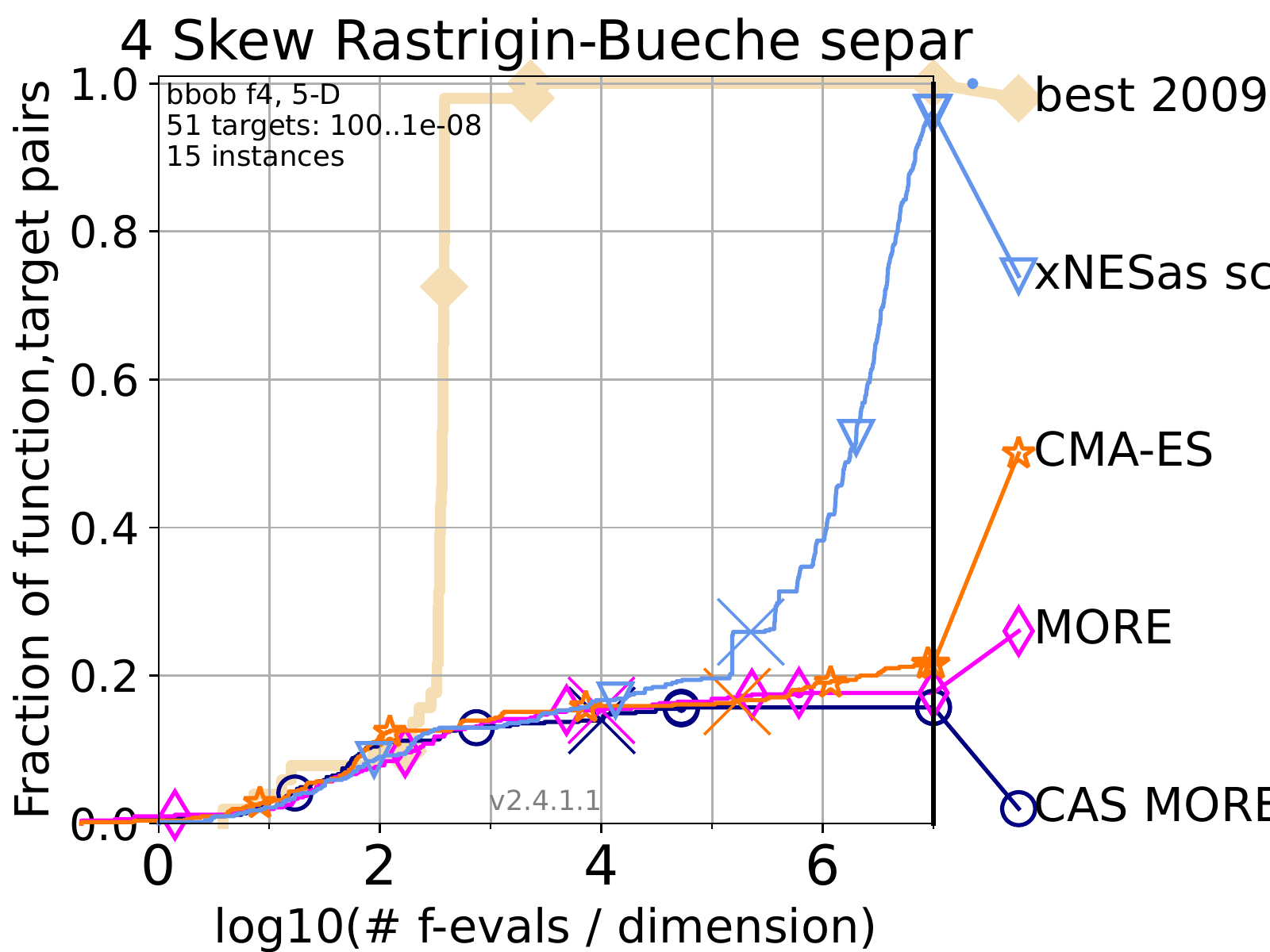}\\
		\includegraphics[width=0.24\textwidth]{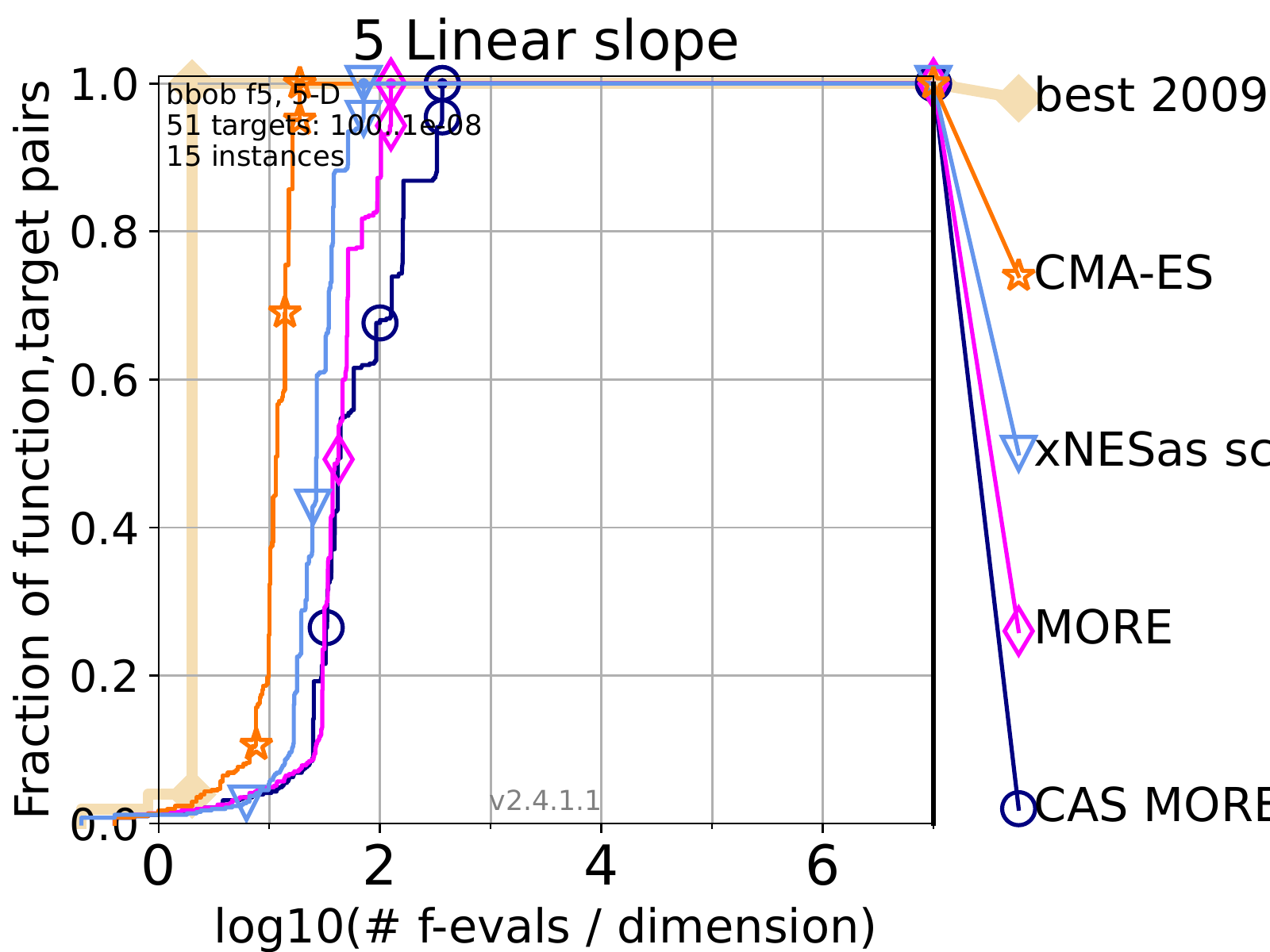}&
		\includegraphics[width=0.24\textwidth]{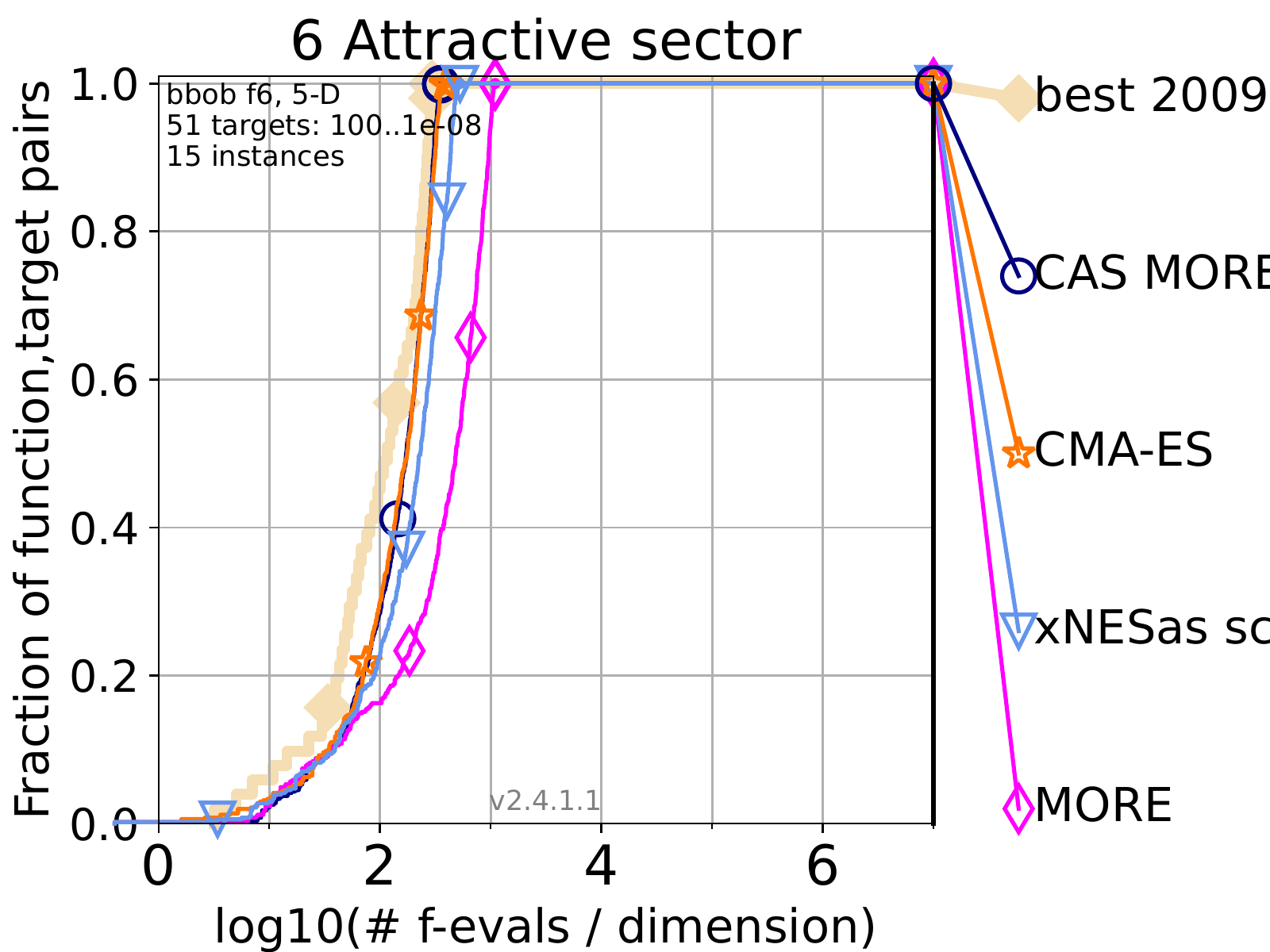}&
		\includegraphics[width=0.24\textwidth]{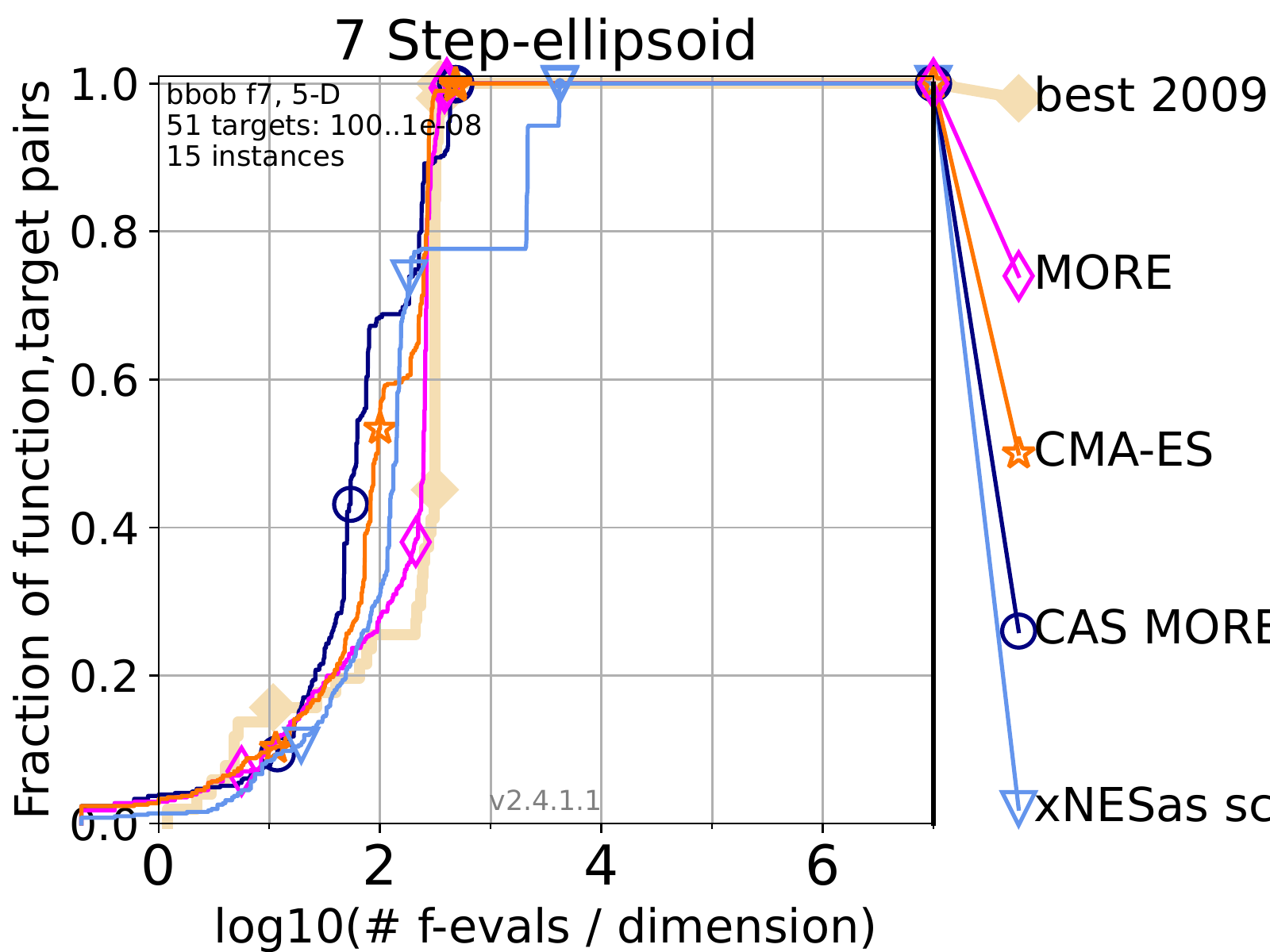}&
		\includegraphics[width=0.24\textwidth]{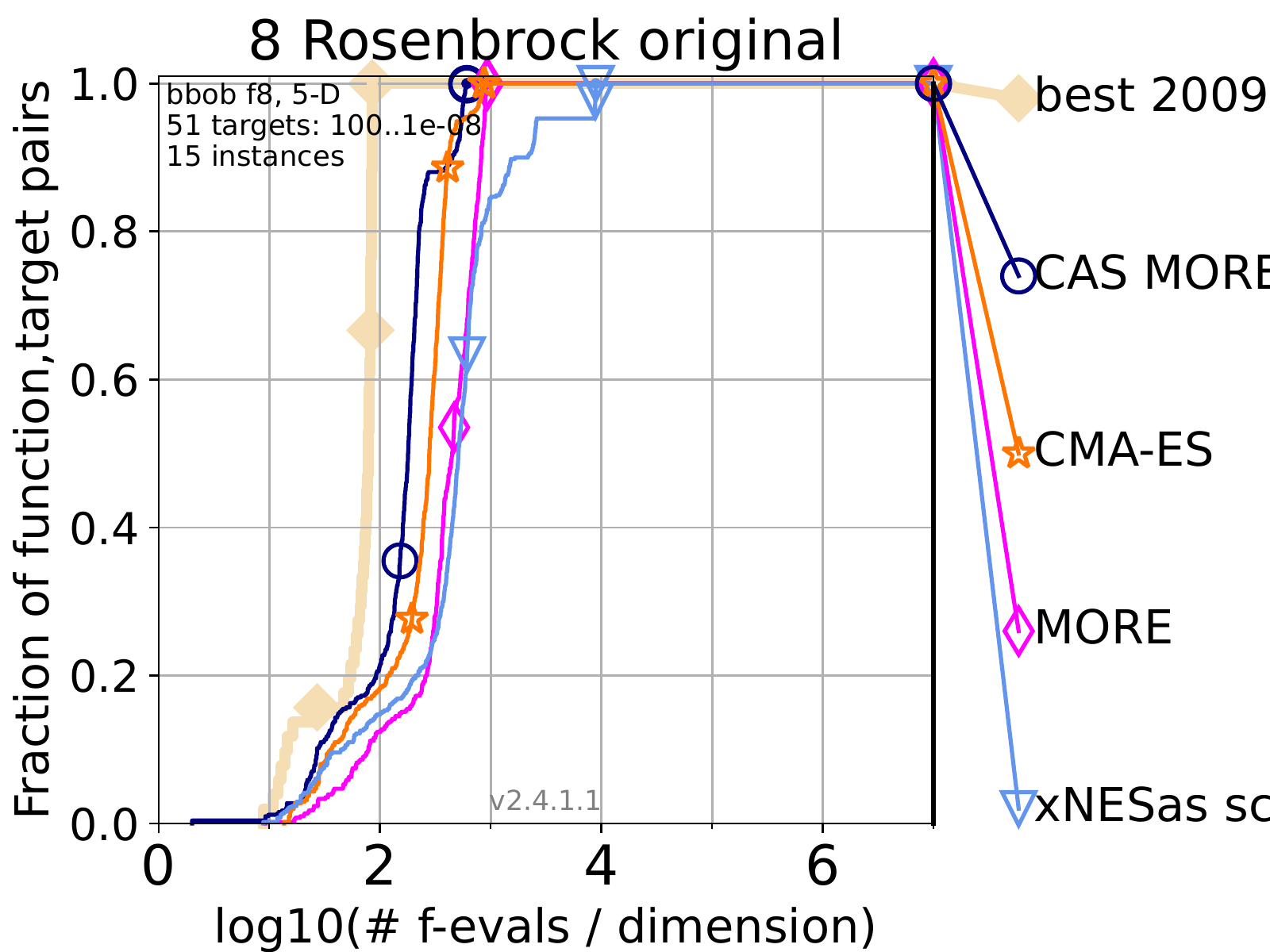}\\
		\includegraphics[width=0.24\textwidth]{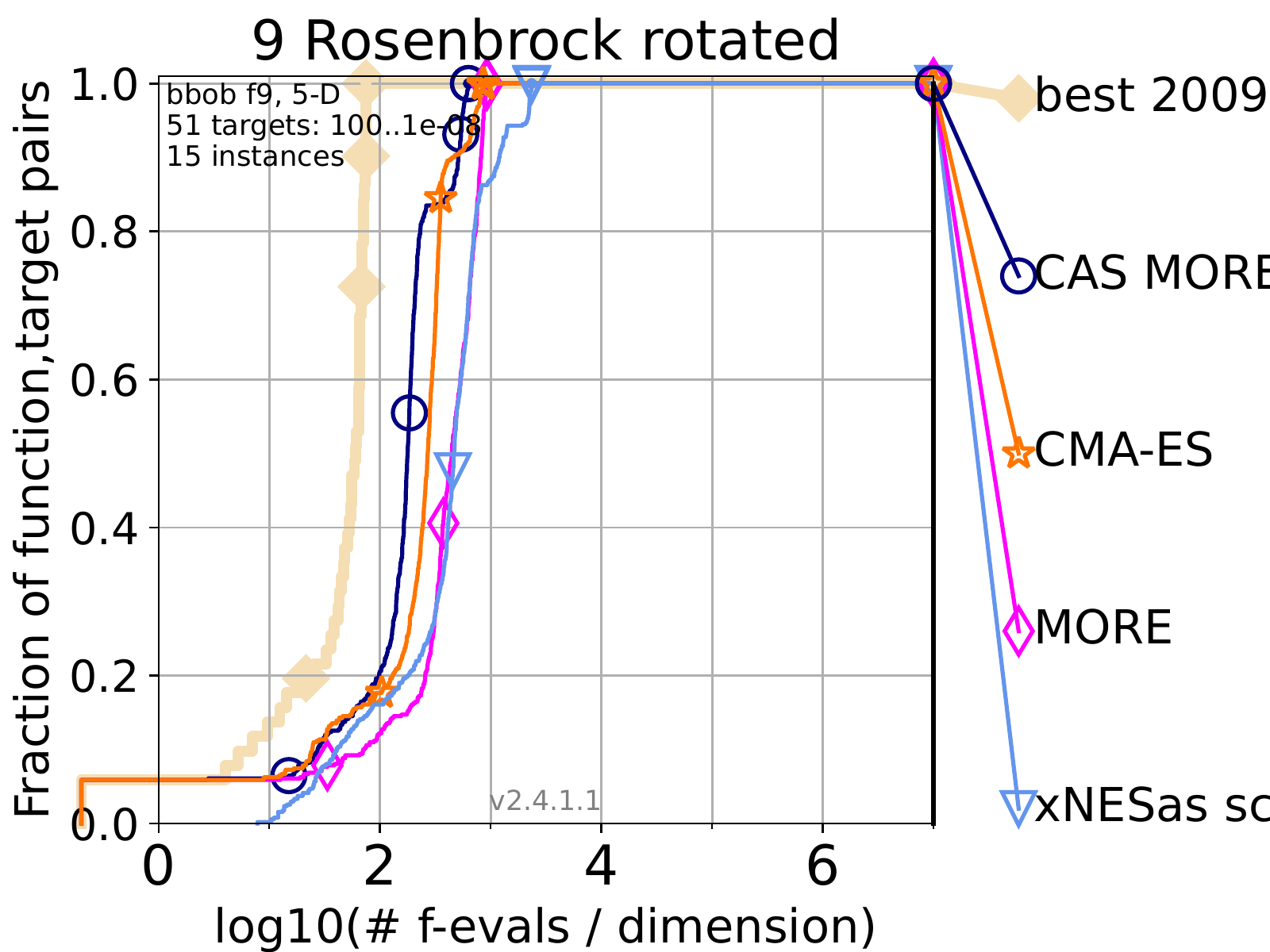}&
		\includegraphics[width=0.24\textwidth]{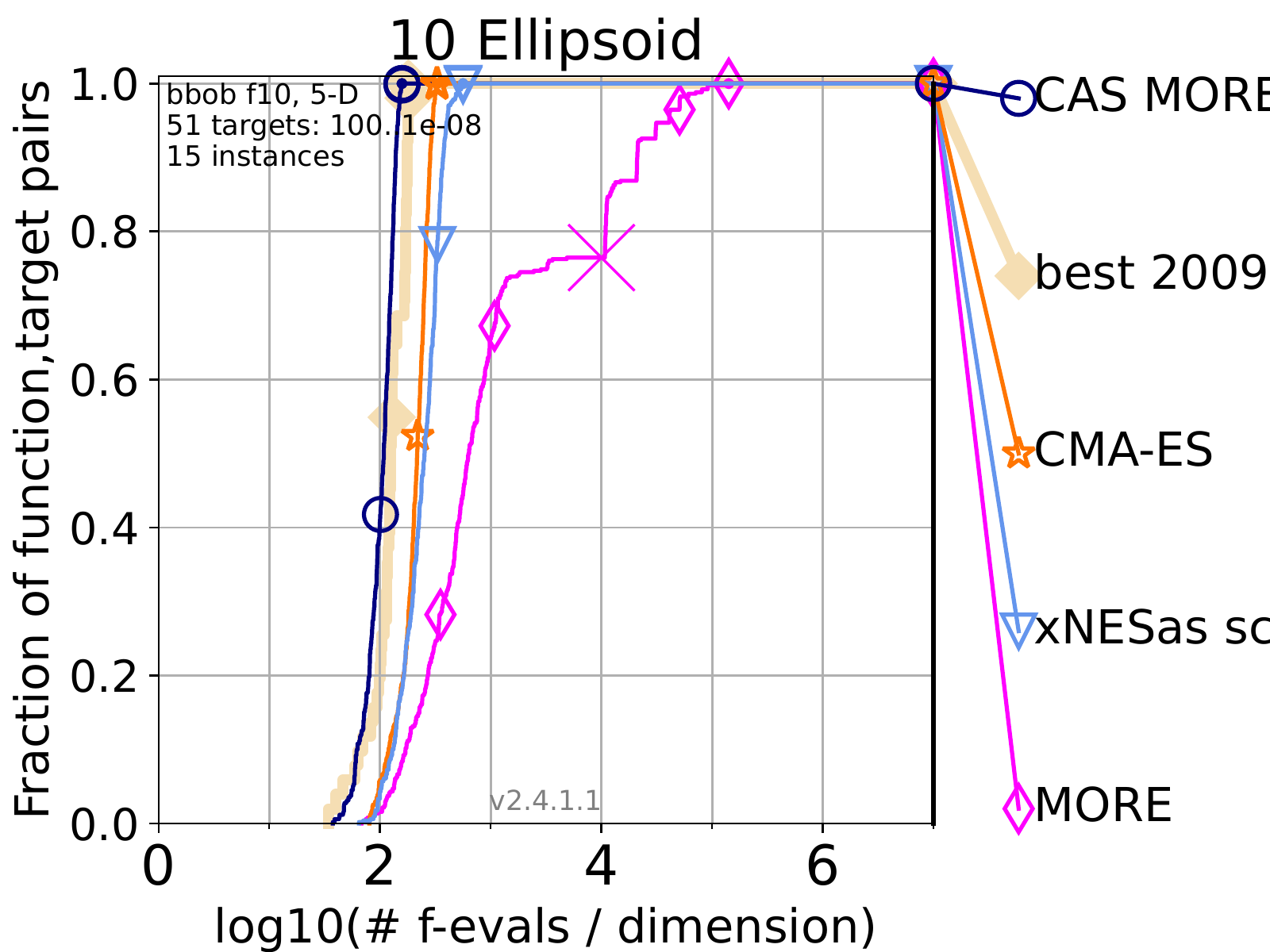}&
		\includegraphics[width=0.24\textwidth]{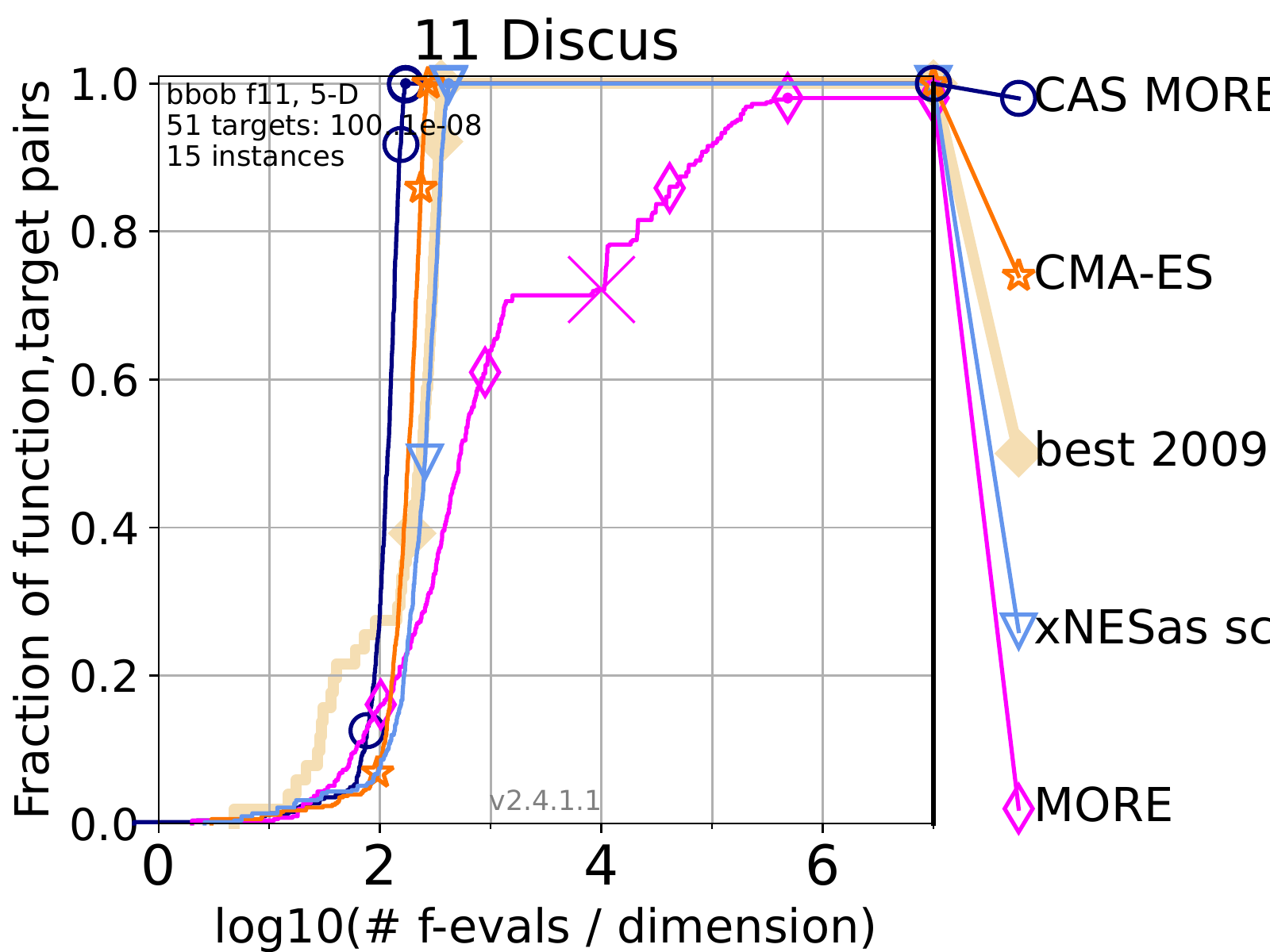}&
		\includegraphics[width=0.24\textwidth]{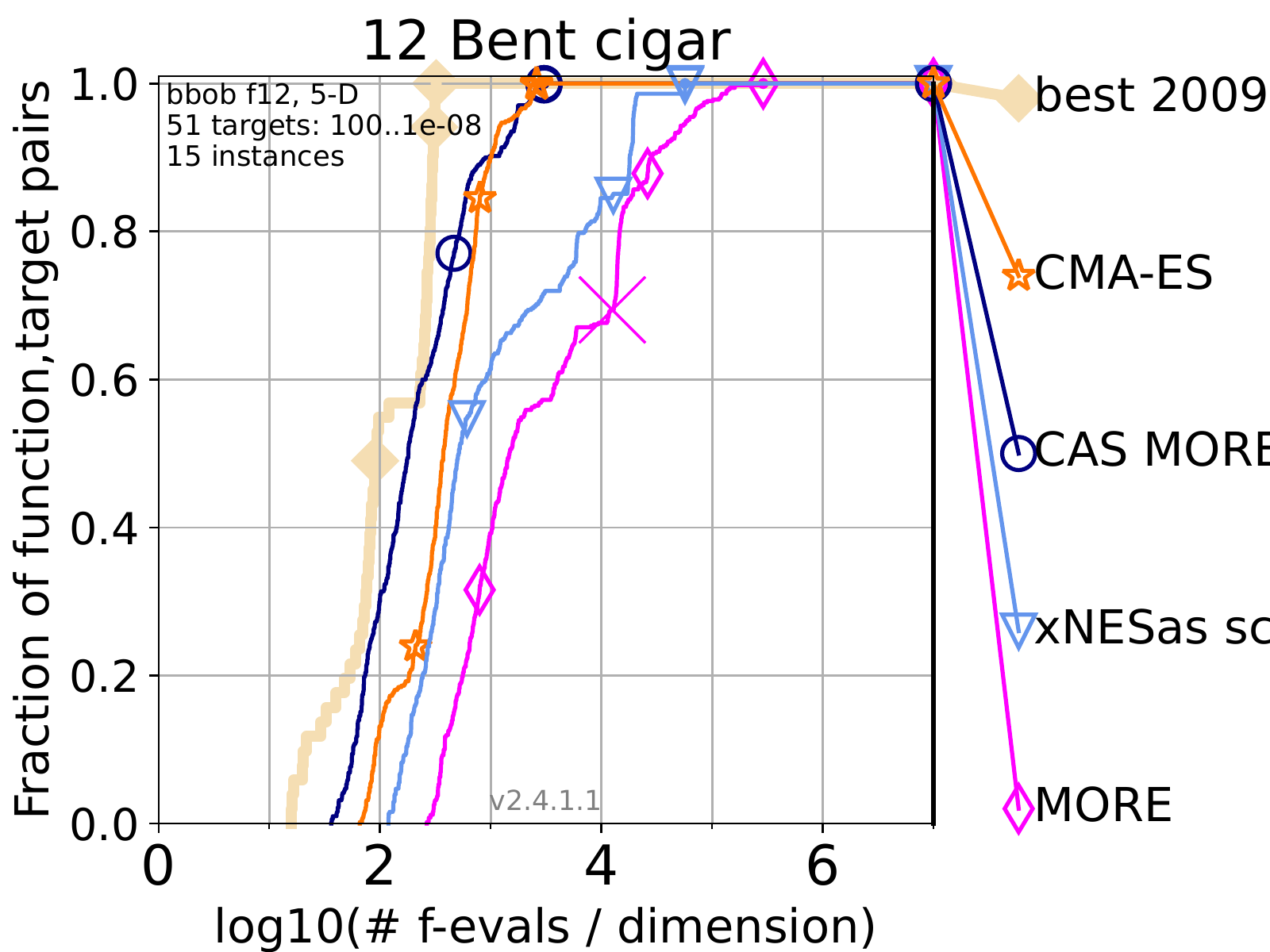}\\
		\includegraphics[width=0.24\textwidth]{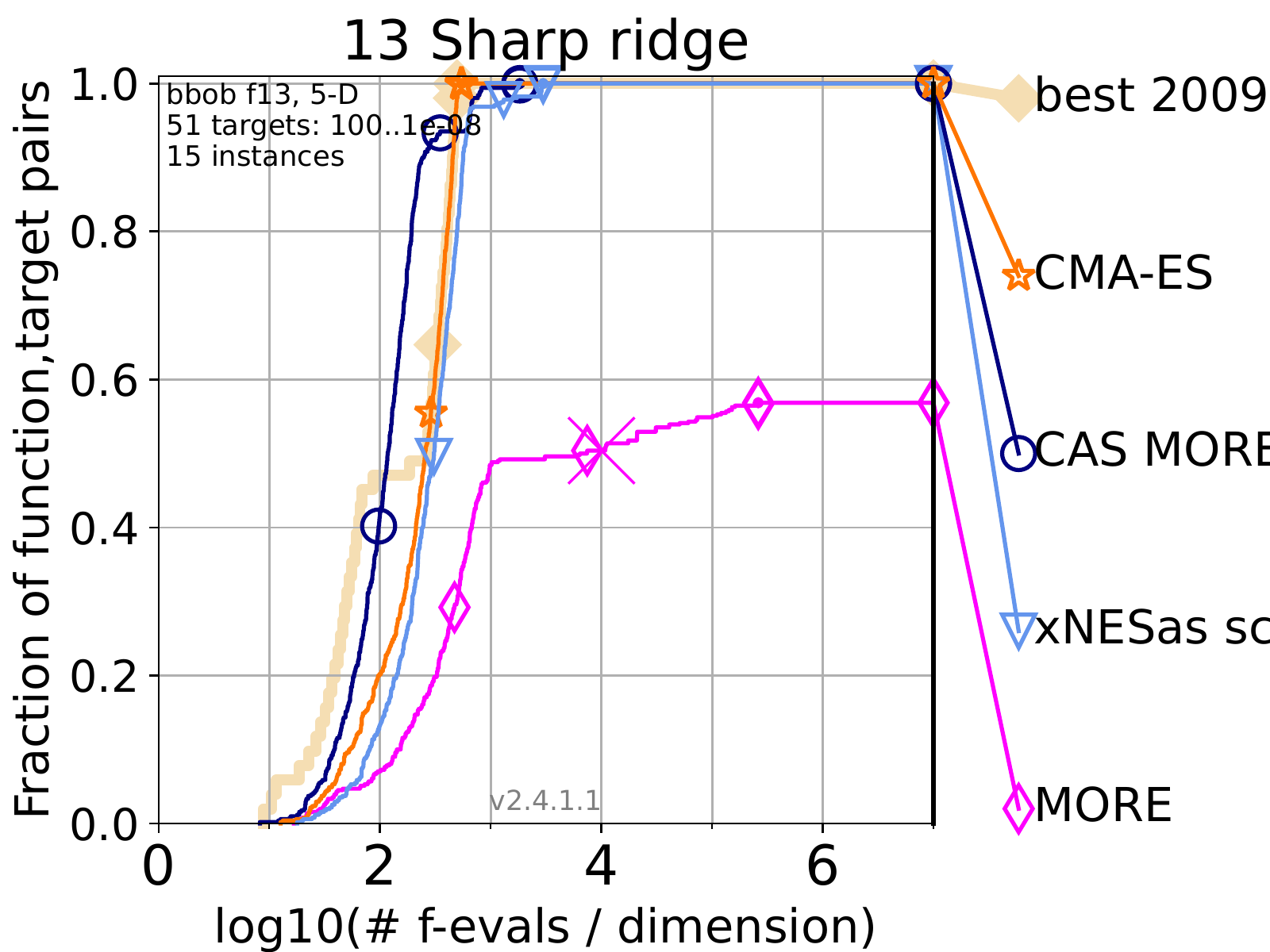}&
		\includegraphics[width=0.24\textwidth]{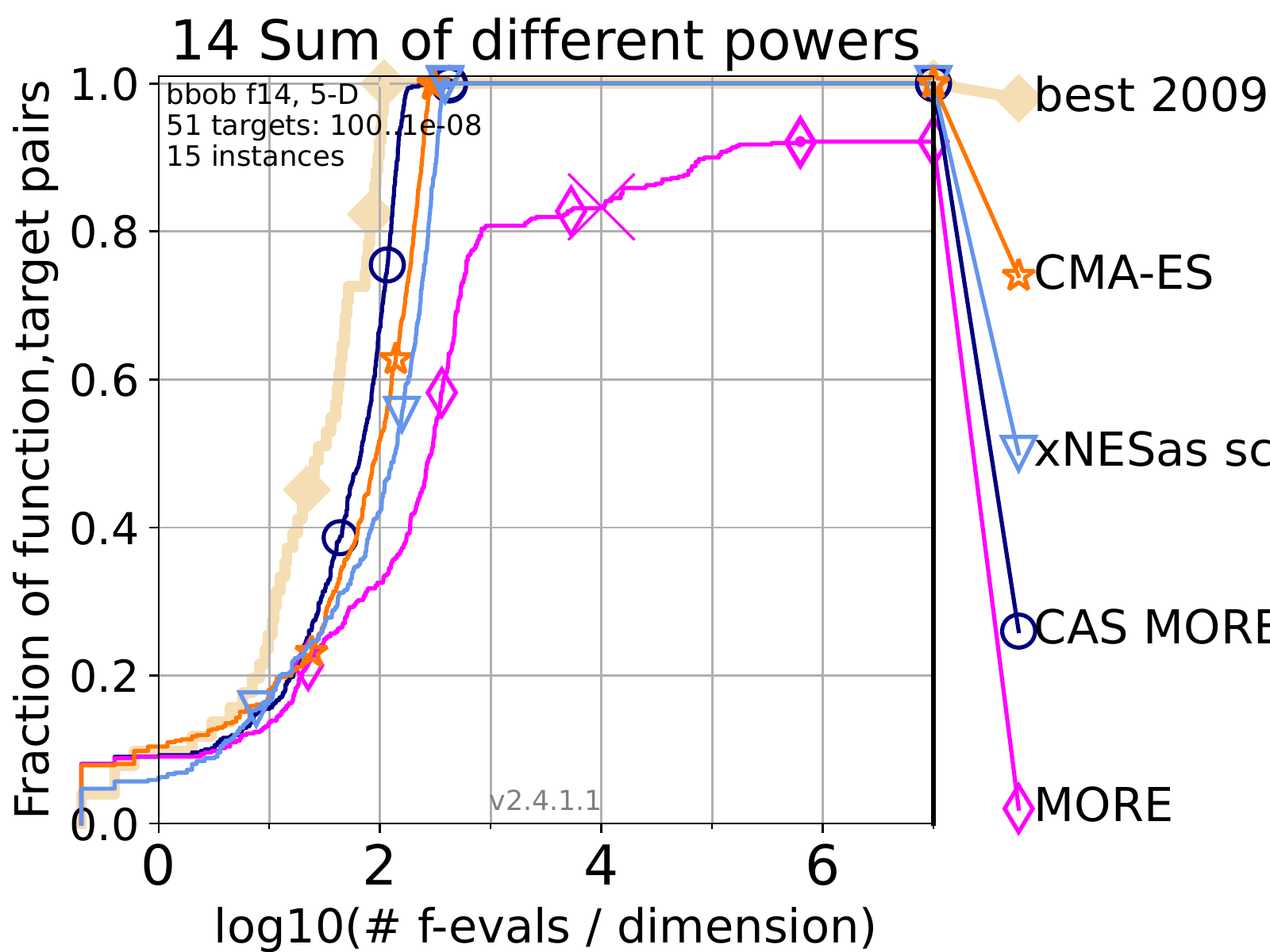}&
		\includegraphics[width=0.24\textwidth]{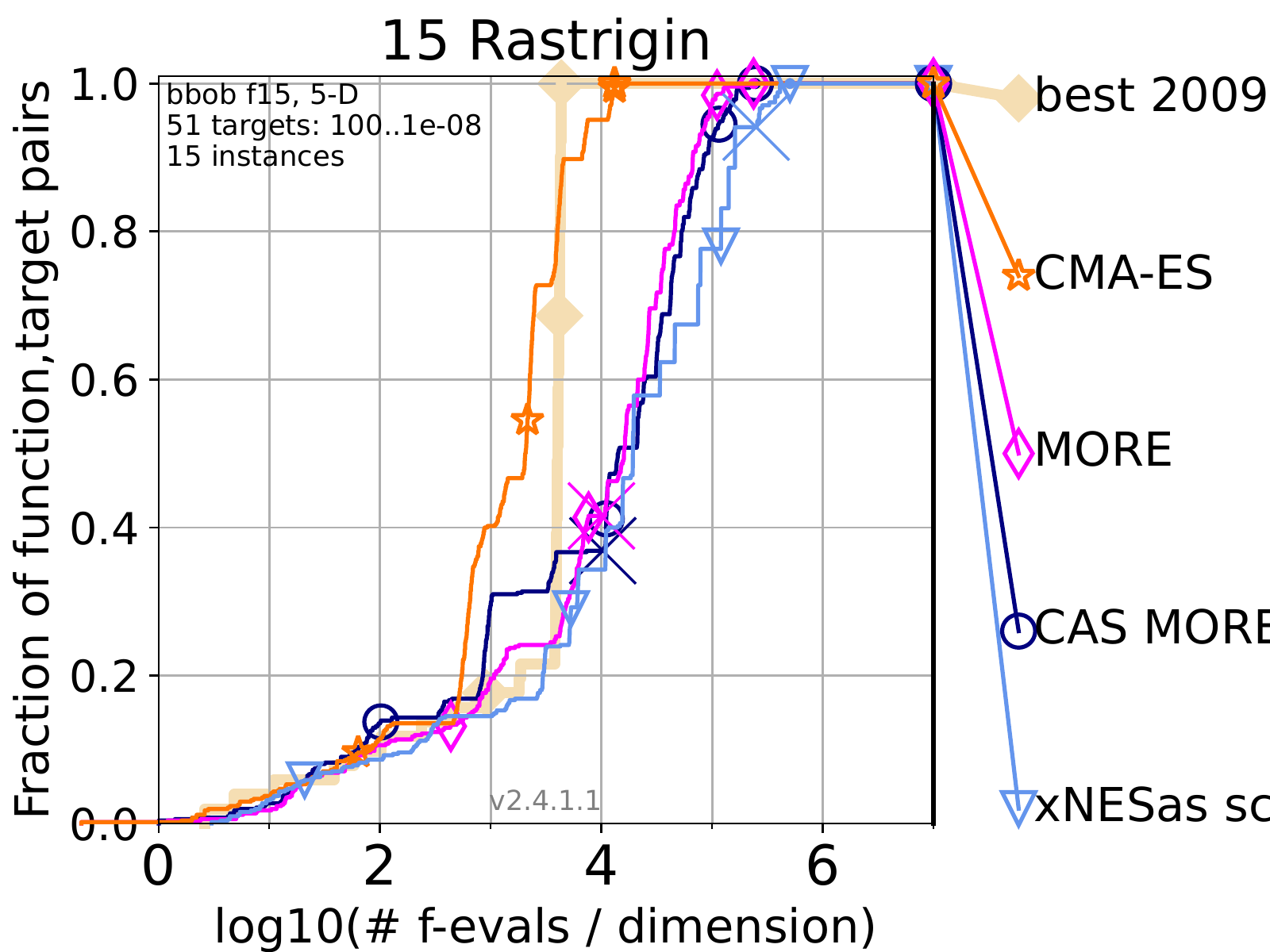}&
		\includegraphics[width=0.24\textwidth]{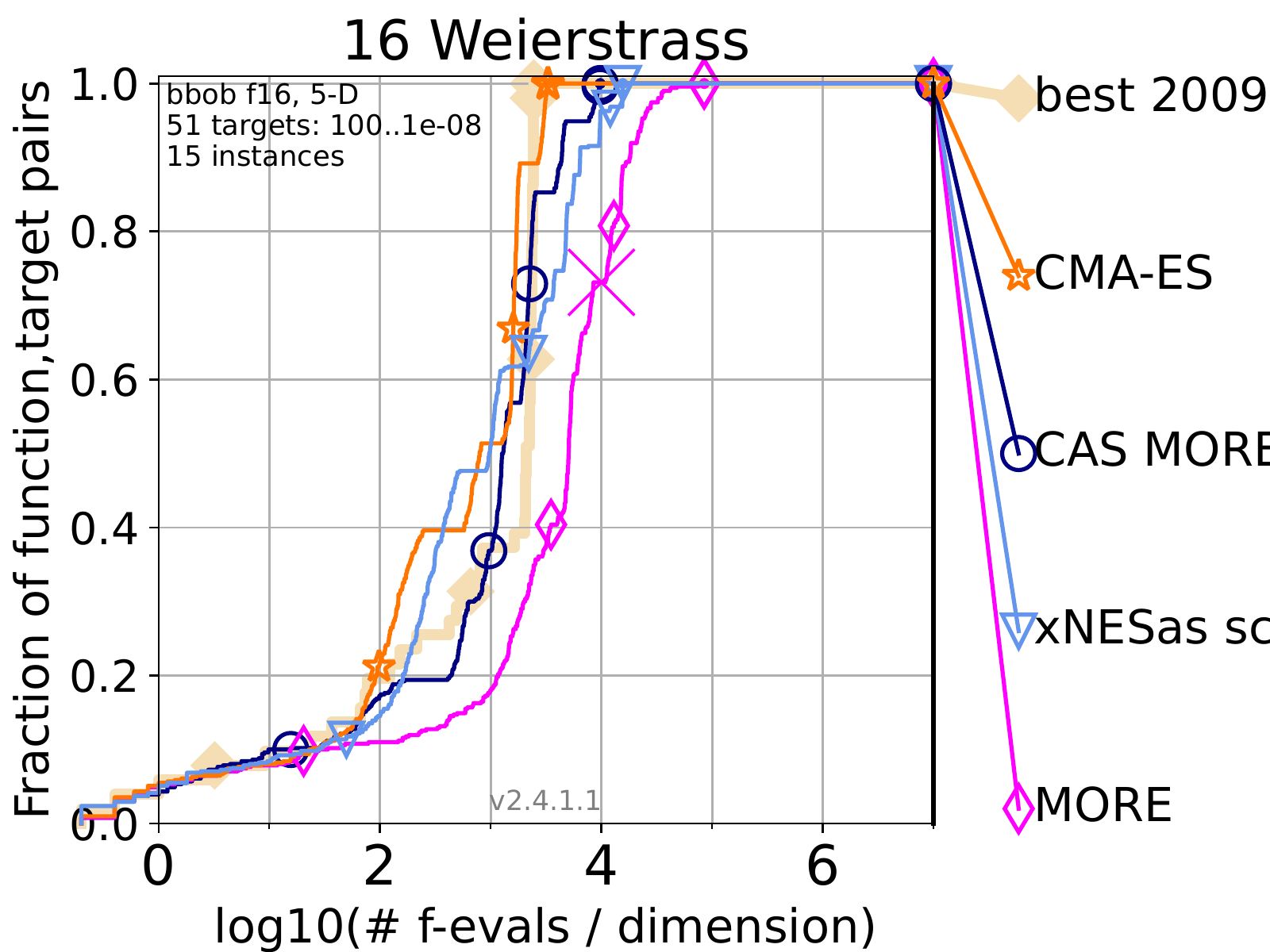}\\
		\includegraphics[width=0.24\textwidth]{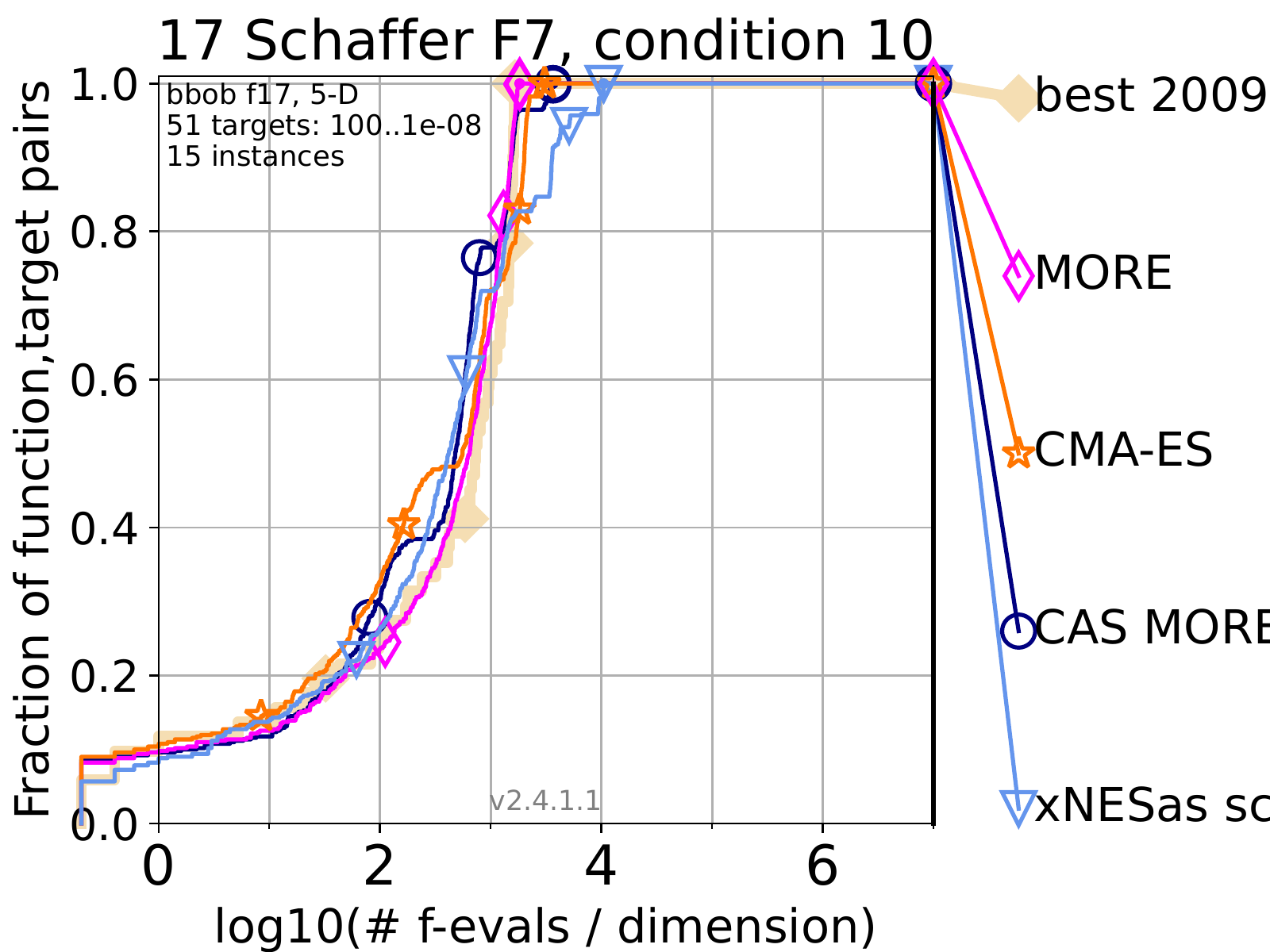}&
		\includegraphics[width=0.24\textwidth]{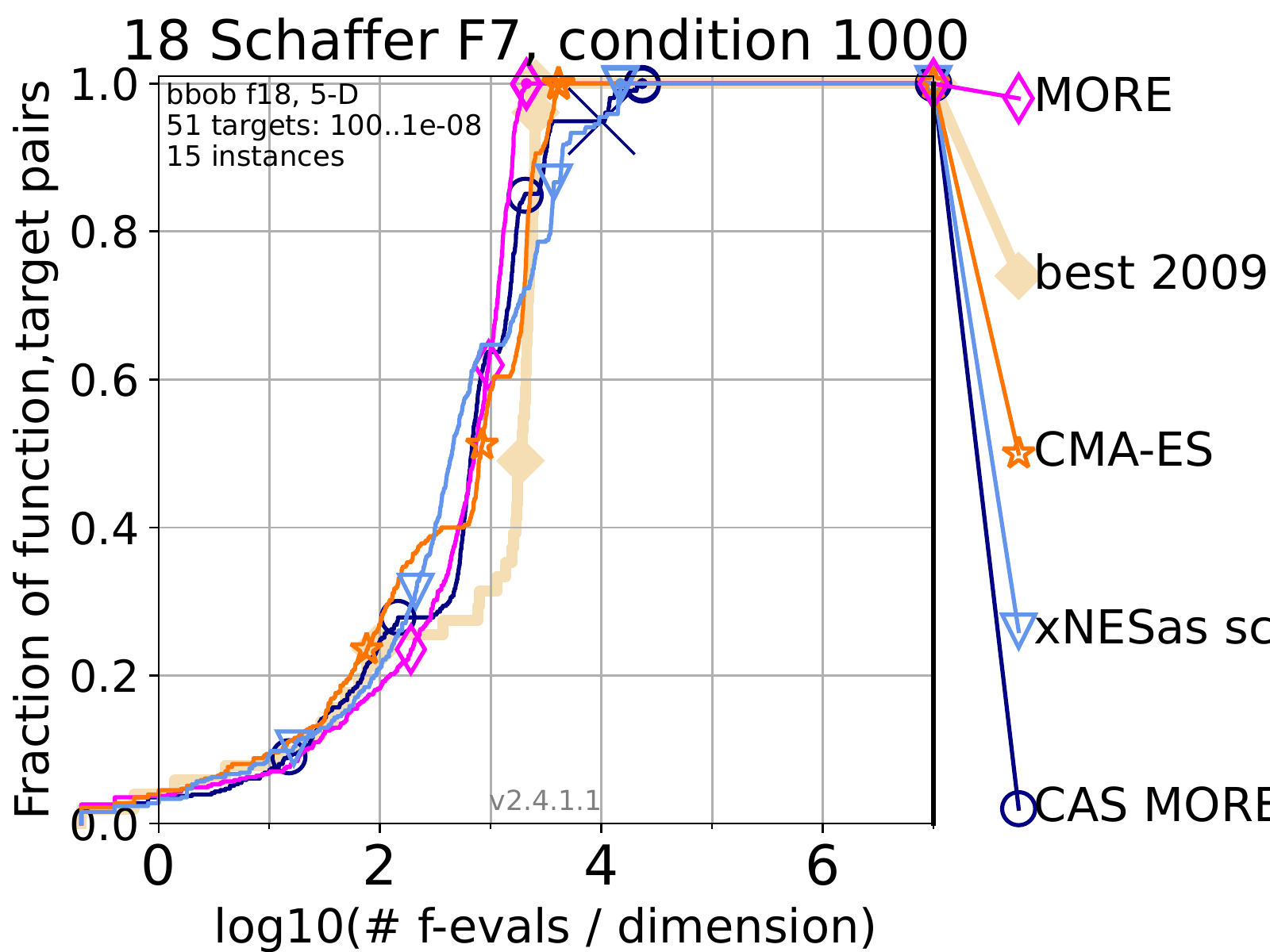}&
		\includegraphics[width=0.24\textwidth]{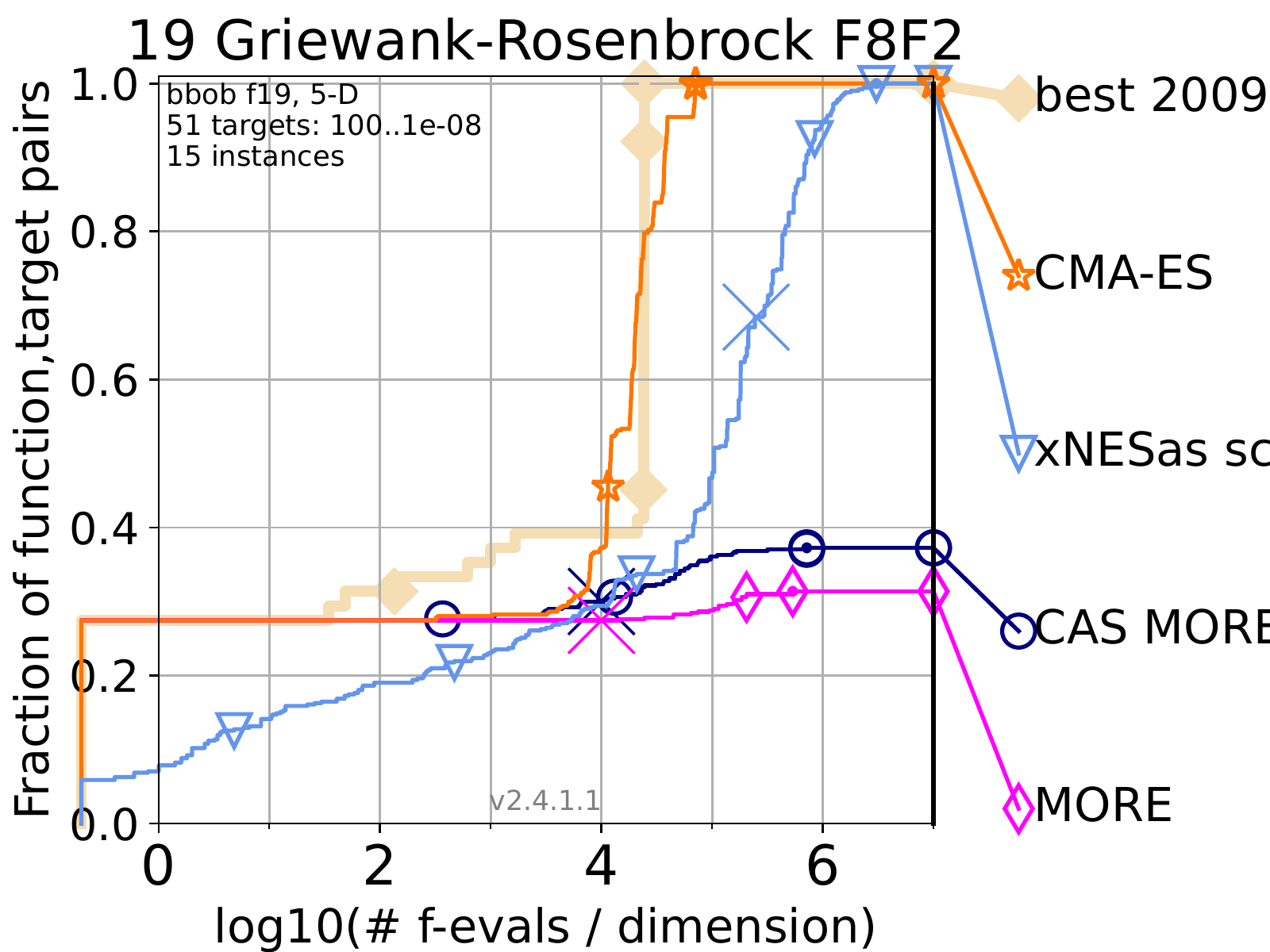}&
		\includegraphics[width=0.24\textwidth]{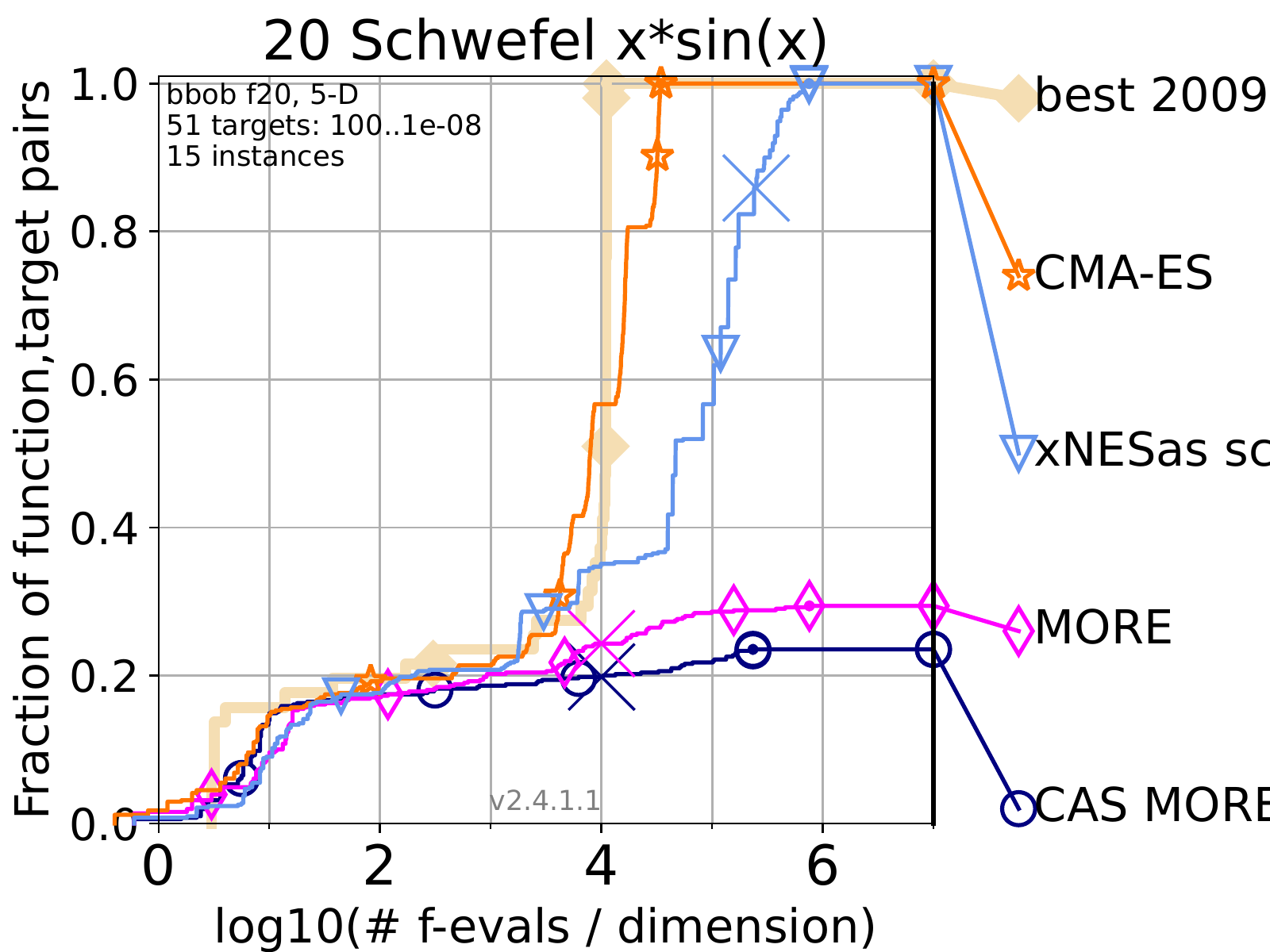}\\
		\includegraphics[width=0.24\textwidth]{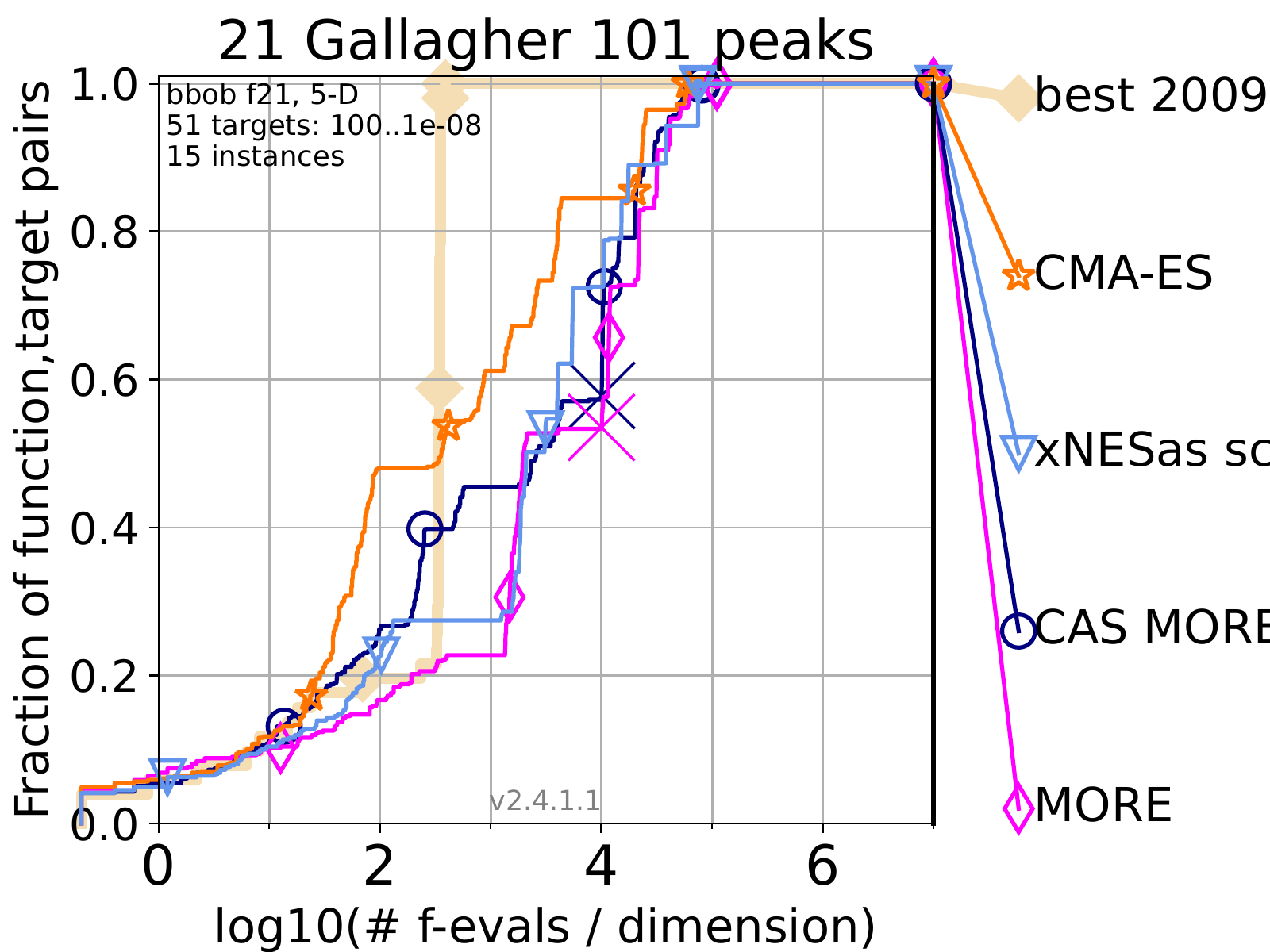}&
		\includegraphics[width=0.24\textwidth]{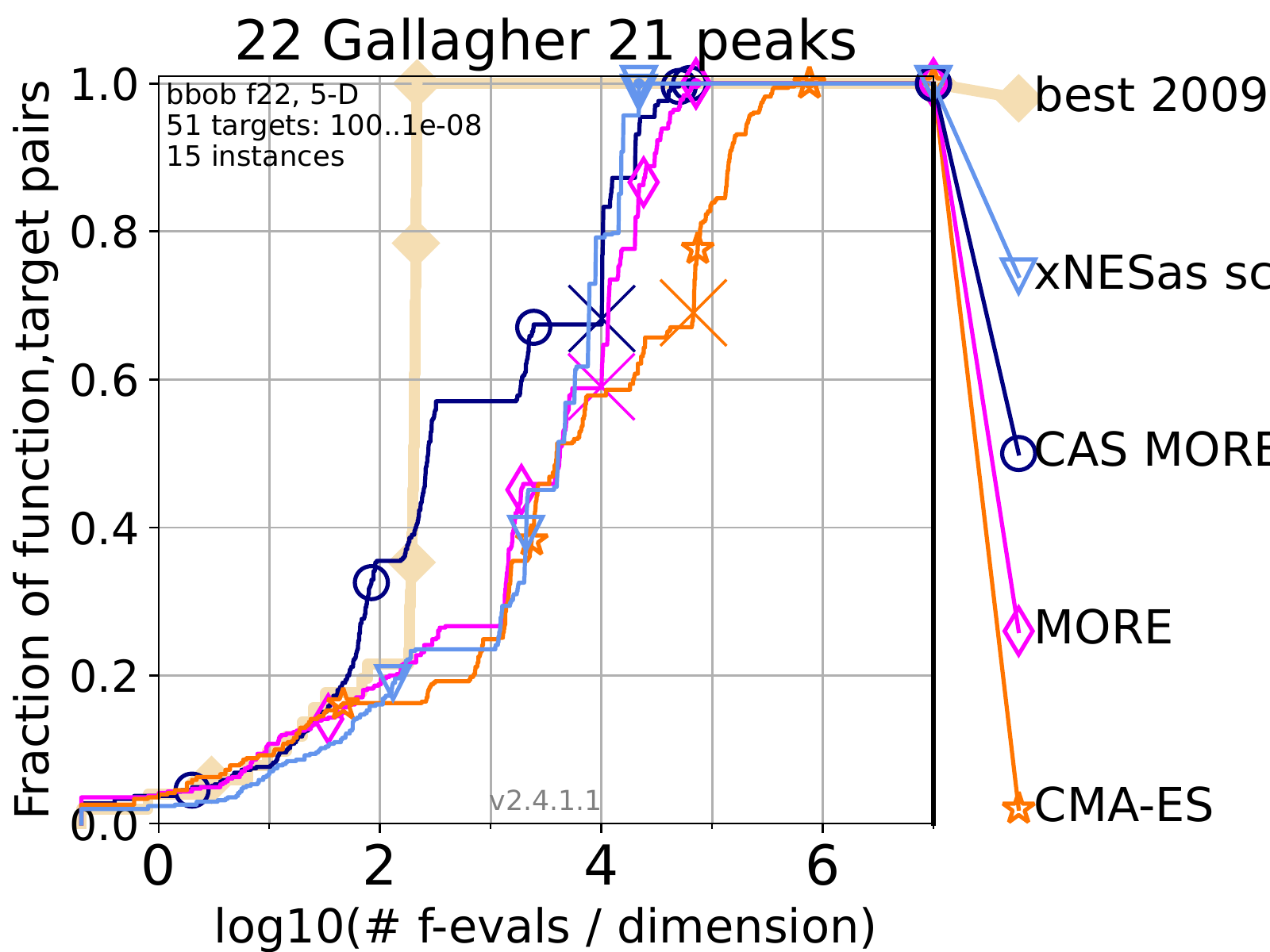}&
		\includegraphics[width=0.24\textwidth]{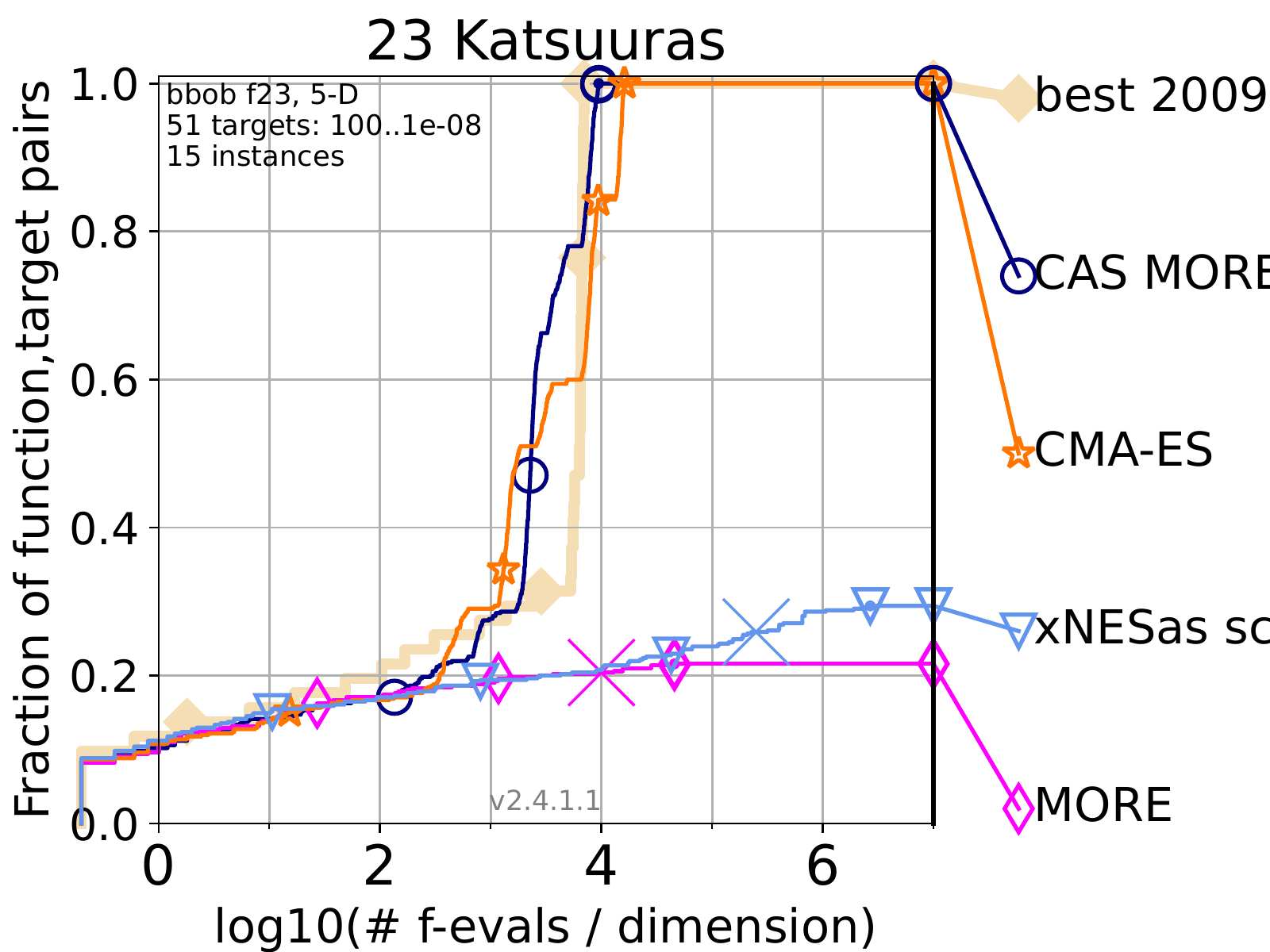}&
		\includegraphics[width=0.24\textwidth]{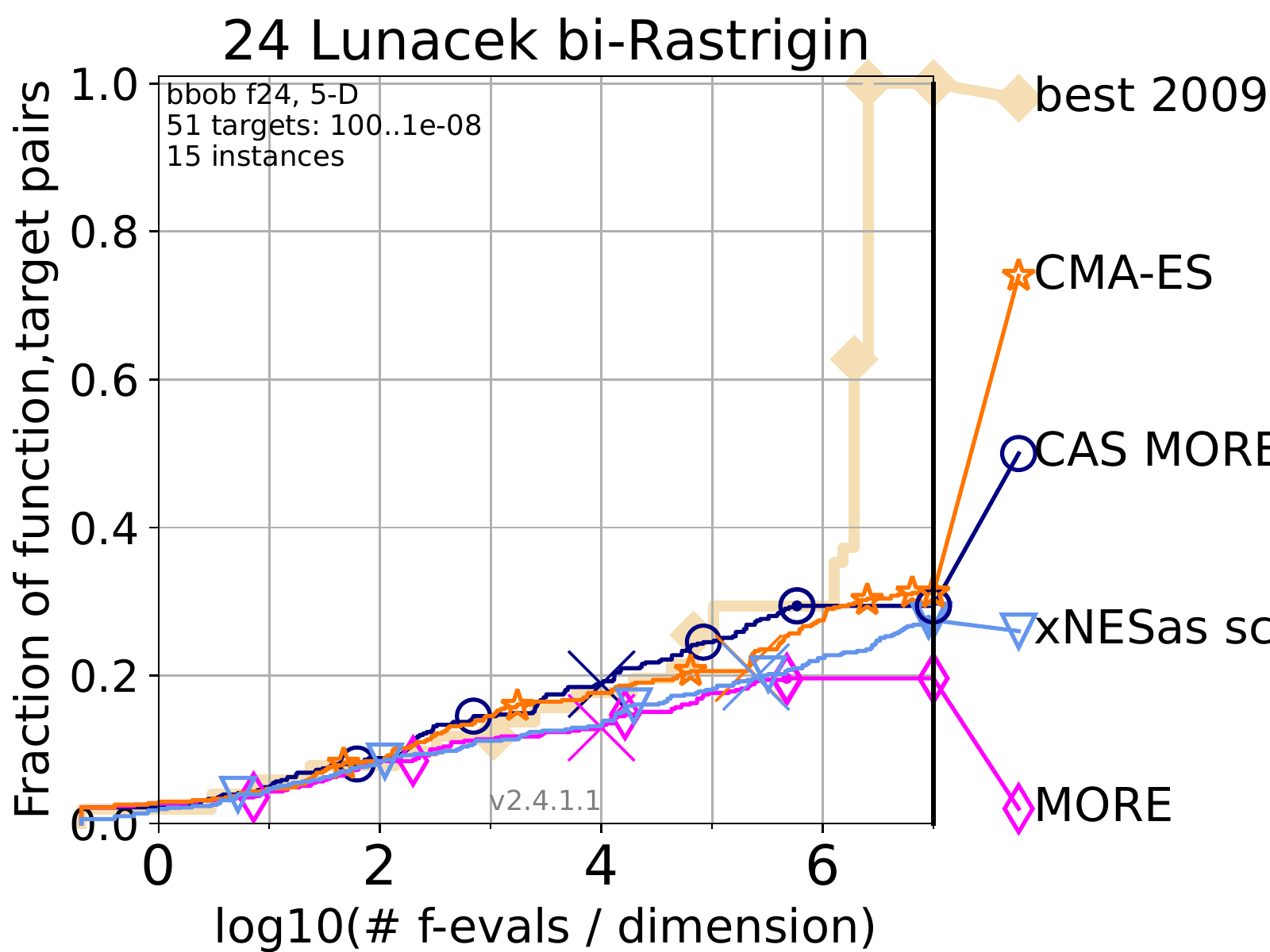}
	\end{tabular}
	\caption{\label{fig:ECDFsingleOne}
		\bbobecdfcaptionsinglefunctionssingledim{5}
	}
\end{figure*}

\FloatBarrier

\bibliographystyle{plainnat}
\bibliography{more}

\begin{thebibliography}{57}
\providecommand{\natexlab}[1]{#1}
\providecommand{\url}[1]{\texttt{#1}}
\expandafter\ifx\csname urlstyle\endcsname\relax
  \providecommand{\doi}[1]{doi: #1}\else
  \providecommand{\doi}{doi: \begingroup \urlstyle{rm}\Url}\fi

\bibitem[Abdolmaleki et~al.(2015)Abdolmaleki, Lioutikov, Peters, Lau,
  Pualo~Reis, and Neumann]{NIPS2015_36ac8e55}
Abbas Abdolmaleki, Rudolf Lioutikov, Jan~R Peters, Nuno Lau, Luis Pualo~Reis,
  and Gerhard Neumann.
\newblock Model-based relative entropy stochastic search.
\newblock In C.~Cortes, N.~Lawrence, D.~Lee, M.~Sugiyama, and R.~Garnett,
  editors, \emph{Advances in Neural Information Processing Systems}, volume~28.
  Curran Associates, Inc., 2015.

\bibitem[Abdolmaleki et~al.(2018)Abdolmaleki, Springenberg, Tassa, Munos,
  Heess, and Riedmiller]{abdolmaleki2018maximum}
Abbas Abdolmaleki, Jost~Tobias Springenberg, Yuval Tassa, Remi Munos, Nicolas
  Heess, and Martin Riedmiller.
\newblock Maximum a posteriori policy optimisation.
\newblock In \emph{International Conference on Learning Representations}, 2018.

\bibitem[Akimoto et~al.(2010)Akimoto, Nagata, Ono, and
  Kobayashi]{akimoto2010bidirectional}
Youhei Akimoto, Yuichi Nagata, Isao Ono, and Shigenobu Kobayashi.
\newblock Bidirectional relation between cma evolution strategies and natural
  evolution strategies.
\newblock In \emph{International Conference on Parallel Problem Solving from
  Nature}, pages 154--163. Springer, 2010.

\bibitem[Akrour et~al.(2018)Akrour, Abdolmaleki, Abdulsamad, Peters, and
  Neumann]{akrour2018model}
Riad Akrour, Abbas Abdolmaleki, Hany Abdulsamad, Jan Peters, and Gerhard
  Neumann.
\newblock Model-free trajectory-based policy optimization with monotonic
  improvement.
\newblock \emph{The Journal of Machine Learning Research}, 19\penalty0
  (1):\penalty0 565--589, 2018.

\bibitem[Amari(1998)]{amari1998natural}
Shun-Ichi Amari.
\newblock Natural gradient works efficiently in learning.
\newblock \emph{Neural computation}, 10\penalty0 (2):\penalty0 251--276, 1998.

\bibitem[Amos and Yarats(2020)]{amos2020differentiable}
Brandon Amos and Denis Yarats.
\newblock The differentiable cross-entropy method.
\newblock In \emph{International Conference on Machine Learning}, pages
  291--302. PMLR, 2020.

\bibitem[Arenz et~al.(2018)Arenz, Neumann, and Zhong]{arenz2018efficient}
Oleg Arenz, Gerhard Neumann, and Mingjun Zhong.
\newblock Efficient gradient-free variational inference using policy search.
\newblock In \emph{International conference on machine learning}, pages
  234--243. PMLR, 2018.

\bibitem[Auger and Hansen(2005)]{auger2005restart}
Anne Auger and Nikolaus Hansen.
\newblock A restart cma evolution strategy with increasing population size.
\newblock In \emph{2005 IEEE congress on evolutionary computation}, volume~2,
  pages 1769--1776. IEEE, 2005.

\bibitem[Auger et~al.(2004)Auger, Schoenauer, and Vanhaecke]{auger2004ls}
Anne Auger, Marc Schoenauer, and Nicolas Vanhaecke.
\newblock Ls-cma-es: A second-order algorithm for covariance matrix adaptation.
\newblock In \emph{International Conference on Parallel Problem Solving from
  Nature}, pages 182--191. Springer, 2004.

\bibitem[Becker et~al.(2019)Becker, Arenz, and Neumann]{becker2019expected}
Philipp Becker, Oleg Arenz, and Gerhard Neumann.
\newblock Expected information maximization: Using the i-projection for mixture
  density estimation.
\newblock In \emph{International Conference on Learning Representations}, 2019.

\bibitem[Beyer and Schwefel(2002)]{beyer2002evolution}
Hans-Georg Beyer and Hans-Paul Schwefel.
\newblock Evolution strategies--a comprehensive introduction.
\newblock \emph{Natural computing}, 1\penalty0 (1):\penalty0 3--52, 2002.

\bibitem[Botev et~al.(2013)Botev, Kroese, Rubinstein, and
  L’Ecuyer]{botev2013cross}
Zdravko~I Botev, Dirk~P Kroese, Reuven~Y Rubinstein, and Pierre L’Ecuyer.
\newblock The cross-entropy method for optimization.
\newblock In \emph{Handbook of statistics}, volume~31, pages 35--59. Elsevier,
  2013.

\bibitem[Chatzilygeroudis et~al.(2017)Chatzilygeroudis, Rama, Kaushik, Goepp,
  Vassiliades, and Mouret]{chatzilygeroudis2017black}
Konstantinos Chatzilygeroudis, Roberto Rama, Rituraj Kaushik, Dorian Goepp,
  Vassilis Vassiliades, and Jean-Baptiste Mouret.
\newblock Black-box data-efficient policy search for robotics.
\newblock In \emph{2017 IEEE/RSJ International Conference on Intelligent Robots
  and Systems (IROS)}, pages 51--58. IEEE, 2017.

\bibitem[Deisenroth et~al.(2013)Deisenroth, Neumann, Peters,
  et~al.]{deisenroth2013survey}
Marc~Peter Deisenroth, Gerhard Neumann, Jan Peters, et~al.
\newblock A survey on policy search for robotics.
\newblock \emph{Foundations and trends in Robotics}, 2\penalty0 (1-2):\penalty0
  388--403, 2013.

\bibitem[End et~al.(2017)End, Akrour, Peters, and Neumann]{end2017layered}
Felix End, Riad Akrour, Jan Peters, and Gerhard Neumann.
\newblock Layered direct policy search for learning hierarchical skills.
\newblock In \emph{2017 IEEE International Conference on Robotics and
  Automation (ICRA)}, pages 6442--6448. IEEE, 2017.

\bibitem[Finck et~al.(2009)Finck, Hansen, Ros, and Auger]{wp200901_2010}
S.~Finck, N.~Hansen, R.~Ros, and A.~Auger.
\newblock Real-parameter black-box optimization benchmarking 2009: Presentation
  of the noiseless functions.
\newblock Technical Report 2009/20, Research Center PPE, 2009.
\newblock Updated February 2010.

\bibitem[Glasmachers et~al.(2010)Glasmachers, Schaul, Yi, Wierstra, and
  Schmidhuber]{glasmachers2010exponential}
Tobias Glasmachers, Tom Schaul, Sun Yi, Daan Wierstra, and J{\"u}rgen
  Schmidhuber.
\newblock Exponential natural evolution strategies.
\newblock In \emph{Proceedings of the 12th annual conference on Genetic and
  evolutionary computation}, pages 393--400, 2010.

\bibitem[Hansen et~al.(2009{\natexlab{a}})Hansen, Finck, Ros, and
  Auger]{hansen2010fun}
N.~Hansen, S.~Finck, R.~Ros, and A.~Auger.
\newblock Real-parameter black-box optimization benchmarking 2009: Noiseless
  functions definitions.
\newblock Technical Report RR-6829, INRIA, 2009{\natexlab{a}}.
\newblock Updated February 2010.

\bibitem[Hansen et~al.(2012)Hansen, Auger, Finck, and Ros]{hansen2012exp}
N.~Hansen, A.~Auger, S.~Finck, and R.~Ros.
\newblock Real-parameter black-box optimization benchmarking 2012: Experimental
  setup.
\newblock Technical report, INRIA, 2012.

\bibitem[Hansen et~al.(2016{\natexlab{a}})Hansen, Auger, Brockhoff, Tu{\v s}ar,
  and Tu{\v s}ar]{hansen2016perfass}
N.~Hansen, A~Auger, D.~Brockhoff, D.~Tu{\v s}ar, and T.~Tu{\v s}ar.
\newblock {COCO}: Performance assessment.
\newblock \emph{ArXiv e-prints},
  \href{https://arxiv.org/abs/1605.03560}{arXiv:1605.03560},
  2016{\natexlab{a}}.

\bibitem[Hansen et~al.(2016{\natexlab{b}})Hansen, Tu{\v s}ar, Mersmann, Auger,
  and Brockhoff]{hansen2016exp}
N.~Hansen, T.~Tu{\v s}ar, O.~Mersmann, A.~Auger, and D.~Brockhoff.
\newblock {COCO}: The experimental procedure.
\newblock \emph{ArXiv e-prints},
  \href{https://arxiv.org/abs/1603.08776}{arXiv:1603.08776},
  2016{\natexlab{b}}.

\bibitem[Hansen et~al.(2020)Hansen, Auger, Ros, Mersmann, Tu{\v s}ar, and
  Brockhoff]{hansen2020cocoplat}
N.~Hansen, A.~Auger, R.~Ros, O.~Mersmann, T.~Tu{\v s}ar, and D.~Brockhoff.
\newblock {COCO}: A platform for comparing continuous optimizers in a black-box
  setting.
\newblock \emph{Optimization Methods and Software}, 2020.
\newblock \doi{https://doi.org/10.1080/10556788.2020.1808977}.

\bibitem[Hansen(2009)]{hansen2009benchmarking}
Nikolaus Hansen.
\newblock Benchmarking a bi-population cma-es on the bbob-2009 function
  testbed.
\newblock In \emph{Proceedings of the 11th annual conference companion on
  genetic and evolutionary computation conference: late breaking papers}, pages
  2389--2396, 2009.

\bibitem[Hansen(2016)]{hansen2016cma}
Nikolaus Hansen.
\newblock The cma evolution strategy: A tutorial.
\newblock \emph{arXiv preprint arXiv:1604.00772}, 2016.

\bibitem[Hansen(2019)]{hansen2019global}
Nikolaus Hansen.
\newblock A global surrogate assisted cma-es.
\newblock In \emph{Proceedings of the Genetic and Evolutionary Computation
  Conference}, pages 664--672, 2019.

\bibitem[Hansen et~al.(2008)Hansen, Niederberger, Guzzella, and
  Koumoutsakos]{hansen2008method}
Nikolaus Hansen, Andr{\'e}~SP Niederberger, Lino Guzzella, and Petros
  Koumoutsakos.
\newblock A method for handling uncertainty in evolutionary optimization with
  an application to feedback control of combustion.
\newblock \emph{IEEE Transactions on Evolutionary Computation}, 13\penalty0
  (1):\penalty0 180--197, 2008.

\bibitem[Hansen et~al.(2009{\natexlab{b}})Hansen, Finck, Ros, and
  Auger]{hansen:inria-00362633}
Nikolaus Hansen, Steffen Finck, Raymond Ros, and Anne Auger.
\newblock {Real-Parameter Black-Box Optimization Benchmarking 2009: Noiseless
  Functions Definitions}.
\newblock Research Report RR-6829, {INRIA}, 2009{\natexlab{b}}.
\newblock URL \url{https://hal.inria.fr/inria-00362633}.

\bibitem[Hansen et~al.(2016{\natexlab{c}})Hansen, Tusar, Mersmann, Auger, and
  Brockhoff]{hansen2016coco}
Nikolaus Hansen, Tea Tusar, Olaf Mersmann, Anne Auger, and Dimo Brockhoff.
\newblock Coco: The experimental procedure.
\newblock \emph{arXiv preprint arXiv:1603.08776}, 2016{\natexlab{c}}.

\bibitem[Hansen et~al.(2019)Hansen, Akimoto, and Baudis]{hansen2019cma}
Nikolaus Hansen, Youhei Akimoto, and Petr Baudis.
\newblock Cma-es/pycma on github.
\newblock \emph{Zenodo, doi}, 10, 2019.

\bibitem[Heidrich-Meisner and Igel(2009)]{heidrich2009hoeffding}
Verena Heidrich-Meisner and Christian Igel.
\newblock Hoeffding and bernstein races for selecting policies in evolutionary
  direct policy search.
\newblock In \emph{Proceedings of the 26th Annual International Conference on
  Machine Learning}, pages 401--408, 2009.

\bibitem[Holland(1992)]{holland1992genetic}
John~H Holland.
\newblock Genetic algorithms.
\newblock \emph{Scientific american}, 267\penalty0 (1):\penalty0 66--73, 1992.

\bibitem[Hwangbo et~al.(2014)Hwangbo, Gehring, Sommer, Siegwart, and
  Buchli]{hwangbo2014rock}
Jemin Hwangbo, Christian Gehring, Hannes Sommer, Roland Siegwart, and Jonas
  Buchli.
\newblock Rock∗—efficient black-box optimization for policy learning.
\newblock In \emph{2014 IEEE-RAS International Conference on Humanoid Robots},
  pages 535--540. IEEE, 2014.

\bibitem[Ib{\'a}{\~n}ez et~al.(2009)Ib{\'a}{\~n}ez, Ballerini, Cord{\'o}n,
  Damas, and Santamar{\'\i}a]{ibanez2009experimental}
Oscar Ib{\'a}{\~n}ez, Lucia Ballerini, Oscar Cord{\'o}n, Sergio Damas, and
  Jos{\'e} Santamar{\'\i}a.
\newblock An experimental study on the applicability of evolutionary algorithms
  to craniofacial superimposition in forensic identification.
\newblock \emph{Information Sciences}, 179\penalty0 (23):\penalty0 3998--4028,
  2009.

\bibitem[Johnson(2014)]{johnson2014nlopt}
Steven~G Johnson.
\newblock The nlopt nonlinear-optimization package, 2014.

\bibitem[Kakade(2001)]{kakade2001natural}
Sham~M Kakade.
\newblock A natural policy gradient.
\newblock \emph{Advances in neural information processing systems}, 14, 2001.

\bibitem[Kupcsik et~al.(2013)Kupcsik, Deisenroth, Peters, and
  Neumann]{kupcsik2013data}
A~Kupcsik, MP~Deisenroth, J~Peters, and G~Neumann.
\newblock Data-efficient contextual policy search for robot movement skills.
\newblock In \emph{Proceedings of the National Conference on Artificial
  Intelligence (AAAI)}. Bellevue, 2013.

\bibitem[Larson et~al.(2019)Larson, Menickelly, and Wild]{larson2019derivative}
Jeffrey Larson, Matt Menickelly, and Stefan~M Wild.
\newblock Derivative-free optimization methods.
\newblock \emph{Acta Numerica}, 28:\penalty0 287--404, 2019.

\bibitem[Loshchilov et~al.(2012{\natexlab{a}})Loshchilov, Schoenauer, and
  Sebag]{loshchilov2012alternative}
Ilya Loshchilov, Marc Schoenauer, and Michele Sebag.
\newblock Alternative restart strategies for cma-es.
\newblock In \emph{International Conference on Parallel Problem Solving from
  Nature}, pages 296--305. Springer, 2012{\natexlab{a}}.

\bibitem[Loshchilov et~al.(2012{\natexlab{b}})Loshchilov, Schoenauer, and
  Sebag]{loshchilov2012self}
Ilya Loshchilov, Marc Schoenauer, and Michele Sebag.
\newblock Self-adaptive surrogate-assisted covariance matrix adaptation
  evolution strategy.
\newblock In \emph{Proceedings of the 14th annual conference on Genetic and
  evolutionary computation}, pages 321--328, 2012{\natexlab{b}}.

\bibitem[Nelder and Mead(1965)]{nelder1965simplex}
John~A Nelder and Roger Mead.
\newblock A simplex method for function minimization.
\newblock \emph{The computer journal}, 7\penalty0 (4):\penalty0 308--313, 1965.

\bibitem[Osborne et~al.(2009)Osborne, Garnett, and
  Roberts]{osborne2009gaussian}
Michael~A Osborne, Roman Garnett, and Stephen~J Roberts.
\newblock Gaussian processes for global optimization.
\newblock In \emph{3rd international conference on learning and intelligent
  optimization (LION3)}, pages 1--15, 2009.

\bibitem[Otto et~al.(2021)Otto, Becker, Vien, Ziesche, and
  Neumann]{otto2021differentiable}
Fabian Otto, Philipp Becker, Ngo~Anh Vien, Hanna~Carolin Ziesche, and Gerhard
  Neumann.
\newblock Differentiable trust region layers for deep reinforcement learning.
\newblock \emph{arXiv preprint arXiv:2101.09207}, 2021.

\bibitem[Pajarinen et~al.(2019)Pajarinen, Thai, Akrour, Peters, and
  Neumann]{pajarinen2019compatible}
Joni Pajarinen, Hong~Linh Thai, Riad Akrour, Jan Peters, and Gerhard Neumann.
\newblock Compatible natural gradient policy search.
\newblock \emph{Machine Learning}, 108\penalty0 (8):\penalty0 1443--1466, 2019.

\bibitem[Paraschos et~al.(2013)Paraschos, Daniel, Peters, and
  Neumann]{NIPS2013_e53a0a29}
Alexandros Paraschos, Christian Daniel, Jan~R Peters, and Gerhard Neumann.
\newblock Probabilistic movement primitives.
\newblock In C.~J.~C. Burges, L.~Bottou, M.~Welling, Z.~Ghahramani, and K.~Q.
  Weinberger, editors, \emph{Advances in Neural Information Processing
  Systems}, volume~26. Curran Associates, Inc., 2013.

\bibitem[Peters et~al.(2010)Peters, Mulling, and Altun]{peters2010relative}
Jan Peters, Katharina Mulling, and Yasemin Altun.
\newblock Relative entropy policy search.
\newblock In \emph{Twenty-Fourth AAAI Conference on Artificial Intelligence},
  2010.

\bibitem[Price(1997)]{price1997dev}
Kenneth Price.
\newblock Differential evolution vs. the functions of the second {ICEO}.
\newblock In \emph{Proceedings of the {IEEE} International Congress on
  Evolutionary Computation}, pages 153--157, Piscataway, NJ, USA, 1997. IEEE.
\newblock \doi{10.1109/ICEC.1997.592287}.

\bibitem[Price(1983)]{price1983global}
WL1551847 Price.
\newblock Global optimization by controlled random search.
\newblock \emph{Journal of optimization theory and applications}, 40\penalty0
  (3):\penalty0 333--348, 1983.

\bibitem[Rubinstein and Kroese(2004)]{rubinstein2004cross}
Reuven~Y Rubinstein and Dirk~P Kroese.
\newblock \emph{The cross-entropy method: a unified approach to combinatorial
  optimization, Monte-Carlo simulation, and machine learning}, volume 133.
\newblock Springer, 2004.

\bibitem[Schaul et~al.(2010)Schaul, Bayer, Wierstra, Sun, Felder, Sehnke,
  R{\"u}ckstie{\ss}, and Schmidhuber]{schaul2010pybrain}
Tom Schaul, Justin Bayer, Daan Wierstra, Yi~Sun, Martin Felder, Frank Sehnke,
  Thomas R{\"u}ckstie{\ss}, and J{\"u}rgen Schmidhuber.
\newblock Pybrain.
\newblock \emph{Journal of Machine Learning Research}, 11\penalty0
  (ARTICLE):\penalty0 743--746, 2010.

\bibitem[Schulman et~al.(2015)Schulman, Levine, Abbeel, Jordan, and
  Moritz]{schulman2015trust}
John Schulman, Sergey Levine, Pieter Abbeel, Michael Jordan, and Philipp
  Moritz.
\newblock Trust region policy optimization.
\newblock In \emph{International conference on machine learning}, pages
  1889--1897. PMLR, 2015.

\bibitem[Spall(2005)]{spall2005introduction}
James~C Spall.
\newblock \emph{Introduction to stochastic search and optimization: estimation,
  simulation, and control}, volume~65.
\newblock John Wiley \& Sons, 2005.

\bibitem[Sun et~al.(2009)Sun, Wierstra, Schaul, and
  Schmidhuber]{sun2009efficient}
Yi~Sun, Daan Wierstra, Tom Schaul, and J{\"u}rgen Schmidhuber.
\newblock Efficient natural evolution strategies.
\newblock In \emph{Proceedings of the 11th Annual conference on Genetic and
  evolutionary computation}, pages 539--546, 2009.

\bibitem[Sutton et~al.(1999)Sutton, McAllester, Singh, and
  Mansour]{sutton1999policy}
Richard~S Sutton, David McAllester, Satinder Singh, and Yishay Mansour.
\newblock Policy gradient methods for reinforcement learning with function
  approximation.
\newblock \emph{Advances in neural information processing systems}, 12, 1999.

\bibitem[Todorov et~al.(2012)Todorov, Erez, and Tassa]{todorov2012mujoco}
Emanuel Todorov, Tom Erez, and Yuval Tassa.
\newblock Mujoco: A physics engine for model-based control.
\newblock In \emph{2012 IEEE/RSJ international conference on intelligent robots
  and systems}, pages 5026--5033. IEEE, 2012.

\bibitem[Wierstra et~al.(2014)Wierstra, Schaul, Glasmachers, Sun, Peters, and
  Schmidhuber]{wierstra2014natural}
Daan Wierstra, Tom Schaul, Tobias Glasmachers, Yi~Sun, Jan Peters, and
  J{\"u}rgen Schmidhuber.
\newblock Natural evolution strategies.
\newblock \emph{The Journal of Machine Learning Research}, 15\penalty0
  (1):\penalty0 949--980, 2014.

\bibitem[Winter et~al.(2008)Winter, Brendel, Pechlivanis, Schmieder, and
  Igel]{winter2008registration}
Susanne Winter, Bernhard Brendel, Ioannis Pechlivanis, Kirsten Schmieder, and
  Christian Igel.
\newblock Registration of ct and intraoperative 3-d ultrasound images of the
  spine using evolutionary and gradient-based methods.
\newblock \emph{IEEE Transactions on Evolutionary Computation}, 12\penalty0
  (3):\penalty0 284--296, 2008.

\bibitem[Zabinsky(2010)]{zabinsky2010random}
Zelda~B Zabinsky.
\newblock Random search algorithms.
\newblock \emph{Wiley Encyclopedia of Operations Research and Management
  Science}, 2010.

\end{thebibliography}

\end{document}